\documentclass{article}
\usepackage{arxiv}

\usepackage[utf8]{inputenc} 
\usepackage[T1]{fontenc}    
\usepackage{hyperref}       
\usepackage{url}            
\usepackage{booktabs}       
\usepackage{amsfonts}       
\usepackage{nicefrac}       
\usepackage{microtype}      
\usepackage{lipsum}
\usepackage{cite}
\usepackage{amsfonts}
\usepackage{graphicx}
\usepackage{soul}
\usepackage{textcomp}
\usepackage{setspace}
\usepackage{placeins}
\usepackage[utf8]{inputenc} 
\usepackage{tabularx}
\usepackage{multirow}
\usepackage{adjustbox}
\usepackage{tabulary}
\usepackage{textcomp}
\usepackage{mathtools}
\usepackage{gensymb}
\usepackage[percent]{overpic}

\usepackage{amsmath}
\usepackage{amssymb}
\usepackage{mathrsfs}
\usepackage{amsthm}
\usepackage{algorithm}
\usepackage{subcaption}
\usepackage{hyperref}
\usepackage{graphicx}
\usepackage{mwe}
\usepackage[noend]{algpseudocode}
\usepackage{bm}
\usepackage{float}
\usepackage{parskip}

\DeclarePairedDelimiter\floor{\lfloor}{\rfloor}

\DeclareMathOperator*{\argmin}{arg\,min}
\newcommand{\integerset}[1]{\left\{1,\ldots,{#1}\right\}}

\usepackage{array}
\newcolumntype{L}{>{\arraybackslash}p{5cm}}
\newlength{\temp}
\newtheorem{assumption}{Assumption}

\graphicspath{ {./figures/} }
\def\BibTeX{{\rm B\kern-.05em{\sc i\kern-.025em b}\kern-.08em
		T\kern-.1667em\lower.7ex\hbox{E}\kern-.125emX}}

\usepackage{url} 

\DeclareMathOperator{\atantwo}{atan2}
\DeclareMathOperator{\clip}{clip}
\usepackage[noend]{algpseudocode}
\usepackage{bm}
\usepackage{float}
\usepackage{parskip}
\title{COLREG-Compliant Collision Avoidance for Unmanned Surface Vehicle using Deep Reinforcement Learning}

\author{
	Eivind Meyer \\
	Department of Engineering Cybernetics\\
	Norwegian University of Science and Technology\\
	Trondheim, Norway \\
	\texttt{eiv.meyer@gmail.com} \\
	\And
	Amalie Heiberg \\
	Department of Engineering Cybernetics\\
	Norwegian University of Science and Technology\\
	Trondheim, Norway \\
	\texttt{heibergamalie@gmail.com} \\
	\And
	Adil Rasheed
	\thanks{Corresponding author (adil.rasheed@ntnu.no)} \\
	Department of Engineering Cybernetics\\
	Norwegian University of Science and Technology\\
	Trondheim, Norway \\
	\texttt{adil.rasheed@ntnu.no} \\
	\And
	Omer San \\
	School of Mechanical and Aerospace Engineering\\
	Oklahoma State University \\
	Stillwater, Oklahoma, 74078-5016 USA \\
	\texttt{osan@okstate.edu} \\
}

\begin{document}
	\maketitle
	
	\begin{abstract}
		Path Following and Collision Avoidance, be it for unmanned surface vessels or other autonomous vehicles, are two fundamental guidance problems in robotics. For many decades, they have been subject to academic study, leading to a vast number of proposed approaches. However, they have mostly been treated as separate problems, and have typically relied on non-linear first-principles models with parameters that can only be determined experimentally. The rise of Deep Reinforcement Learning (DRL) in recent years suggests an alternative approach: end-to-end learning of the optimal guidance policy from scratch by means of a trial-and-error based approach. In this article, we explore the potential of Proximal Policy Optimization (PPO), a DRL algorithm with demonstrated state-of-the-art performance on Continuous Control tasks, when applied to the dual-objective problem of controlling an underactuated Autonomous Surface Vehicle in a COLREGs compliant manner such that it follows an a priori known desired path while avoiding collisions with other vessels along the way. Based on high-fidelity elevation and AIS tracking data from the Trondheim Fjord, an inlet of the Norwegian sea, we evaluate the trained agent's performance in challenging, dynamic real-world scenarios where the ultimate success of the agent rests upon its ability to navigate non-uniform marine terrain while handling challenging, but realistic vessel encounters.
	\end{abstract}

	\twocolumn
	\section{Introduction}
	Autonomous vehicles is one of the most interesting prospects associated with the rise of Artificial Intelligence (AI) and Machine Learning (ML) in recent years. Specifically, the success of Deep Learning (DL) applications in an ever-increasing number of domains, ranging from computer vision to imperfect-information games, has put the former pie-in-the-sky proposal of self-driving vehicles on the horizon of technological development.
	
	While automated path following, at least in the maritime domain, has been a relatively trivial endeavor in the light of classical control theory and is a well-established field of research \cite{BreivikPathFollowing, CaharijaIntegralLoS, moe2014path, fossen2003line, GuerreroObsBasedPDTrajectoryTracking, belleter2018observer, ReisRobustPathFollowing, PaliottaPathFollowing,
		liu2015path,liu2017path,
		singh2018constrained}, considerably more advanced capabilities are required to navigate unknown, dynamic environments; characteristics that, generally speaking, apply to the real world. Reactive collision avoidance, i.e. the ability to, based on a sensor-based perception of the local environment, perform evasive manoeuvres that mitigate collision risk, remains a very challenging undertaking (e.g., see \cite{campbell2012review,zhao2016real,xie2019ship,shi2019study}). 
	
	This is not to say, however, that the topic is not well-researched; a wide variety of approaches have been proposed, including especially (but not exhaustively) artificial potential field methods \cite{Khatib:1986:ROA:6806.6812, BorensteinKoren, Panagou}, dynamic window methods \cite{FoxDyn, highspeeddynwin, ModDynWid}, velocity obstacle methods \cite{journals/ijrr/FioriniS98, proactivecolavoidbrekke} and optimal control-based methods \cite{chenBarrier, hamiltonJacobiMITCHELL, branchingMPCEriksen, ibHagenMPC, BITAR2018389}. However, it appears from a literature review that, when applied to autonomous vehicles with non-holonomic and real-time constraints, the approaches suggested so far suffer from one or more of the following drawbacks \cite{YanObstacleAvoidance, KorenPFMLimitations, moe2016set, MINGUEZ2002397}:
	
	\begin{itemize}
		\item Unrealistic assumptions, or neglect, of the vessel dynamics.
		\item Inability to scale to environments of non-trivial complexity (e.g. multi-obstacle scenarios).
		\item Excessive computation time requirements.
		\item Disregard for desirable output trajectory properties, including smoothness, continuity, feasibility and safety.
		\item Incompatibility with external environmental forces such as wind, currents and waves.
		\item Stability issues caused by singularities.
		\item Sub-optimal outputs due to local minima.
		\item Requirement of a precise mathematical model of the controlled vessel.
	\end{itemize}
	
	Focusing on the maritime domain, this paper will explore how Deep Reinforcement Learning (DRL), a machine learning paradigm concerned with using DL for iteratively approximating optimal behavior policies in unknown environments, can be used for training an end-to-end autopilot mechanism capable of avoiding collisions at sea. For the simpler problem of treating path following and collision avoidance as separate challenges, DRL-based methods have already demonstrated remarkable potential, yielding promising results in a multitude of studies, including especially \cite{MARTINSEN2018Straight, martinsen2018curved, zhang2020deepPF, WOO2019155DRLPF, MartinsenRLTrackingFrontiers} for the former problem domain and \cite{Guo_2020, LIN2019106327, ZHAO2019106436, Lin2019ResearchOU} for the latter.  
	
	For a preliminary study, we simulate a small-sized supply ship model, equip it with a rangefinder sensor suite and train a DRL-based controller using Proximal Policy Optimization (PPO). A carefully constructed reward function, which balances the prioritization of path adherence versus that of collision avoidance (which can be considered competing objectives), is used to guide the agent's learning process. Finally, we evaluate its performance in challenging, dynamic test scenarios reconstructed from real-world terrain and maritime traffic data.
	\begin{figure}[ht!]
		\includegraphics[trim={0cm 0cm 0cm 0cm},clip,width=0.5\textwidth]{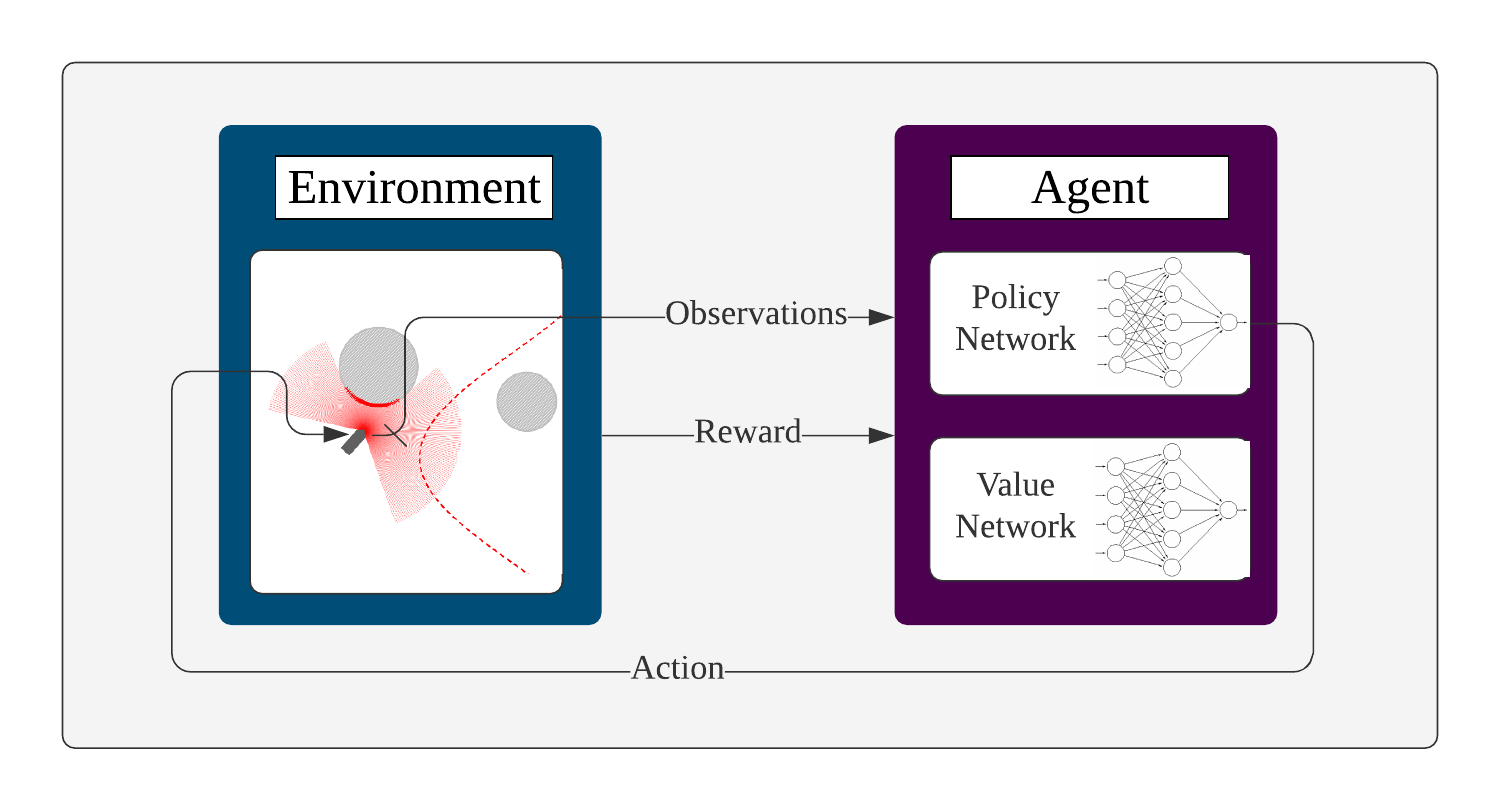}
		\caption[Flowchart of reinforcement learning]{Flowchart outlining the structure of the guidance system explored in this study. At each time-step, the agent receives an observation vector $s^{(t)}$, and then, according to its policy $\pi$, which is implemented as a neural network, outputs an action (i.e. control vector) $a^{(t)}$, influencing the state of the simulated environment. During training, the agent's policy is continuously improved by means of gradient ascent based on the reward signal $r^{(t)}$ that it receives at each time-step. This constant feedback enables the agent, whose policy is initially nothing more than a clean slate with no intelligent characteristics, to improve its capabilities through a trial-and-error based approach. Its learning objective is simple: Find the policy that yields the highest expectation of the agent's long-term future reward.}
		\label{fig:rlflowchart}
	\end{figure}
	\section{Motivation}
	
	Arguably, the most promising aspect of autonomous vessels is not the obvious economic impact resulting from increased efficiency and the replacement of costly human labor, but instead the potential to eliminate injuries and material damage caused by collisions. According to the European Maritime Safety Agency, which annually publishes statistics on maritime accidents related to the EU member states, almost half of casualties at sea are ``navigational in nature, including contact, collision and grounding or stranding'' \cite{EuropeanMaritimeSafetyAgencyReport2020}.
	
\begin{figure}[ht!]
		\centering
		\includegraphics[width=1.0\linewidth]{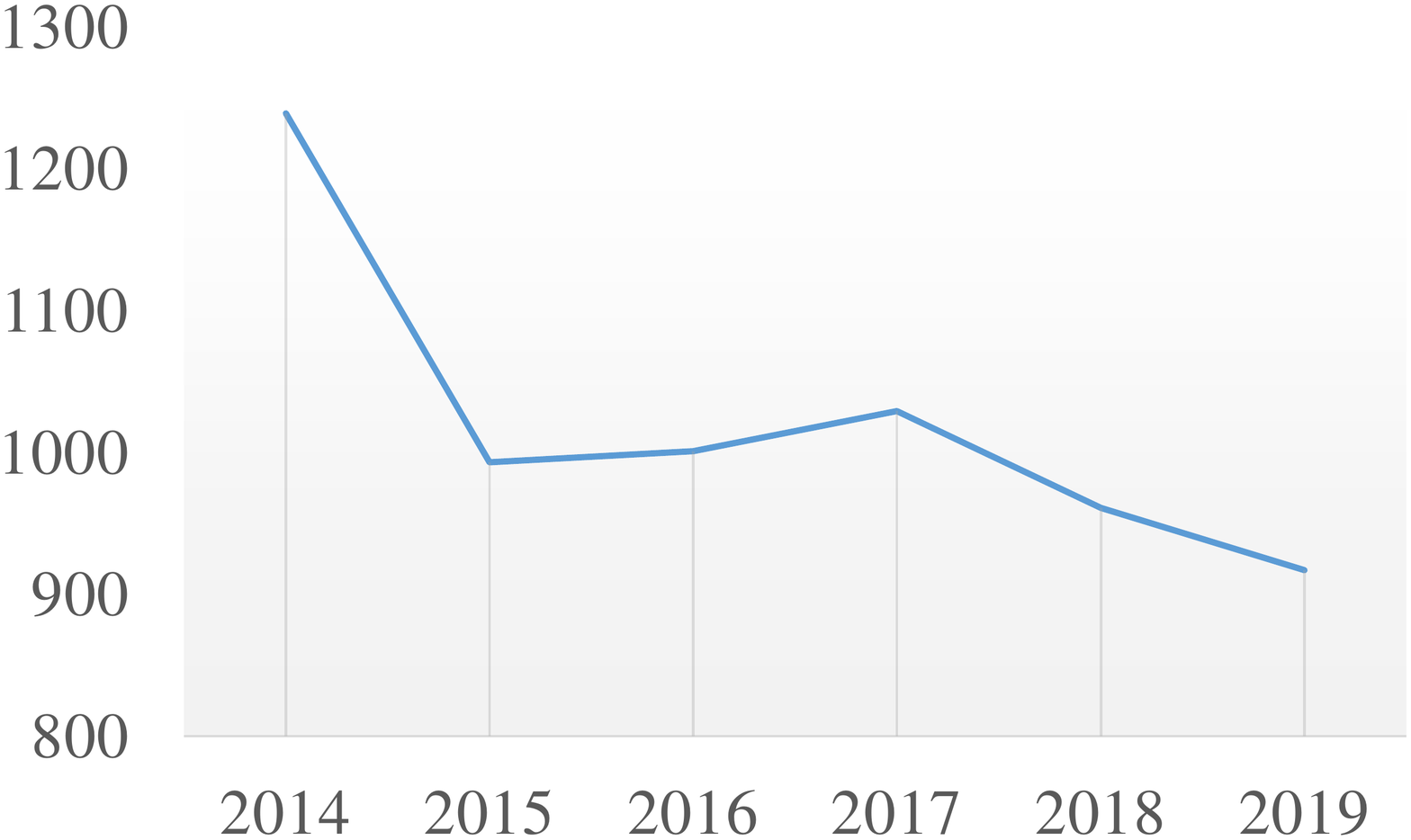}
		\caption[Yearly injuries plot]{Human injuries per year according to maritime accident statistics published by the European Maritime Safety Agency.}
		\label{fig:accidentplot}
	\end{figure}

	Validating a DRL-based approach to vessel guidance in a simulated environment can pave the way for applying the technology on a real, physical vessel. Since maritime collisions, of which 65.8\% can be attributed to human error \cite{EuropeanMaritimeSafetyAgencyReport2019}, account for hundreds of injuries each year in the EU alone (as shown in Figure \ref{fig:accidentplot}), a positive result could be a preliminary step on the important path towards the adoption of AI systems for autonomous vessel guidance. Due to the limitations of existing methods, this is yet to take place on a large scale. 
	\section{Background}
	
	\subsection{Maritime Navigation Rules}
	
	For collision avoidance at sea, adherence to the International Regulations for Preventing Collisions at Sea (COLREGs) \cite{COLREGS} is crucial. Before autonomous vessels became a possibility, the COLREGs were formulated to prevent collisions between two or more vessels. The two main takeaways from these rules relevant for this work are; 1) the give way vessel should take early and substantial action, and 2) safe speed should be ensured at all times, such that course alteration is effective towards avoiding collisions where there is sufficient sea-room. Furthermore, the following rules provide clear instructions on how maritime vessels should behave upon encounters with other ships.
	
	\textbf{Rule 14: Head-on situation}
	
	\begin{quote}
		\textit{(a) When two power-driven vessels are meeting on reciprocal or nearly reciprocal courses so as to involve risk of collision each shall alter her course to starboard so that each shall pass on the port side of the other.}
	\end{quote}
	
	\textbf{Rule 15: Crossing situation}
	
	\begin{quote}
		\textit{When two power-driven vessels are crossing so as to involve risk of collision, the vessel which has the other on her own starboard side shall keep out of the way and shall, if the circumstances of the case admit, avoid crossing ahead of the other vessel. }
	\end{quote}
	
	\textbf{Rule 16: Action by give-way vessel}
	
	\begin{quote}
		\textit{Every vessel which is directed to keep out of the way of another vessel shall, so far as possible, take early and substantial action to keep well clear.}
	\end{quote}
	
	\textbf{Rule 18: Responsibilities between vessels}
	
	\begin{quote}
		\textit{(a) A power-driven vessel underway shall keep out of the way of: \\ (ii) a vessel restricted in her ability to manoeuvre.}
	\end{quote}
	
	Since the vessel controlled by the RL agent (the own-ship) is significantly smaller than the vessels encountered, it is, as a result of Rule 18, required to act as the give-way vessel in all situations.

		\begin{figure}[ht!]
		\centering
		\begin{subfigure}{0.49\linewidth}
			{\includegraphics[width=\linewidth]{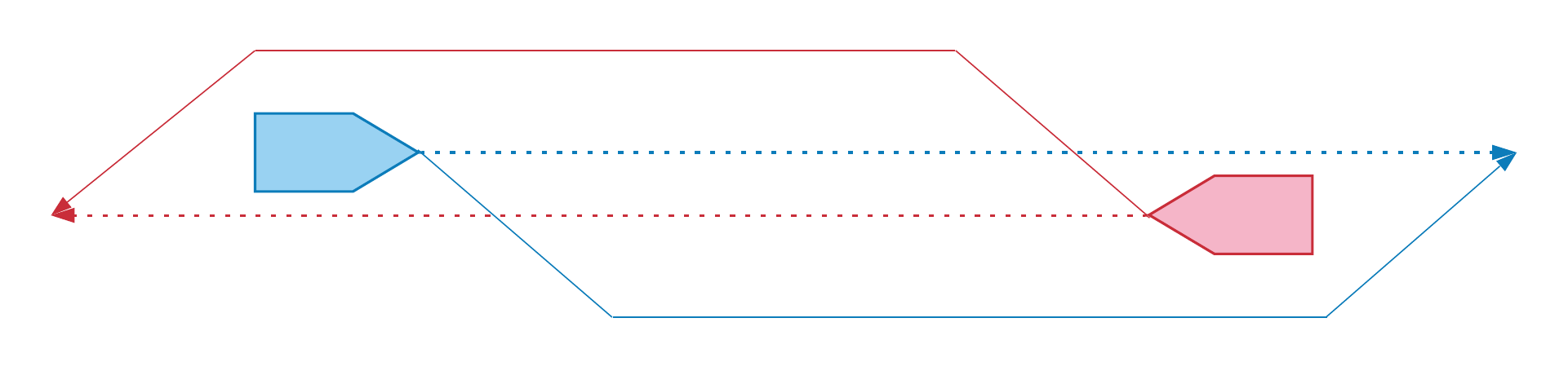}}
			\subcaption{Head-on situation, target-vessel free to move.}
			\label{fig:headon_free}
		\end{subfigure}
		\begin{subfigure}{0.49\linewidth}
			{\includegraphics[width=\linewidth]{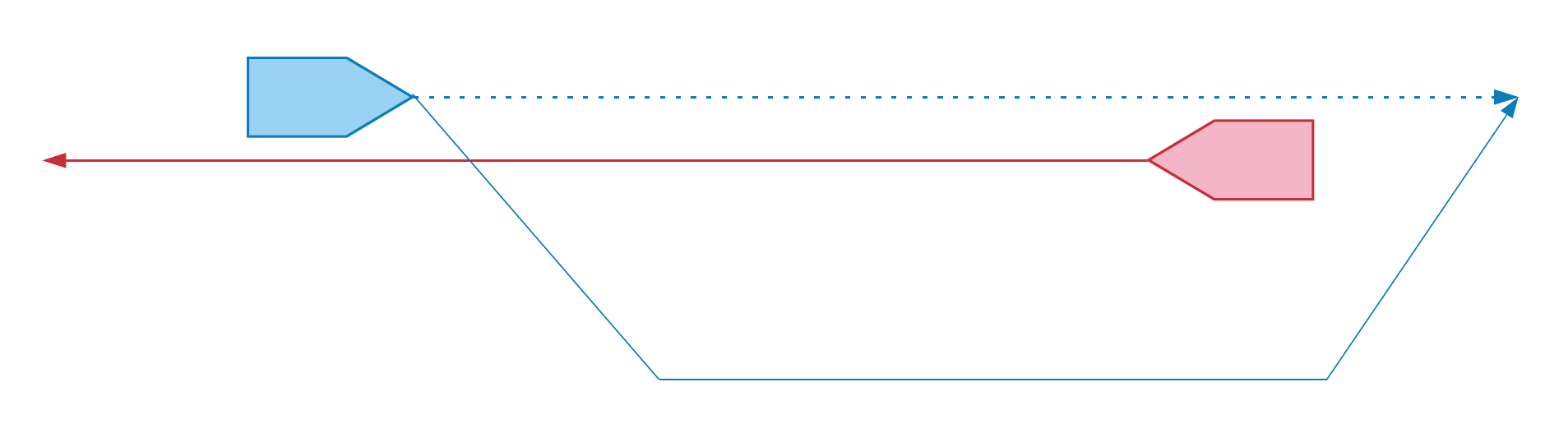}}
			\subcaption{Head-on situation, target-vessel restricted in movement.}
			\label{fig:headon_restricted}
		\end{subfigure}
		\begin{subfigure}{0.49\linewidth}
			{\includegraphics[width=\linewidth]{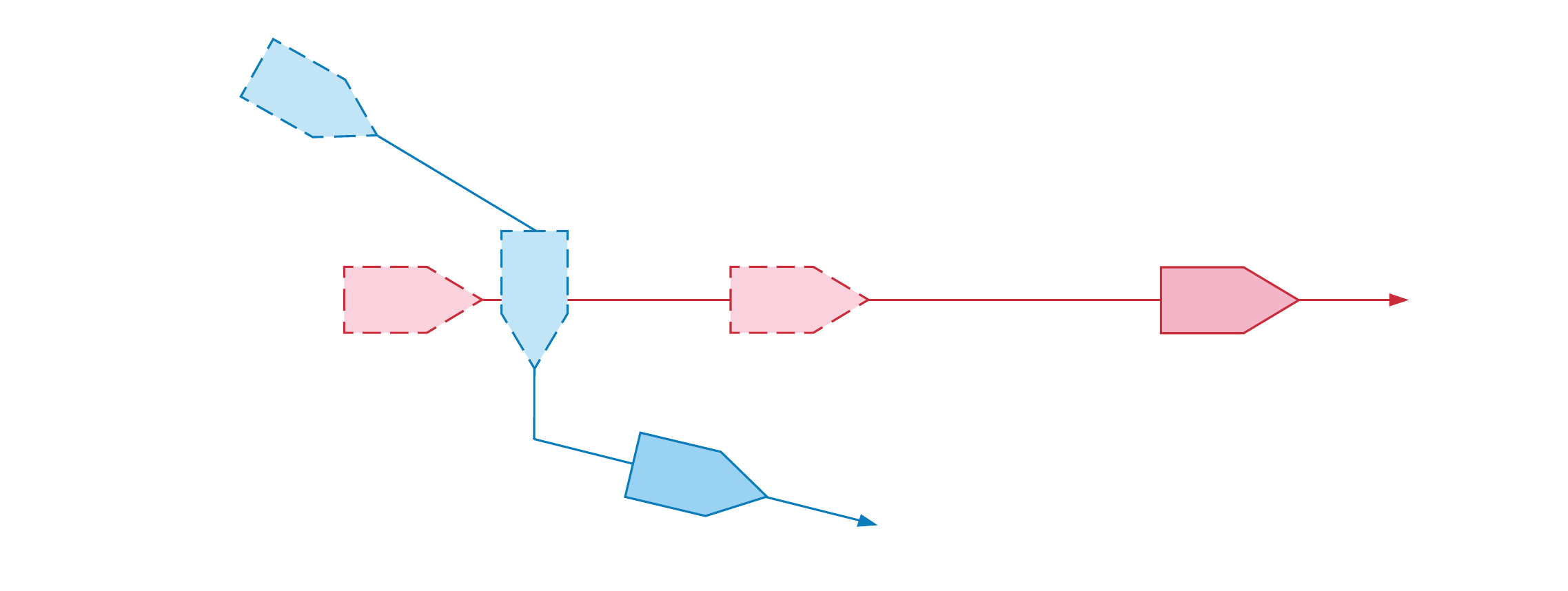}}
			\subcaption{Crossing situation, target-vessel free to move.}
		\end{subfigure}
	\begin{subfigure}{0.49\linewidth}
		{\includegraphics[width=\linewidth]{figures/crossing.pdf}}
		\subcaption{Crossing situation, target-vessel restricted in movement.}
	\end{subfigure}
		\caption{Expected behavior from the own-ship (colored in blue) in head-on and crossing encounters according to COLREGs.}
		\label{fig:colreg_complianec}
	\end{figure}

	%
	
	\label{section:backgrund}
	
	\subsection{Dynamics of a marine vessel}
	\subsubsection{Coordinate frames}
	
	In order to model the dynamics of marine vessels, one must first define the coordinate frames. Two coordinate frames typically used in vehicle control applications are of particular interest: The geographical North-East-Down (NED) and the body frame. The NED reference frame $\{n\} = (x_n, y_n, z_n)$ forms a tangent plane to the Earth's surface, making it useful for terrestrial navigation. Here, the $x_n$-axis is directed north, the $y_n$-axis is directed east and the $z_n$-axis is directed towards the center of the earth.
	
	\begin{assumption}[State space restriction]\label{as:restricted_motion}The vessel is always located on the surface and thus there is no heave motion. Also, there is no pitching or rolling motion.\end{assumption}
	
	The origin of the body-fixed reference frame $\{b\} = (x_b, y_b, z_b)$ is fixed to the current position of the vessel in the NED-frame, and its axes are aligned with the heading of the vessel such that $x_b$ is the longitudinal axis, $y_b$ is the transversal axis and $z_b$ is the normal axis pointing downwards. However, as the vessel is restricted to surface level motion in our application, only the North and East components are of interest.
	\begin{figure}[ht!]
		\includegraphics[clip,trim={1.7cm 1.7cm 1.7cm 1.7cm},width=0.5\textwidth]{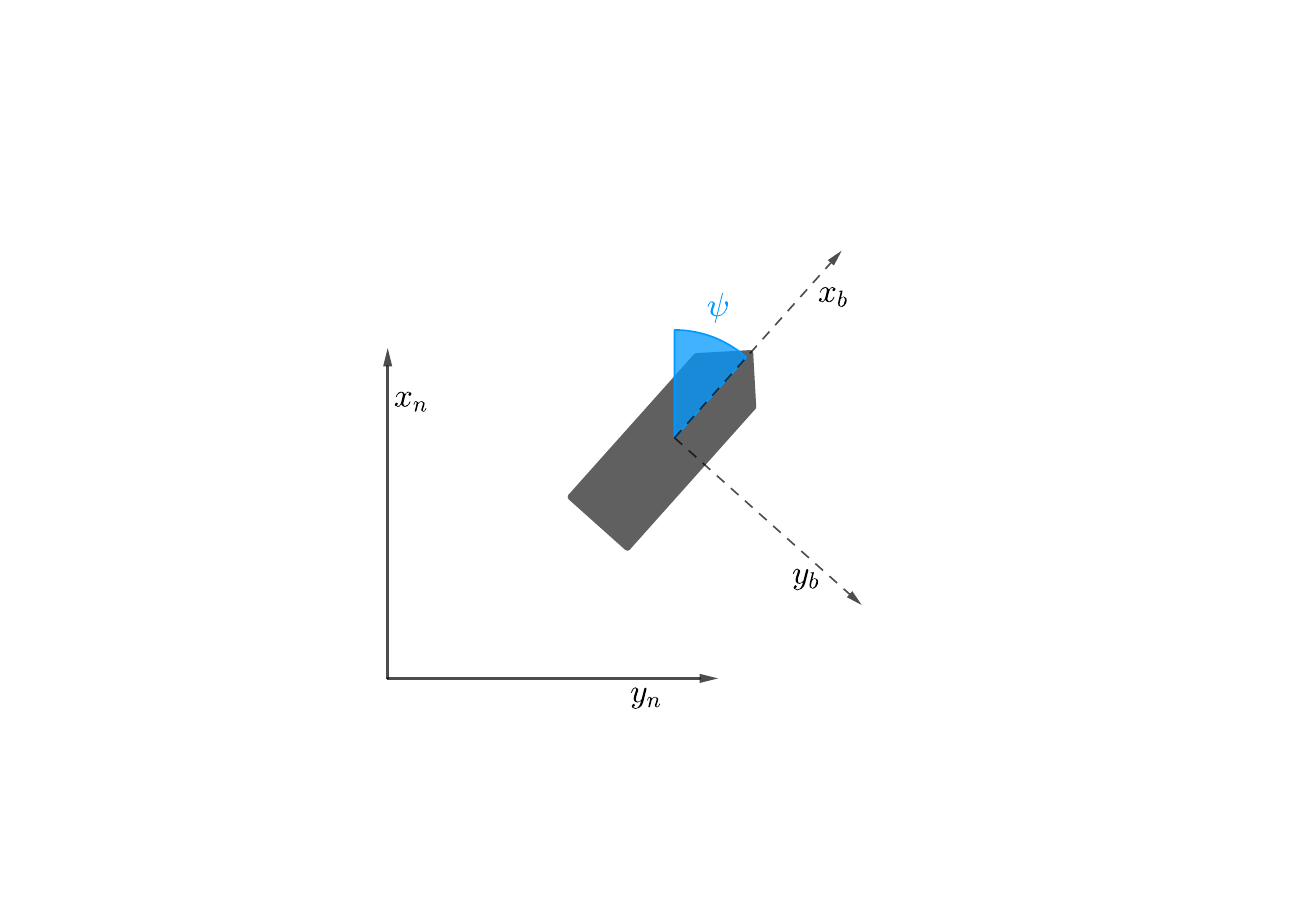}
		\caption[Coordinate frames]{Illustration of the NED and body coordinate frames.}
		\label{fig:coordinateframes}
	\end{figure}
	\subsubsection{State variables}\label{section:sname_state_variables}
	
	Following SNAME notation \cite{SNAME}, the state vector consists of the generalized coordinates $\bm{\eta} = \left[ x^n, y^n, \psi \right]^T$, where $x^n$ and $y^n$ are the North and East positions, respectively, in the reference frame $\{n\}$, and $\psi$ is the yaw angle, i.e. the current angle between the vessel's longitudinal axis $x_b$ and the North axis $x_n$. Correspondingly, the translational and angular velocity vector $\bm{\nu} = \left[ u, v, r \right]^T$ consists of the surge (i.e. forward) velocity $u$, the sway (i.e. sideways) velocity $v$ as well as yaw rate $r$.
	
	\subsubsection{Vessel model}\label{section:vessel_dynamics}
	
	To facilitate further research, we base the vessel dynamics on CyberShip II, a 1:70 scale replica of a supply ship which has a length of 1.255 m and mass of 23.8 kg \cite{Skjetne2004ANS}. Training the RL algorithm on a small vessel, such as CyberShip II, would allow for a relatively straight-forward deployment on a real-world model ship for further testing of the algorithm.  However, the symbolic representation of the dynamics of a surface vessel, which is obtained from well-researched ship maneuvering theory, is the same regardless of the vessel - the distinctions lie solely in the numerical matrix parameters. Thus, if it can be demonstrated that an RL agent can control a small-sized model ship in an intelligent manner, there is reason to believe that controlling a full-sized ship would be within its reach.
	
	As it is equipped with rudders and propellers aft, as well as one bow thruster fore, Cybership II is a fully actuated ship. This means that it could, in principle, be commanded to follow an arbitrary trajectory in the state space, as it is able to accelerate independently in every relevant
	DOF simultaneously. However, for the purpose of simplifying the RL agent's action space, we disregard the bow thruster in this study and allow only the aft thrusters and control surfaces to be applied by the Reinforcement Learning (RL) agent as control signals. This omission is further motivated by the fact that bow thrusters have limited effectiveness at higher speeds     \cite{ShipHeadingSpeedControlSorensenBreivikEriksen}. Thus, the control vector can be modelled as $\bm{f} = \left[T_u, T_r\right]^T$, where $T_u$ represents the force input in surge and $T_r$ represents the moment input in yaw.
	
	\begin{assumption}[Calm sea]\label{as:calm_sea} There are no external disturbances to the vessel such as wind, ocean currents or waves.\end{assumption}
	
	\noindent Given Assumption \ref{as:calm_sea}, the 3-DOF vessel dynamics can be expressed in a compact matrix-vector form as 
	\begin{equation}
		\begin{aligned}
			\dot{\bm{\eta}} &= \mathbf{R}_{z,\psi}(\bm{\eta})   \bm{\nu} \\
			\mathbf{M} \bm{\dot{\nu}} + \mathbf{C}(\bm{\nu}) \bm{\nu} + \mathbf{D}(\bm{\nu}) \bm{\nu} &= \mathbf{B} \bm{f}
		\end{aligned}
	\end{equation}
	where $\mathbf{R}_{z,\psi}$ represents a rotation of $\psi$ radians around the $z_n$-axis as defined by
	\begin{equation*}
		\begin{aligned}
			\mathbf{R}_{z,\psi} &= \begin{bmatrix}
				\cos{\psi} & -\sin{\psi} & 0 \\
				\sin{\psi} & \cos{\psi} & 0 \\
				0 & 0 & 1
			\end{bmatrix}
		\end{aligned}
	\end{equation*}
	\noindent Furthermore, $\mathbf{M} \in \mathbb{R}^{3\times3}$ is the mass matrix and includes the effects of both rigid-body and added mass, $\mathbf{C}(\bm{\nu}) \in \mathbb{R}^{3\times3}$ incorporates centripetal and Coriolis effects and $\mathbf{D}(\bm{\nu}) \in \mathbb{R}^{3\times3}$ is the damping matrix. Finally, $\mathbf{B} \in \mathbb{R}^{3\times2}$ is the actuator configuration matrix. The numerical values of the matrices are taken from \cite{SKJETNE2004203}, where the model parameters were estimated experimentally for CyberShip II in a marine control laboratory.
	
	\subsection{Deep reinforcement learning}
	
	Applications of RL on high-dimensional, continuous control tasks heavily rely on function approximators to generalize over the state space. Even if classical, tabular solution methods such as Q-learning can be made to work (provided a discretizing of the continuous action space), this is not considered an efficient approach for control applications \cite{lillicrap2015continuous}. In recent years, given their remarkable generalization ability over high-dimensional input spaces, the dominant approach has been the application of deep neural networks which are optimized by means of gradient methods. There are, however, different approaches to how the networks are utilized, and thus their semantic interpretation in the context of the learning agent differs. In Q-Learning-based methods such as Deep Q-Learning (DQN) \cite{MnihDQN}, a deep neural network is used to predict the expected value (i.e. long-term, cumulative reward) of state-action pairs, which reduces the policy to an optimization problem over the set of available actions given the current state. In gradient-based policy methods, on the other hand, the policy itself is implemented as a deep neural network whose weights are optimized by means of gradient ascent (or approximations thereof). Lately, several algorithms built on this principle have gained a large traction in the RL research community, most notably Deep Deterministic Policy Gradient (DDPG) \cite{lillicrap2015continuous}, Asynchronous Advantage Actor Critic (A3C) \cite{mnih2016asynchronous} and Proximal Policy Optimization (PPO) \cite{schulman2017proximal}. For continuous control tasks, this family of DRL methods is commonly considered to be the more efficient approach \cite{tai2016survey}. Based on previous work, where the PPO algorithm significantly outperformed other methods on a learning problem similar to the one covered in this study \cite{meyer_ASV_IEEE}, we focus our efforts on this method.
	
	\subsubsection{RL Preliminaries}
	First, we model the interplay between the agent and the environment as an infinite-horizon discounted Markov Decision Process (MDP), formally defined by the 6-tuple $(\mathcal{S}, \mathcal{A}, p, p_0, r, \Omega, o, \gamma)$ where
	\begin{itemize}
		\item $\mathcal{S}$ is the state space,
		\item $\mathcal{A}$ is the action space,
		\item $p : \mathcal{S} \times \mathcal{A} \times \mathcal{S} \to [0, 1]$ defines the conditional transition probabilities for the next state $s'$ such that $p(s' |s, a) = Pr(S_{t+1}=s'| S_{t}=s, A_{t}=a)$,
		\item $p_0 : \mathcal{S} \to [0, 1]$ is initial state distribution, i.e. $p_o(s) = Pr(S_0=s)$,
		\item $r : \mathcal{S} \times \mathcal{A} \to \mathbb{R}$ returns the numeric reward at each time-step as function of the current state and applied action,
		\item $\gamma \in [0, 1]$ is the discount factor for future rewards.
	\end{itemize}
	The agent draws its actions from its policy $\pi$. The policy may be a deterministic function (as in DDPG), but in the context of PPO, it is modelled as a stochastic function. The conditional action distribution given the current state $s$ is given by $\pi(a|s) : \mathcal{S} \times \mathcal{A} \to \to [0, 1] = Pr(A_{t}=a| S_{t}=s)$. Specifically, we assume that the agent is drawing actions from a non-uniform multivariate Gaussian distribution whose mean is outputted by a neural network parametrized by the weights $\theta$. Formally, this translates to $a_{t} \sim \pi(s_{t})$, where $t$ is the current time-step.
	
	Next, we introduce the state-value function $V^{\pi}(s)$ and the action-value function $Q^{\pi}(s, a)$. $V^{\pi}(s)$ is the expected return from time $t$ onwards given an initial state $s$, whereas $Q^{\pi}(s, a)$ is the expected return from time $t$ onwards, but conditioned on the current action $a_t$. Formally, we have that
	\begin{subequations}
		\begin{align}
			V^{\pi}(s_t) &= \mathbb{E}_{s_{i \geq t}, a_{i \geq t} \sim \pi} \left[ R_t | s_t \right]\\
			Q^{\pi}(s_t, a_t) &= \mathbb{E}_{s_{i \geq t}, a_{i \geq t} \sim \pi} \left[ R_t | s_t, a_t \right]
		\end{align}
	\end{subequations}
	\noindent where the random variable $R_t$ represents the reward at time-step $t$.
	
	\subsubsection{Policy gradients}
	
	The stochasticity of the policy enables us to translate the RL problem, i.e. the search for the optimal policy, into the problem of optimizing the expectation 
	\begin{equation}
		J(\theta) = \mathbb{E}_{s_i, a_i \sim \pi(\bm{\theta)}} \left[ R_0 \right]
	\end{equation}
	The family of policy gradient methods, to which PPO belongs, approach gradient ascent by updating the parameter vector $\theta$ according to the approximation $ \theta_{t+1} \gets \alpha \theta_{t} + \widehat{\nabla_{\theta} J(\theta)}$, where $\widehat{\nabla_{\theta} J(\theta)}$ is a stochastic estimate of $\nabla_{\theta} J(\theta)$ satisfying $\mathbb{E}{\left[ \widehat{\nabla_{\theta} J(\theta)} \right]} = \nabla_{\theta} J(\theta)$. From the policy gradient theorem \cite{pgmforreinforcementlearning} we have that the policy gradient $\nabla_{\theta} J(\theta)$ satisfies
	\begin{equation}\label{eq:policy_grad_theorem}
		\nabla_{\theta} J(\theta) \propto \sum_{s}{\mu(s)\sum_{a}{\nabla_{\theta} \pi(a | s) Q^{\pi}(s, a)}}
	\end{equation}
	\noindent where $\mu$ is the steady state distribution under $\pi$ such that $\mu(s) = \lim_{t\to\infty}{Pr\{S_t = s | A_{0:{t-1}} \sim \pi\}}$. Following the steps outlined in \cite{sutton1998introduction}, this can be algebraically transformed to
	\begin{equation}
		\nabla_{\theta} J(\theta) \propto \mathbb{E}_{\pi}{\left[ \nabla_{\theta}{\ln{\pi(A_t | S_t)}} Q^{\pi}(S_t, A_t) \right]}   
	\end{equation}
	\noindent Also, it can be shown that one can greatly reduce the variance of this expression by replacing the state-action value function $Q^{\pi}(s, a)$ in Equation \ref{eq:policy_grad_theorem} by $Q^{\pi}(s, a) - b(s)$, where the \textbf{baseline} function $b(s)$ can be an arbitrary function not depending on the action $a$, without introducing a bias in the estimate. Commonly, $b(s)$ is set to be the state value function $V^{\pi}$, which yields the \textbf{advantage} function
	\begin{equation}\label{eq:advantage_function}
		A^{\pi}(s, a) = Q^{\pi}(s, a) - V^{\pi}(s)
	\end{equation}
	\noindent which represents the expected improvement obtained by an action compared to the default behavior. This leads to
	\begin{equation}
		\nabla_{\theta} J(\theta) \propto \mathbb{E}_{\pi}{\left[ \nabla_{\theta}{\log{\pi(A_t | S_t)}} A^{\pi}(s, a) \right]}
	\end{equation}
	\noindent Thus, an unbiased empirical estimate based on $N$ episodic policy rollouts of the policy gradient $\nabla_{\theta} J(\theta)$ is
	\begin{equation}\label{eq:reinforce_emp_grad}
		\widehat{\nabla_{\theta} J(\theta)} = \frac{1}{N} \sum_{n=1}^{N}{\sum_{t=0}^{\infty}{\hat{A}_{t}^{n} \nabla_{\theta} \log{\pi(a_t^n | s_t^n})   }}
	\end{equation}
	$A^{\pi}(s, a)$ is, like $Q^{\pi}(s, a)$ and $V^{\pi}(s)$, unknown, and must thus be estimated by the function approximator $\hat{A}(s)$. Generalized Advantage Estimation (GAE), as proposed in \cite{schulmanhighdimcontrol}, is the most notable approach. GAE makes use of a function approximator (commonly a neural network) $\hat{V}(s)$ to approximate the actual value function $V(s)$. A common approach is to use an artificial neural network, which is trained on the discounted empirical returns.
	\subsubsection{Proximal policy optimization}
	
	PPO, as well as its predecessor (Trust Region Policy Optimization \cite{SchulmanLMJA15_TRPO}) do not, even though it is feasible, optimize the policy directly via the expression in Equation \ref{eq:reinforce_emp_grad}. TRPO instead optimizes the surrogate objective function
	\begin{equation}
		J^{CPI}(\theta') = \hat{\mathbb{E}}_t {\left[ \frac{\pi_{\theta'}{(a_t | s_t)}}{\pi_{\theta}{(a_t | s_t)}} \hat{A}_t^{\pi_{\theta}}  \right]}
	\end{equation}
	which provides theoretical guarantees for policy improvement. However, as this relies on an approximation that is valid only in the local neighborhood, carefully choosing the step size is critical to avoid instabilities. Unlike in TRPO, where this is achieved by imposing a hard constraint on the relative entropy between the current and next policy, PPO elegantly incorporates the preference for a modest step-size in the optimization target, yielding a more efficient algorithm \cite{schulman2017proximal}. Specifically, it instead focuses on maximizing
	\begin{equation}
		\begin{aligned}
			&J^{CLIP}(\theta^{\prime}) = \hat{\mathbb{E}}_t {\left[\min{\left( r_t(\theta) \hat{A}_t^{\pi_{\theta}}, \clip_\epsilon{(r_t(\theta))} \hat{A}_t^{\pi_{\theta}} \right)}  \right]}\\
			&\clip_\epsilon(x) = \clip{(x, 1 - \epsilon, 1 + \epsilon)}
		\end{aligned}
	\end{equation}
	\noindent where $r_t(\theta)$ is a shorthand for the probability ratio $\frac{\pi_{\theta'}{(a_t | s_t)}}{\pi_{\theta}{(a_t | s_t)}}$.
	
	The PPO training process, which is written in pseudocode format in Algorithm \ref{alg:ppo}, can then be summarized as follows: At each iteration, PPO first collects batches of Markov trajectories from concurrent rollouts of the current policy. Next, the policy is updated according to a stochastic gradient descent update scheme.
	\begin{algorithm}[!htb]
		\scriptsize
		\begin{algorithmic}
			\For{iteration = $1, 2, ...$}
			\For{actor = $1, 2, ... N$}
			\State For $T$ time-steps, execute policy $\pi_{\theta}$.
			\State Compute advantage estimates $\hat{A}_1, \dots \hat{A}_T$
			\EndFor
			\For{epoch = $1, 2, ... N_E$}
			\State Obtain mini batch of $N_{MB}$ samples from the $N_A T$ simulated time-steps.
			\State Perform SGD update from minibatch $(\mathbf{X}_{MB}, \mathbf{Y}_{MB})$.
			\State $\theta \gets \theta'$
			\EndFor
			\EndFor
		\end{algorithmic}
		\caption{Proximal Policy Optimisation}
		\label{alg:ppo}
	\end{algorithm}
	\subsection{Terrain data}
	\begin{figure}[ht!]
		\centering
		\includegraphics[width=0.7\linewidth]{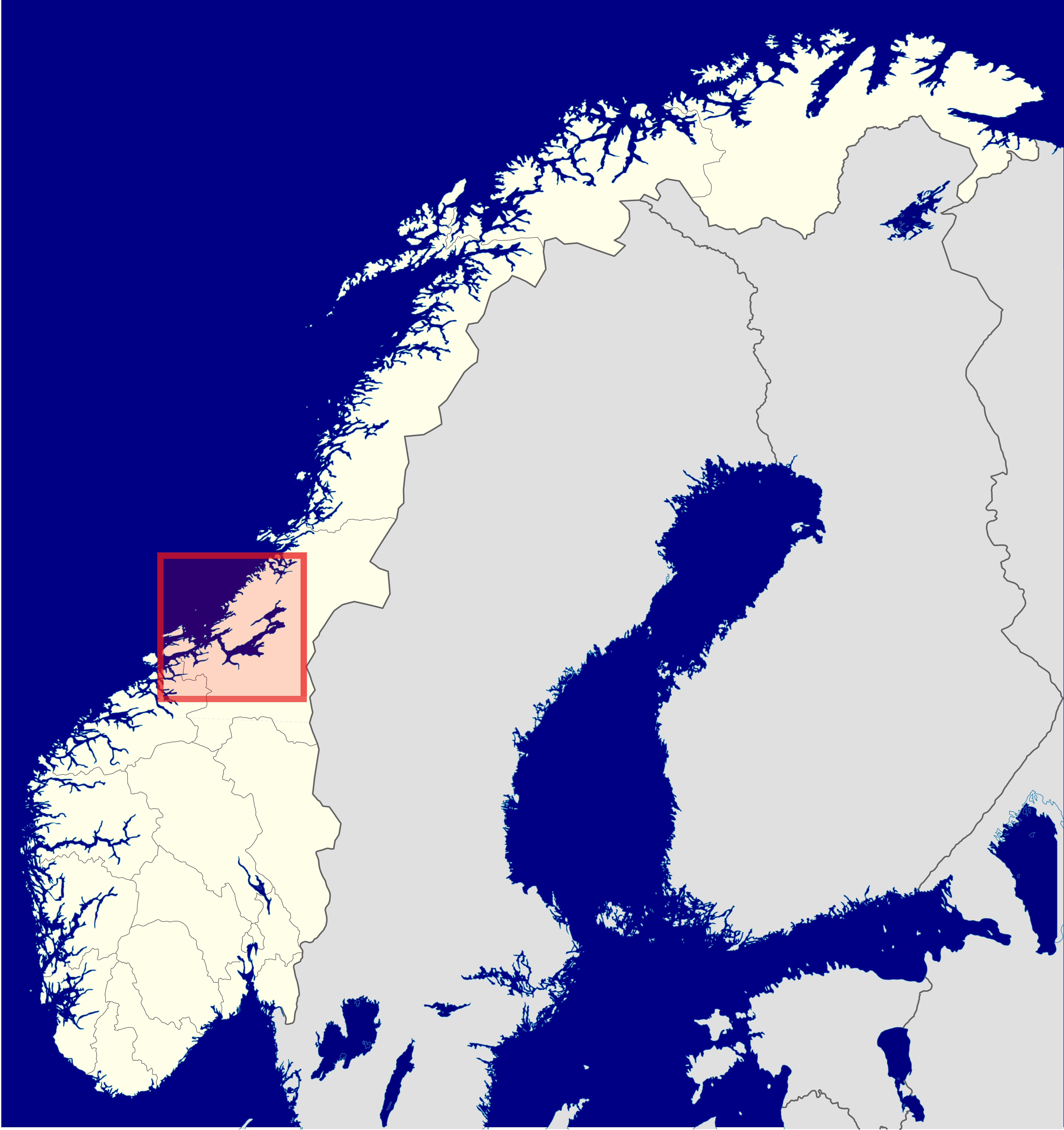}
		\caption[Area of interest]{Map of the Norwegian mainland highlighting the area of interest.\footnotemark}
		\label{fig:trondheimfjordmap}
	\end{figure}
	Our maritime simulation environment is made from a digital reconstruction of the Trondheim Fjord (Figure \ref{fig:trondheimfjordmap}), an inlet of the Norwegian sea. Specifically, it is based on a digital terrain model (DTM) provided by the Norwegian Mapping Authority (Kartverket). The data set, which is called DTM10, is generated from airborne laser scanning, and has a horizontal resolution of 10x10 meters with coverage of the entire Norwegian mainland \cite{kartverket_2019}. The coordinates are given according to the Universal Transverse Mercator (UTM) rectangular projection system, which partitions the Earth into 60 north-south zones, each of which has a 6 degree longitudinal span. Within each zone, which is indexed consecutively from zone 1 (180°W to 174°W) to zone 60 (174°E to 180°E), a mapping from latitude/longitude coordinates to a Cartesian x-y coordinate system is performed based on a local flat earth-assumption. Given the vast number of zones used in the UTM projection system, the approximated coordinates, which of course have inherent distortions because of the spherical shape of the Earth, are of relatively high accuracy. The DTM10 data set is given with respect to zone 33. 
	\footnotetext{Original image source: NordNordWest (\url{https://commons.wikimedia.org/wiki/File:Norway_location_map.svg}), "Norway location map"}
	\begin{figure}[htb!]
		\includegraphics[trim={1cm 0cm 3cm 0cm},clip,width=\linewidth]{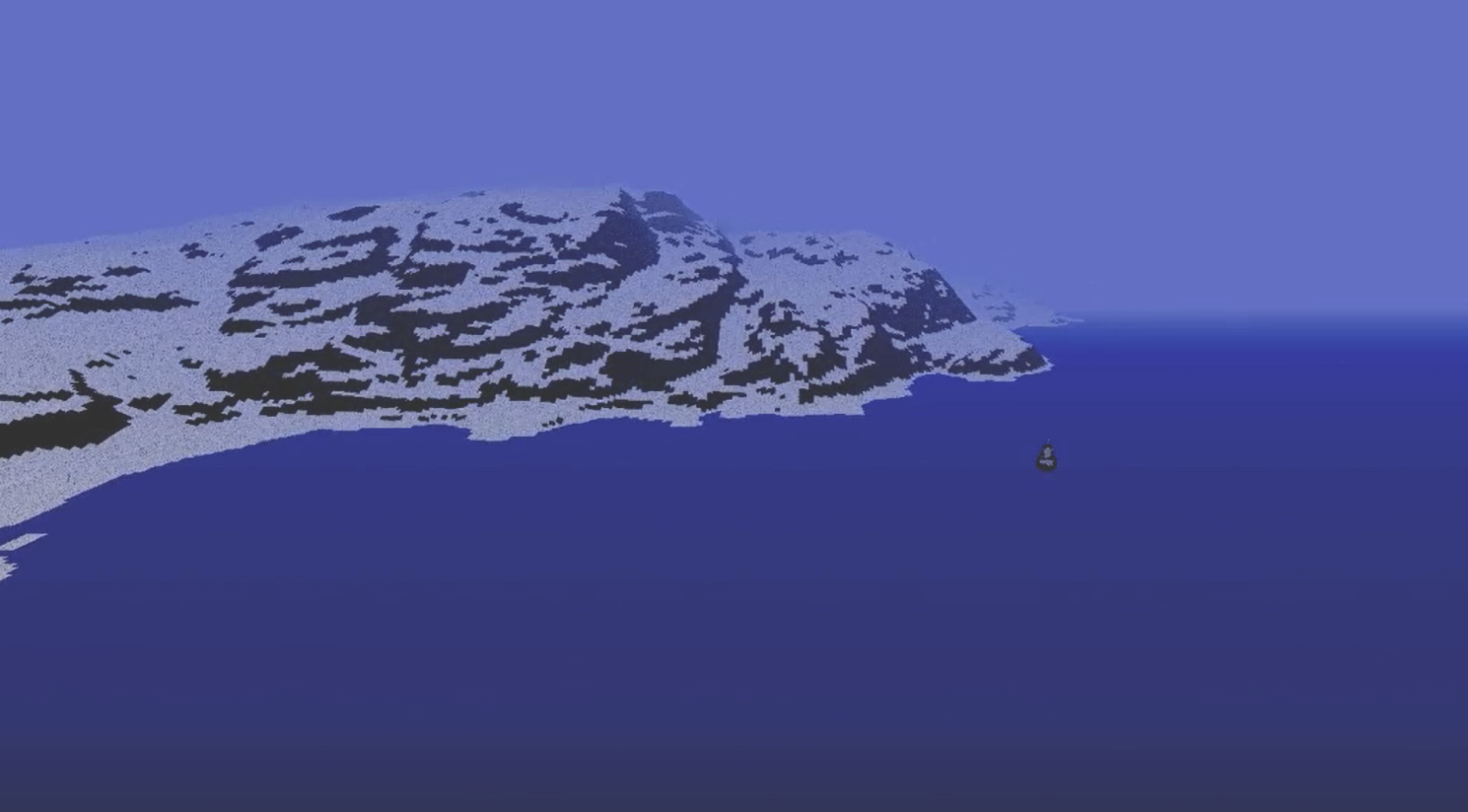}
		\caption[Digital terrain reconstruction]{Digital terrain reconstructed from DTM10 (Norwegian Mapping Authority) rendered in 3D for debugging and showcasing purposes. Specifically, this shows a view of the Bymarka area, a nature reserve on the west side of Trondheim.}
		\label{fig:3dterrain}
	\end{figure}
	\subsection{Tracking data}
	
\begin{figure}[ht!]
		\hspace*{-0.70cm}\includegraphics[trim={0.9cm 0cm 0cm 0cm},clip,width=0.62\textwidth]{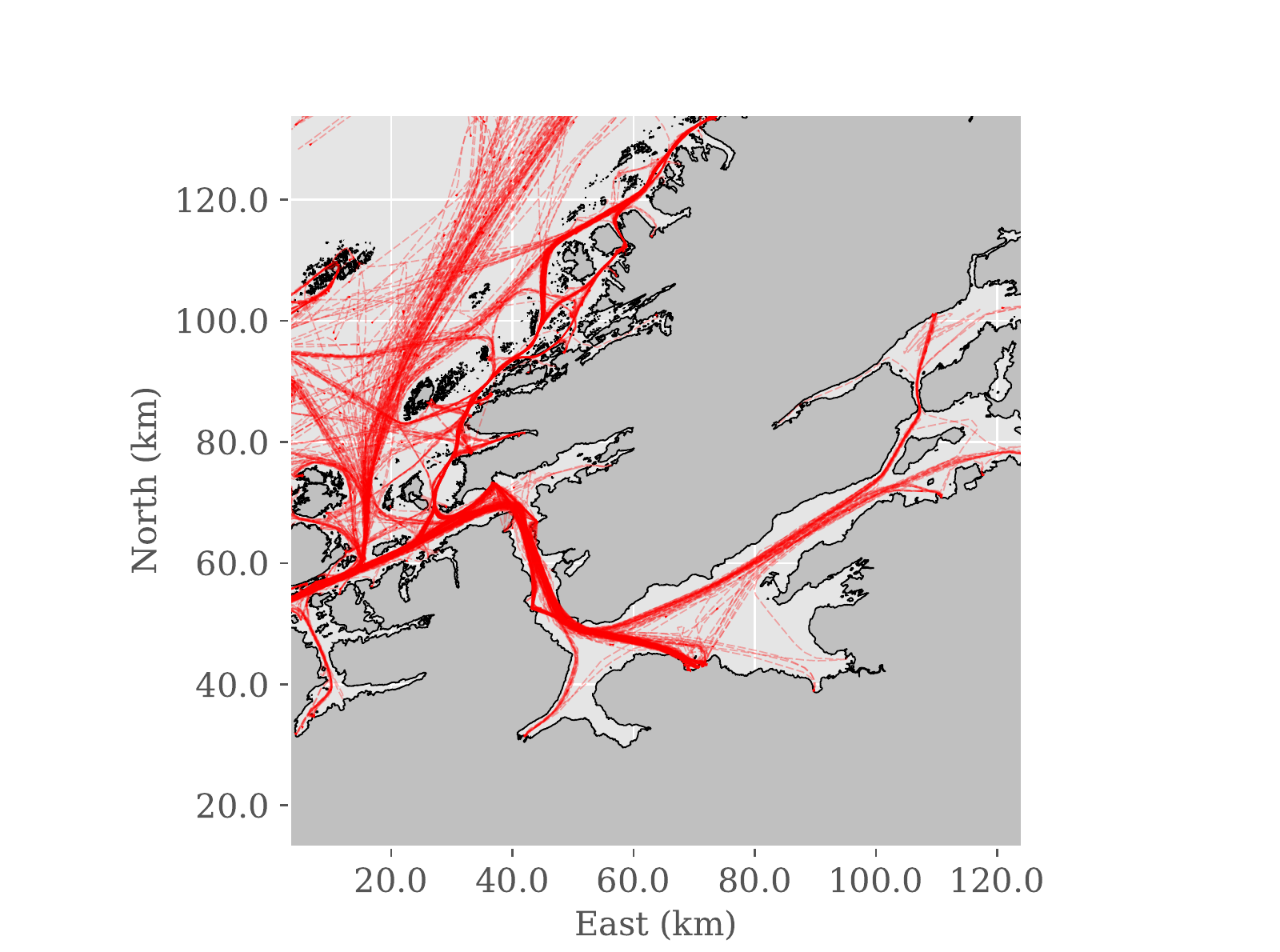}
		\caption[Traffic in the Trondheim Fjord area]{Snapshot of the marine traffic from January 26, 2020 to February 6, 2020 in the Trondheim Fjord area based on AIS tracking data. Each line represents one recorded travel.}
		\label{fig:marinetraffic}
	\end{figure}

	We obtain a sample of historical vessel tracking data in the Trondheim Fjord area from a query of the Norwegian Coastal Administration's AIS Norway data service. The automatic identification system (AIS) is an automatic tracking system which provides both static (e.g. vessel dimensions) and dynamic (e.g. vessel position, heading and speed) information based on vessel transmissions. Within the field of autonomous surface vehicle guidance, AIS information is often used as a supplementary data source that is, by method of sensor fusion, combined with marine radar in collision avoidance algorithms. Additionally, given a large enough sample time within the area of interest, it provides a historical model of the marine traffic in the area. In our case, our historical data results from a 10 day data query ranging from January 26, 2020 to February 6, 2020 of all recorded traffic (Figure \ref{fig:marinetraffic}) within a rectangular area around the Trondheim Fjord. 
	Depending on the transmitter characteristics for each individual vessel, the resulting tracking data resolution varies from 2-20 seconds, facilitating a high-accuracy reconstruction of each vessel's trajectory in our simulation. As the AIS tracking data represents vessel position by latitude/longitude coordinates, a conversion to the zone 33 UTM x-y coordinate system is called for. To do the conversion, we utilize the \textit{from\_latlon} method provided by the Python package \textbf{utm} \cite{UTM}.
	
	\section{Methodology}
	\label{section:methodology}
	\subsection{Training environment}
	
\begin{figure}[ht!]
		\hspace*{-1cm}\includegraphics[trim={0cm 0cm 0cm 0.5cm},clip,width=0.6\textwidth]{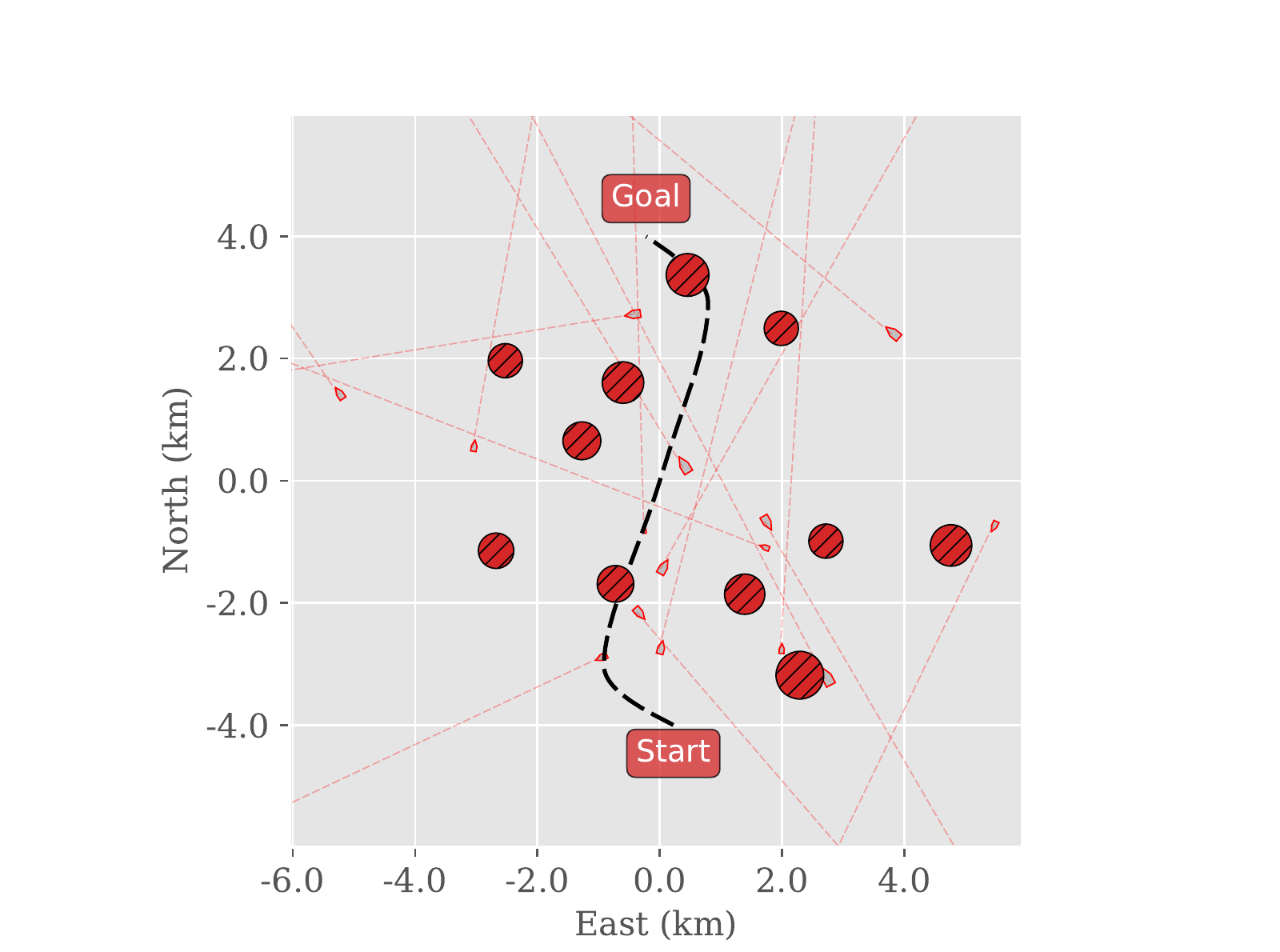}
		\caption[Random moving obstacles training scenario]{Random sample of the stochastically generated path following training scenario with moving obstacles. The circles are static obstacles, whereas the vessel-shaped objects are moving according to the trajectory lines.}
		\label{fig:movingobstacles_demo}
	\end{figure}

	DRL-based autonomous agents have a remarkable ability to generalize their policy over the observation space, including the domain of unseen observations. And given the complexity and heterogeneity of the Trondheim Fjord environment, with archipelagos, shorelines and skerries (see Figure \ref{fig:marinetraffic}), this ability will be fundamental to the agent's performance. However, the training environment, in which the agent is supposed to evolve from a blank slate to an intelligent vessel controller, must be both representative, challenging and unpredictable to facilitate the generalization. Of course, the most representative choice for a training scenario would be the Trondheim Fjord itself, which would, if it was not for the generalization issues associated with this approach \cite{codevilla2019exploring}, also allow for training the agent via behavior cloning based on the available vessel tracking data. However, given the resolution of our terrain data, the resulting obstacle geometry is typically very complex, leading to overly high computational demands for simulating the functioning of the distance sensor suite. Thus, the better choice is to carefully craft an artificial training scenario with simple obstacle geometries. To reflect the dynamics of a real marine environment, we let the stochastic initialization method of the training scenario spawn other target vessels with deterministic, linear trajectories. Additionally, circular obstacles, which are scattered around the environment, are used as a substitute for real-world terrain. A randomly chosen initialization of the training environment is shown in Figure \ref{fig:movingobstacles_demo}.
	\subsection{Observation vector}
	
	Here, the goal is to engineer an observation vector $s$ containing sufficient information about the vessel's state relative to the path, as well as information from the sensors. To achieve this, the full observation vector is constructed by concatenating navigation-based and perception-based features, which formally translates to $s = [s_{n}, s_{p}]^T$. In the context of this paper, we consider the term \textit{navigation} as the characterization of the vessel's state, i.e. its position, orientation and velocity, with respect to the desired path. On the other hand, \textit{perception} refers to the observations made via the rangefinder sensor measurements. In the following, the path navigation feature vector $s_{n}$ and the elements culminating in the perception-based feature vector $s_{p}$ are covered in detail.
	
	\subsubsection{Path navigation}
	
	A sufficiently information-rich path navigation feature vector would be such that it, on its own, could facilitate a satisfactory path-following controller (without any consideration for obstacle avoidance). A few concepts often used in the field of vessel guidance and control are useful in order to formalize this. 
	\begin{figure}[ht!]
		\includegraphics[width=0.9\linewidth]{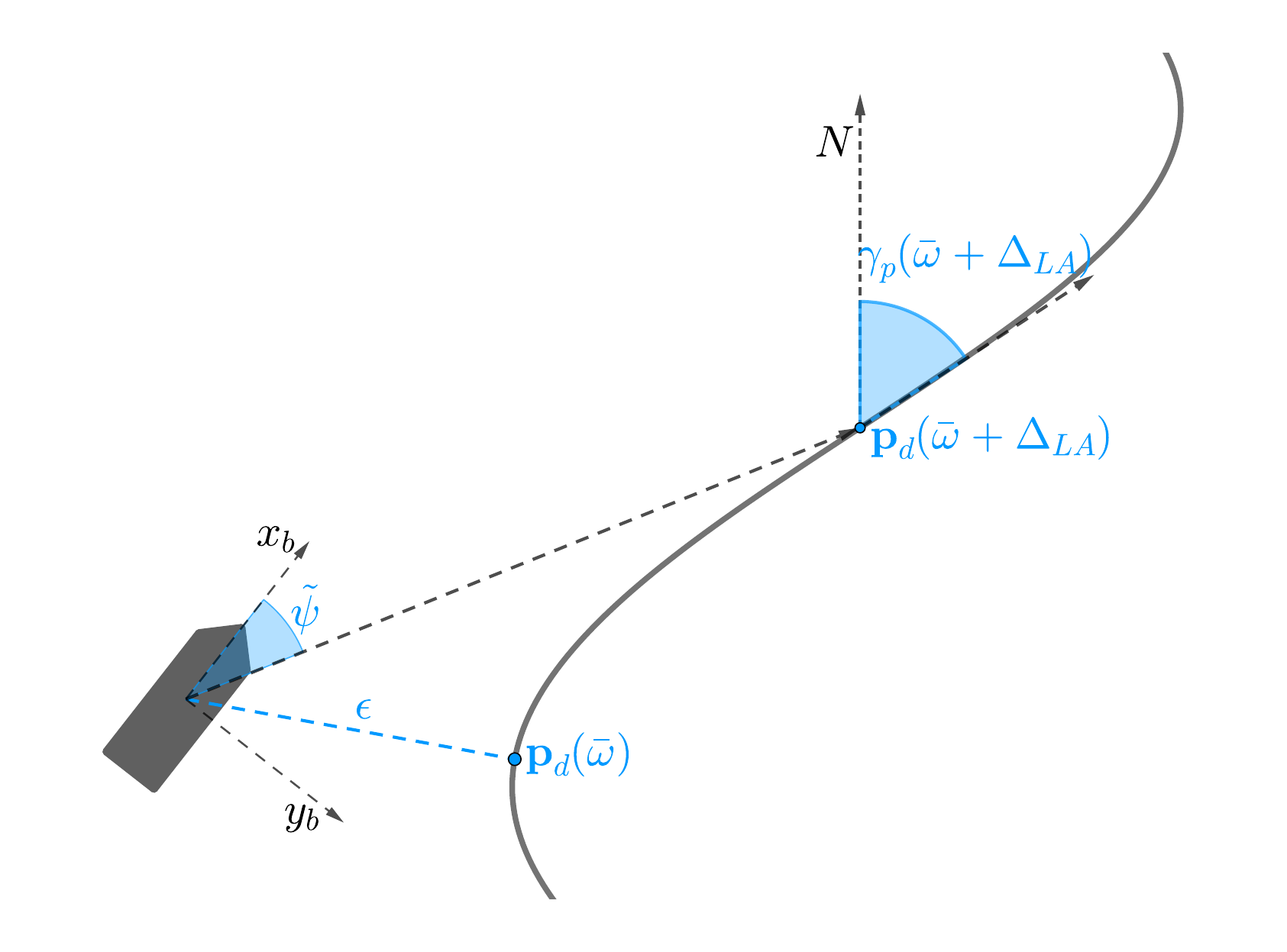}
		\caption[Path following navigation]{Illustration of key concepts for navigation with respect to path following. The path reference point $\bm{p}_d(\omega)$, i.e. point yielding the closest Euclidean distance to the vessel, is here located right of the vessel, while the look-ahead reference point $\bm{p}_d (\bar{\omega} + \Delta_{LA})$ is located a distance $\Delta_{LA}$ further along the path.}
		\label{fig:navigationfig}
	\end{figure}
	First, we introduce the mathematical representation of the parameterized path, which is expressed as
	\begin{equation}
		\bm{p}_d(\omega) = \left[x_d(\omega), y_d(\omega)\right]^T
	\end{equation}
	\noindent where $x_d(\omega)$ and $y_d(\omega)$ are given in the NED-frame. Navigation with respect to the path necessitates a reference point on the path which is continuously updated based on the current vessel position. Even though other approaches exist, this reference point is best thought of as the point on the path that has the closest Euclidean distance to the vessel, given its current position, as visualised in the example illustration shown in Figure \ref{fig:navigationfig}. To find this, we calculate the corresponding value of the path variable $\bar{\omega}$ at each time-step. This is an equivalent problem formulation because the path is defined implicitly by the value of $\omega$. Formally, this translates to the optimization problem
	\begin{equation}
		\bar{\omega} =  \argmin_\omega \left(x^n - x_d(\omega)\right)^2 + \left(y^n - y_d(\omega)\right)^2
	\end{equation}
	Which, using the Newton–Raphson method, can be calculated accurately and efficiently at each time-step. Here, the fact that the Newton–Raphson method only guarantees a local optimum is a useful feature, as it prevents sudden path variable jumps given that the previous path variable value is used as the initial guess \cite{martinsen2018}.
	
	Accordingly, we define the corresponding Euclidean distance to the path, i.e. the deviation between the desired path and the current track, as the cross-track error (CTE) $\epsilon$. Formally, we thus have that
	\begin{equation}
		\epsilon =  \left\lVert \left[ x^n, y^n \right]^T - \bm{p}_d(\bar{\omega}) \right\rVert
	\end{equation}
	Next, we consider the look-ahead point $\bm{p}_d (\bar{\omega} + \Delta_{LA})$ to be the point which lies a constant distance further along the path from the reference point $\bm{p}_d(\bar{\omega})$. The parameter $\Delta_{LA}$, the look-ahead distance, is set by the user and controls how aggressively the vessel should reduce the distance to the path. Look-ahead based steering, i.e. setting the look-ahead point direction as the desired course angle, is a commonly used guidance principle \cite{fossen11}.
	
	We then define the heading error $\tilde{\psi}$ as the change in heading needed for the vessel to navigate straight towards the look-ahead point from its current position, as illustrated in Figure \ref{fig:navigationfig}. This is calculated from
	\begin{equation}
		\tilde{\psi} = \atantwo{\left(\frac{y_d(\bar{\omega} + \Delta_{LA}) - y^n}{x_d(\bar{\omega} + \Delta_{LA}) - x^n}\right)} - \psi
	\end{equation}
	\noindent where $\psi$ is the vessel's current heading and $x^n, y^n$ are the current NED-frame vessel coordinates as defined earlier.
	
	However, even if minimizing the heading error will yield good path adherence, taking into account the path direction at the look-ahead point might improve the smoothness of the resulting vessel trajectory. Referring to the first order path derivatives as $x_{p}^{\prime}(\bar{\omega})$ and $y_{p}^{\prime}(\bar{\omega})$, we have that the path angle $\gamma_p$, in general, can be expressed as a function of arc-length $\omega$ such that
	\begin{equation}
		\gamma_{p}(\bar{\omega}) = \atantwo{(y_{p}^{\prime}(\bar{\omega}), x_d^{\prime}(\bar{\omega}))}
	\end{equation}
	As visualized in Figure \ref{fig:navigationfig}, the path direction at the look-ahead point is then given by $\gamma_{p}(\bar{\omega} + \Delta_{LA})$. Accordingly, we can then define the look-ahead heading error, which is zero in the case when the vessel is heading in a direction that is parallel to the path direction at the look-ahead point, as
	\begin{equation}
		\tilde{\psi}_{LA} = \gamma_{p}(\bar{\omega} + \Delta_{LA}) - \psi
	\end{equation}
	Our assumption is then that the navigation feature vector $s_{n}$, defined as outlined in Table \ref{tab:obs_vector_path}, should provide a sufficient basis for the agent to intelligently adhere to the desired path.
	\begin{table}[h]
		\scriptsize
		\begin{tabular}{ll}
			\hline
			\textbf{Feature} & \textbf{Definition}\\
			\hline
			Surge velocity & $u^{(t)}$ \\
			Sway velocity & $v^{(t)}$ \\
			Yaw rate & $r^{(t)}$ \\
			Cross-track error & $\epsilon^{(t)}$ \\
			Heading error & $\tilde{\psi}^{(t)}$ \\
			Look-ahead heading error & $\tilde{\psi}_{LA}^{(t)}$ \\
			\hline
		\end{tabular}
		\caption[Path-following feature vector]{Path-following feature vector $s_{n}$ at timestep $t$.}\label{tab:obs_vector_path}
	\end{table}
	Formally, we thus have that
	\begin{equation}
		s_{n}^{(t)} = \left[u^{(t)}, v^{(t)}, r^{(t)}, \epsilon^{(t)}, \tilde{\psi}^{(t)}, \tilde{\psi}_{LA}^{(t)}\right]^T
	\end{equation}
	\FloatBarrier
	
	\subsubsection{Sensing}
	Using a set of rangefinder sensors as the basis for obstacle avoidance is a natural choice, as it yields a comprehensive, easily interpretable representation of the neighbouring obstacle environment. This should also enable a relatively straightforward transition from the simulated environment to a real-world one, given the availability of common rangefinder sensors, be it lidars, radars, sonars or depth cameras.
	\begin{figure}[ht!]
		\hspace*{0.4cm}\includegraphics[trim={0cm 0cm 0.0cm 0.0cm},clip,width=0.9\linewidth]{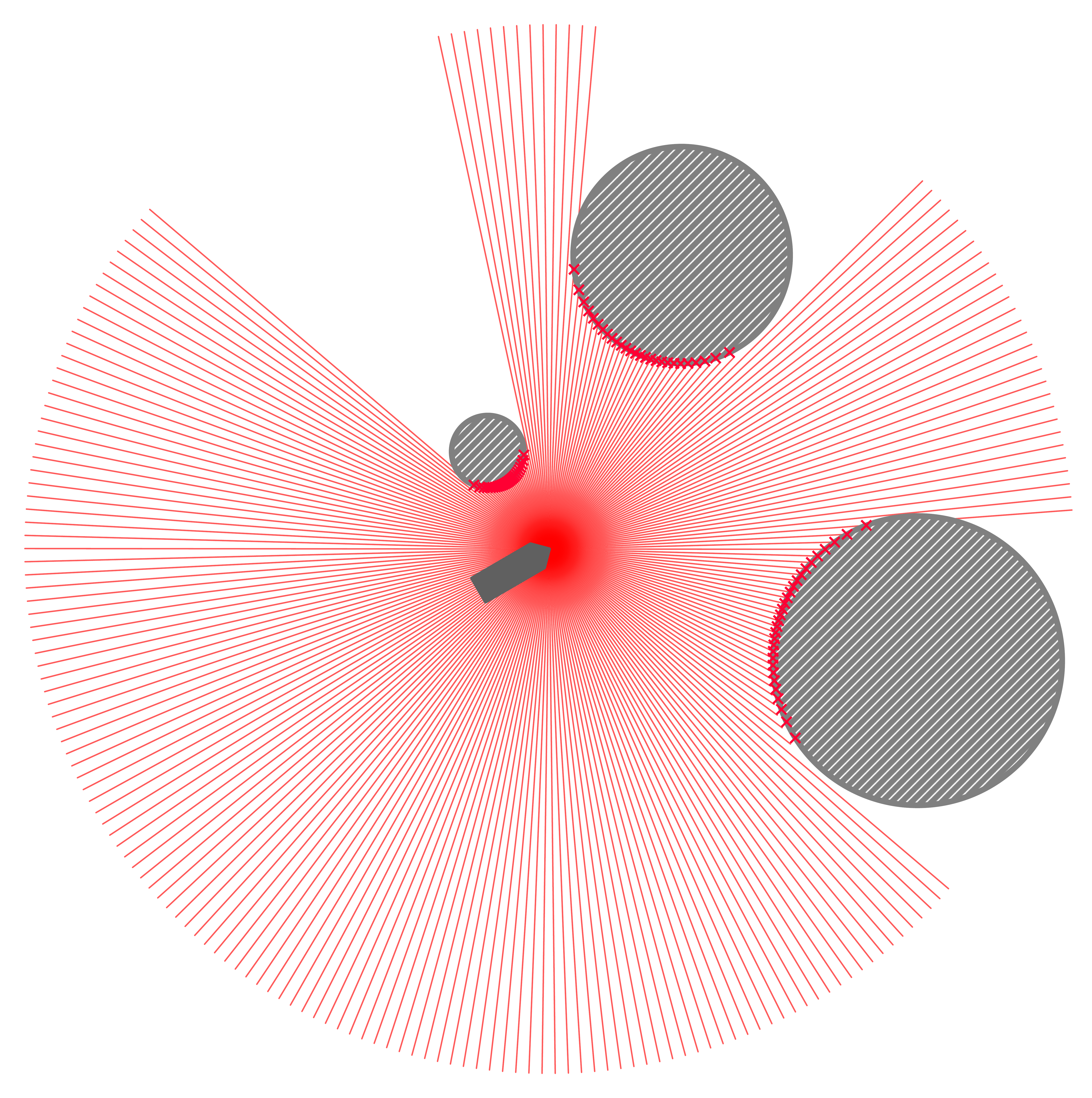}
		\caption[Rangefinder sensor suite]{Rangefinder sensor suite attached to autonomous surface vessel.}
		\label{fig:sensorswithoutsectors}
	\end{figure} 
	In our setup, the vessel is equipped with $N$ distance sensors with a maximum detection range of $S_r$, which are distributed uniformly with 360 degree coverage, as illustrated in Figure \ref{fig:sensorswithoutsectors}. While the area behind the vessel is obviously of lesser importance, and not necessary to consider for navigating purely static terrain \cite{meyer_ASV_IEEE}, the possibility of overtaking situations where the agent must react to another vessel approaching from behind makes full sensor coverage a necessity.
	
	\subsubsection{Sensor partitioning}
	The most natural approach to constructing the final observation vector would then be to concatenate the path information feature vector with the array of sensor outputs. However, initial experiments with this approach were aborted as it became apparent that the training process had stagnated - at a very dissatisfactory agent performance level. A likely explanation for this failure is the size of the observation vector which was fed to the agent's policy and value networks; as it becomes overly large, the agent suffers from the well-known \textit{curse of dimensionality}. Due to the resulting network complexity, as well as the exponential relationship between the dimensionality and volume of the observation space, the agent fails to generalize new, unseen observations in an intelligent manner \cite{goodfellowDL}. This calls for a significant dimensionality reduction. This can, of course, be achieved simply by reducing the number of sensors, something which would also have the fortunate side effect of reducing the simulation's computational needs. Unfortunately, this approach also turned out unsuccessful, even after testing a wide range of smaller sensor setups. Clearly, when the sensor count becomes too low, the agent's perception of the neighboring obstacle environment is simply too scattered to facilitate satisfactory obstacle-avoiding behavior in challenging scenarios such as the ones used for training the agent. As balancing the trade-off between sensor resolution and observation dimensionality appears intractable, this calls for a more involved approach.
	
\begin{figure}[ht!]
		\hspace*{0.4cm}\includegraphics[trim={0cm 0cm 0.0cm 0.0cm},clip,width=0.9\linewidth]{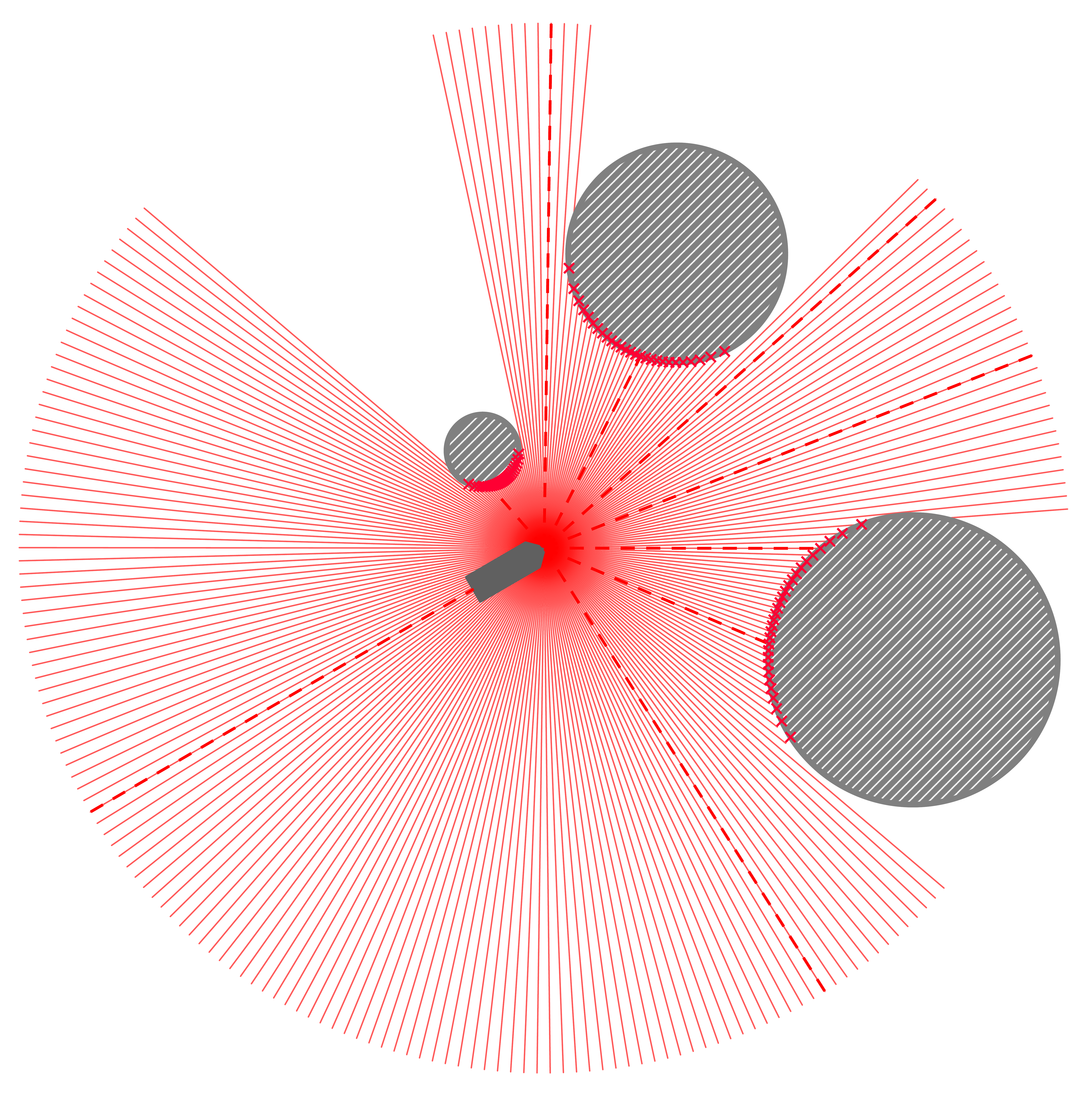}
		\caption[Sector-partitioned rangefinder sensor suite]{Rangefinder sensor suite partitioned into $D=9$ sectors according to the the mapping function $\kappa$ with the scale parameter $\gamma_C = 0.13$.}
		\label{fig:sensorswithsectors}
	\end{figure}

	A natural approach is to partition the sensor suite into $D$ sectors, each of which produces a scalar measurement which is included in the final observation vector, effectively summarizing the local sensor readings within the sector. However, given our desire to minimize its dimensionality, dividing the sensors into sectors of uniform size is likely sub-optimal, as obstacles located in front of the vessel are significantly more critical and thus require a higher degree of perception accuracy than those that are located at its rear. In order to realize such a non-uniform partitioning, we use a logistic function - a choice that also fulfills our general preference for symmetry. Assuming a counter-clockwise ordering of sensors and sectors starting at the rear of the vessel, we map a given sensor index $i \in \integerset{N}$ to sector index $k \in \integerset{D}$ according to
	\begin{equation}
		\kappa : i \mapsto \kappa(i) = \floor*{\underbrace{D \sigma\left(\frac{\gamma_C i}{N} - \frac{\gamma_C}{2} \right)}_\text{Non-linear mapping} - \underbrace{D \sigma \left( - \frac{\gamma_C}{2} \right)}_\text{Constant offset}}
	\end{equation}
	where $\sigma$ is the logistic sigmoid function and $\gamma_C$ is a scaling parameter controlling the density of the sector distribution such that decreasing it will yield a more evenly distributed partitioning. In Figure \ref{fig:sensorswithsectors}, the practical output of this sensor mapping procedure is visualised, with the sectors being the narrowest near the front of the vessel.
	
	We can then formally define the distance measurement vector for the $k^{th}$ sector, which we denote by $\bm{w}_k$, according to
	\begin{align*}
		\bm{w}_{k, i} &=x_i&     & \text{ for } i \in \integerset{N} \text{ such that } \kappa(i) = k
	\end{align*}
	\begin{figure}[ht!]
		\centering
		\begin{subfigure}{0.49\linewidth}
			{\includegraphics[width=\linewidth]{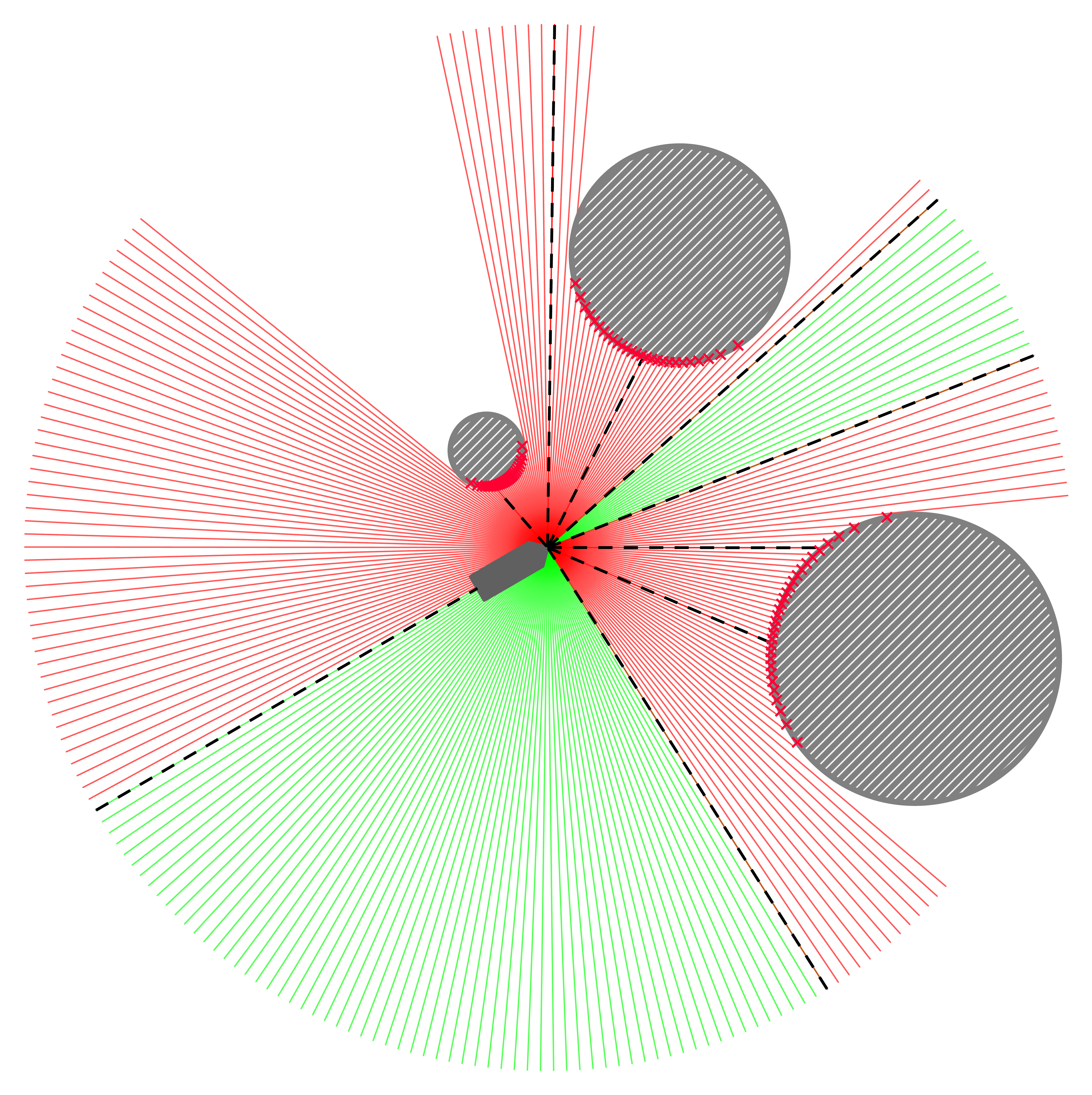}}
			\subcaption{Min pooling}
			\label{fig:minpooling}
		\end{subfigure}
		\begin{subfigure}{0.49\linewidth}
			{\includegraphics[width=\linewidth]{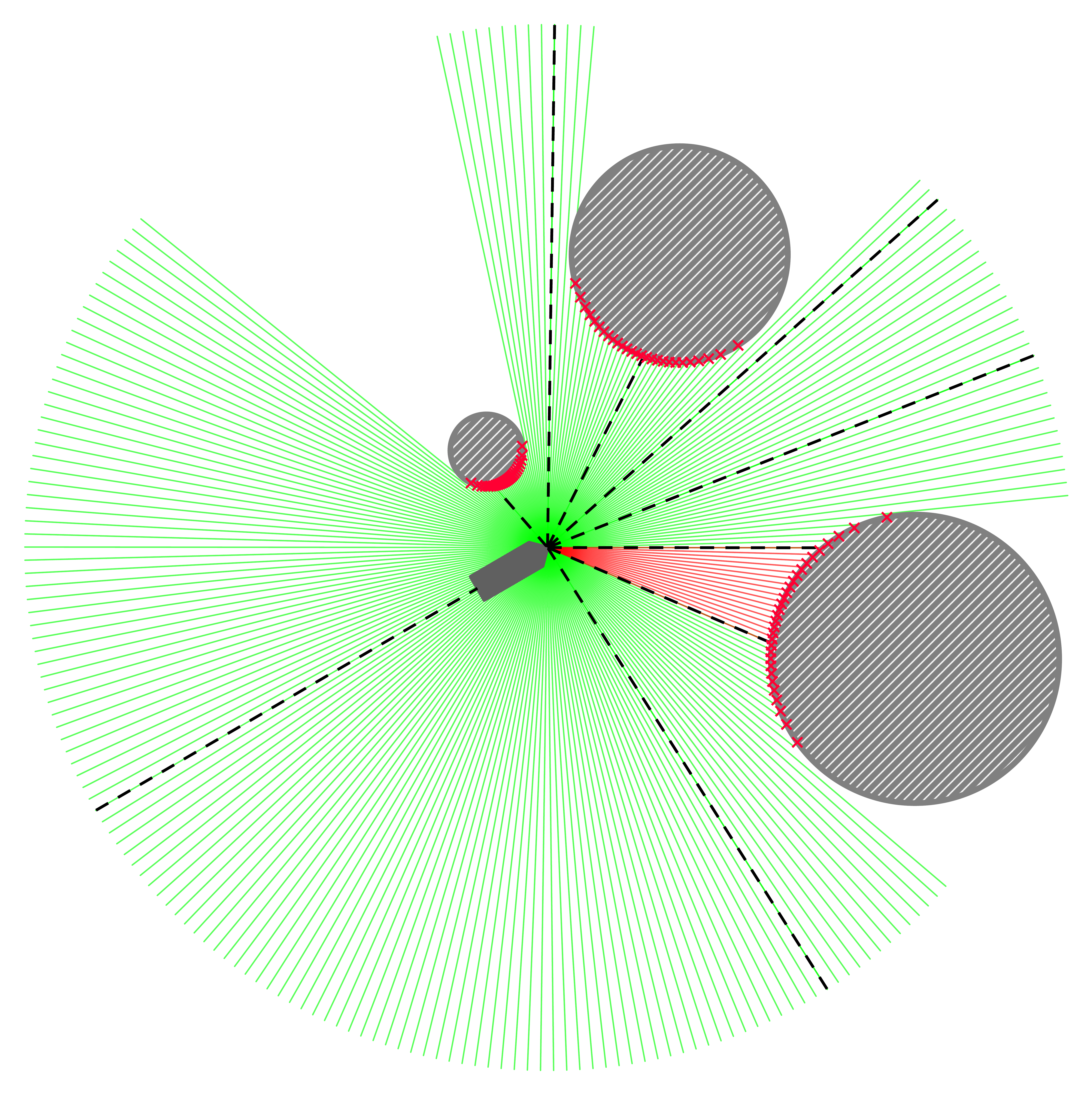}}
			\subcaption{Max pooling}
			\label{fig:maxpooling}
		\end{subfigure}
		\begin{subfigure}{0.49\linewidth}
			{\includegraphics[width=\linewidth]{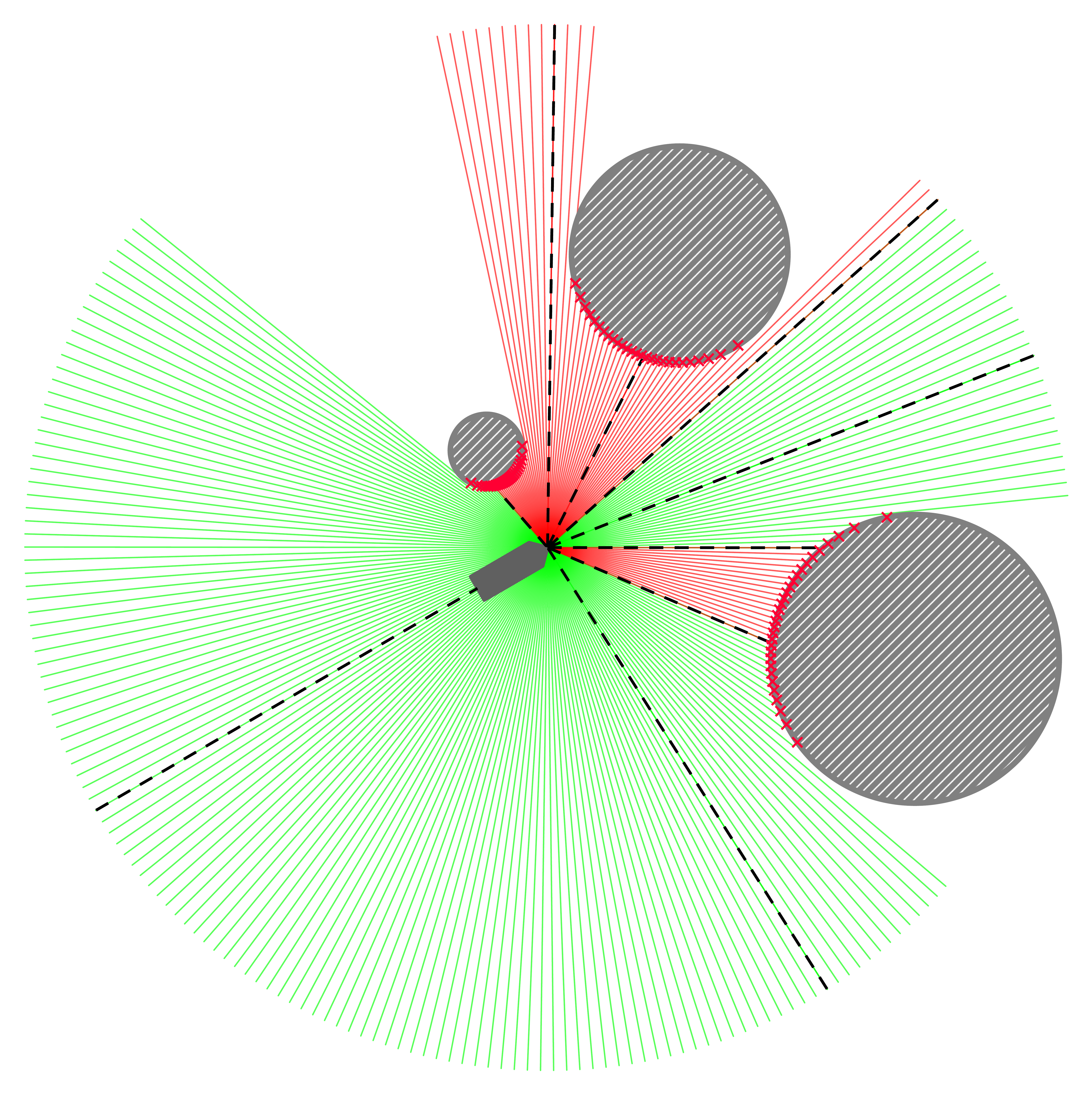}}
			\subcaption{Feasibility pooling}
		\end{subfigure}
		\caption{Pooling techniques for sensor dimensionality reduction. For the sectors colored green, the maximum distance $S_r$ was outputted, implying that the sector is clear of any obstacles. It is obvious that min-pooling yields an overly restrictive observation vector, effectively telling the agent that a majority of the travel directions are blocked. On the other hand, max pooling yields overly optimistic estimates, potentially leading to dangerous situations. The feasibility pooling algorithm, however, mirrors an intuitive reasoning about the reachability within each sector, producing a more intelligent estimate.}
	\end{figure}
	Next, we seek a mapping $f : \mathbb{R}^n \mapsto \mathbb{R}$, which takes the vector of distance measurements $\bm{w}_k$, for an arbitrary sector index $k$, as input, and outputs a scalar value based on the current sensor readings within the sector. Always returning the smallest measured obstacle distance within the sector, i.e. $f = \min$ (in the following referred to as \textit{min pooling}), is a natural approach which yields a conservative and thereby safe observation vector. As can be seen in Figure \ref{fig:minpooling}, however, this approach might be overly restrictive in certain obstacle scenarios, where feasible openings in between obstacles are inappropriately overlooked. However, even if the opposite approach (\textit{max pooling}, i.e. $f = \max$) solves this problem, it is straight-forward to see, e.g. in Figure \ref{fig:maxpooling} by considering the fact that the presence of the small obstacle near the vessel is ignored, that it might lead to dangerous navigation strategies.
	\begin{figure}[ht!]
		\centering
		\begin{subfigure}{0.49\linewidth}
			{\includegraphics[width=\linewidth]{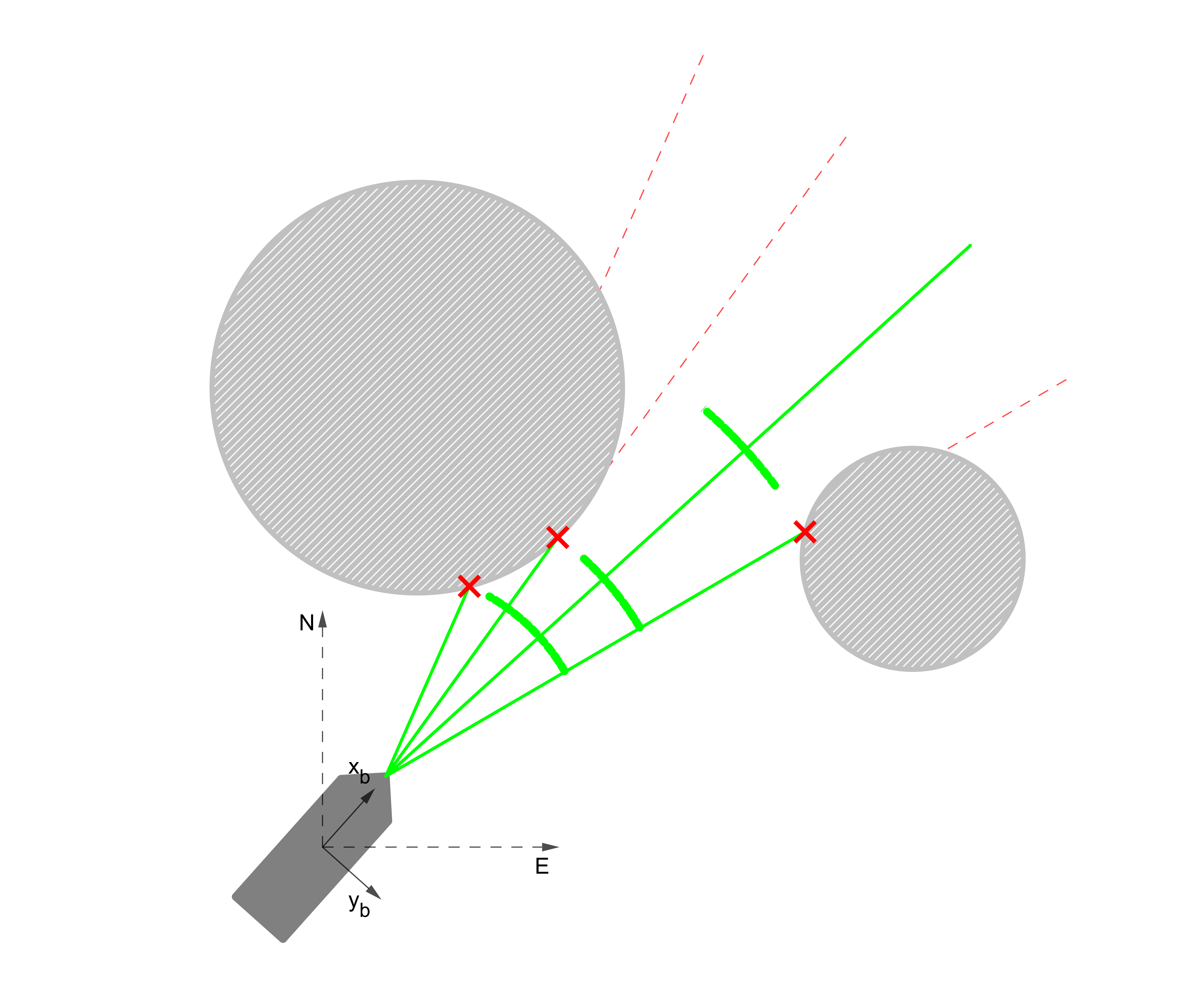}}
			\subcaption{Full distance is reachable.}
			\label{fig:feasible_passing}
		\end{subfigure}
		\begin{subfigure}{0.49\linewidth}
			{\includegraphics[width=\linewidth]{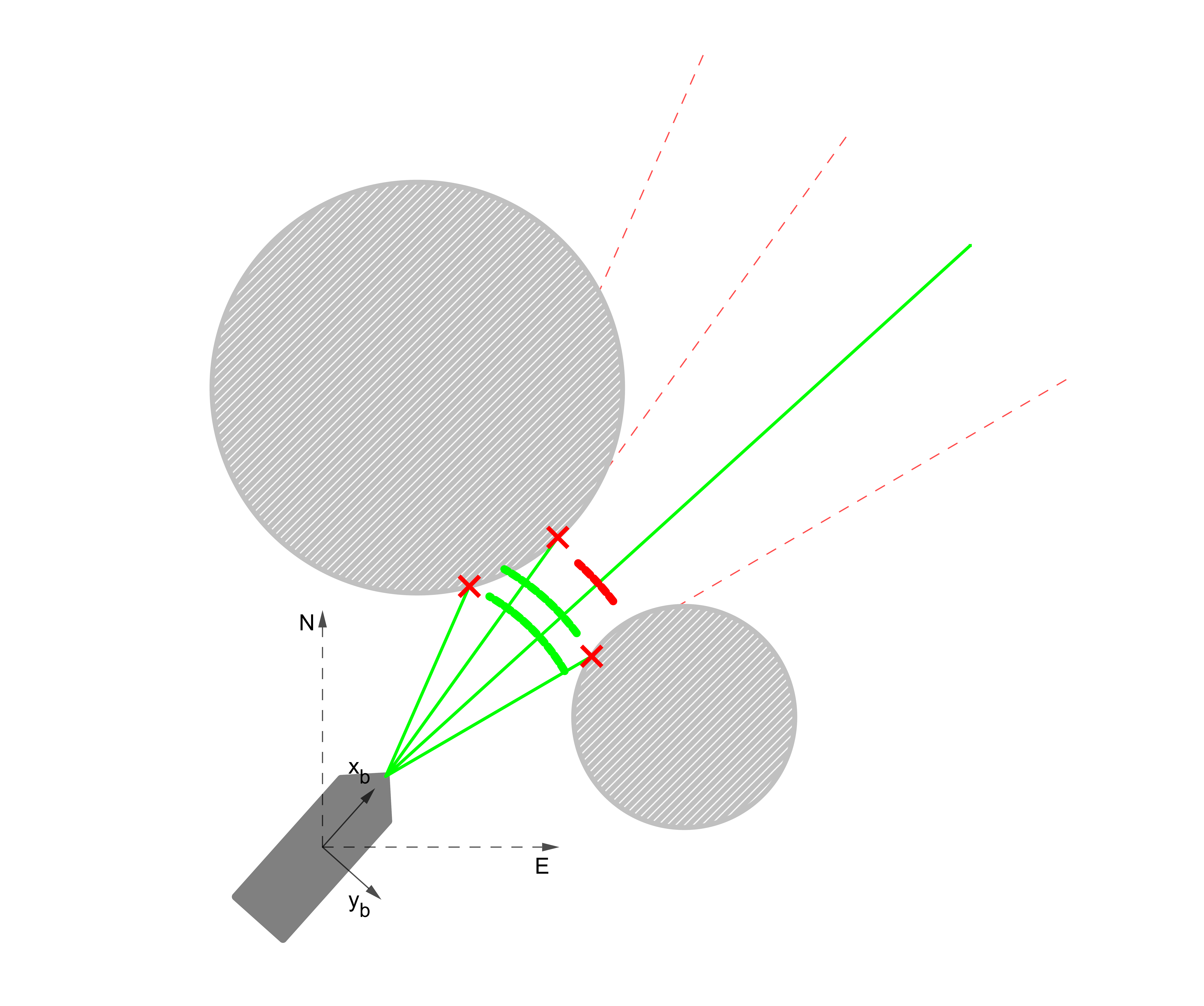}}
			\subcaption{Less than half the distance is reachable.}
			\label{fig:infeasible_passing}
		\end{subfigure}
		\caption[Feasibility pooling example]{Illustration of the feasibility algorithm for two different scenarios. After sorting the sensor indices according to the corresponding distance measurements, the algorithm iterates over them in ascending order, and, at each step, decides if the vessel can feasibly continue past this point. In the scenario displayed in the figure on the right, the opening is deemed too narrow for the full distance to be reachable.}
	\end{figure}
	In order to alleviate the problems associated with min and max pooling mentioned above, a new approach is required. The \textit{feasibility pooling} procedure, which was introduced in \cite{meyer_ASV_IEEE}, calculates the maximum reachable distance within each sector, taking into account the location of the obstacle sensor readings as well as the width of the vessel. This method requires us to iterate over the sensor reading in ascending order corresponding to the distance measurements, and for each resulting distance level check whether it is feasible for the vessel to advance beyond this level. As soon as the widest opening available within a distance level is deemed too narrow given the width of the vessel, the maximum reachable distance has been reached. Formally, we define $f$ to be the algorithm outlined in Algorithm \ref{alg:feasibility_pooling}.
	\begin{algorithm}
		\scriptsize
		\begin{algorithmic}
			\Require
			\Statex Vessel width $W \in \mathbb{R}^+$
			\Statex Angle between neighboring sensors $\theta$
			\Statex Sensor rangefinder measurements for current sector $\bm{x} = \{x_1, \dots, x_n\}$
			\Procedure{FeasibilityPooling}{$\bm{x}$}
			\State Initialize $\mathcal{I}$ to be the indices of $\bm{x}$ sorted in ascending order according to the measurements $x_i$
			\For{$i \in \mathcal{I}$}
			\State Arc-length $d_i \gets \theta x_i$
			\State Opening-width $y \gets d_i/2$
			\State Opening was found $s_i \gets$ $false$
			\For{$j \gets 0$ to $n$}
			\If{$x_j > x_i$}
			\State $y \gets y + d_i$
			\If{$y > W$}
			\State $s_i \gets$ $true$
			\State \textbf{break}
			\EndIf
			\Else
			\State $y \gets y + d_i/2$
			\If{$y > W$}
			\State $s_i \gets$ $true$
			\State \textbf{break}
			\EndIf
			\State $y \gets 0$
			\EndIf
			\EndFor
			\If{$s_i$ is $false$}
			\Return $x_i$
			\EndIf
			\EndFor
			\EndProcedure
		\end{algorithmic}
		\caption[Feasibility pooling algorithm]{Feasibility pooling for rangefinder sensors \cite{meyer_ASV_IEEE}.}
		\label{alg:feasibility_pooling}
	\end{algorithm}
	\subsubsection{Motion detection}
	Simply feeding the pooled current rangefinder sensor readings to the agent's policy network, will, without any doubt, be insufficient for the agent to learn a policy for intelligently avoiding moving obstacles. A continuous snapshot of the environment can facilitate a purely reactive (but still intelligent \cite{meyer_ASV_IEEE}) agent in a static environment, but without explicit or implicit knowledge of the nearby obstacles' velocities, such an agent will invariantly fail when placed in a dynamic environment, as it will be unable to distinguish between stationary and moving obstacles.
	
	An implicit approach worth mentioning is to process the sensor readings sequentially using a Recurrent Neural Network (RNN). In recent years, RNN architectures, such as Long Short-Term Memory LSTM, have gained a lot of traction in the ML research community \cite{lipton2015criticalRNN} and been successfully applied to sequential RL problems. An example of this is the LSTM-based AlphaStar agent, which reached grandmaster level in the popular real-time strategy game StarCraft II \cite{AlphaStar}. It is therefore possible that a high-performing collision avoidance policy could be found by feeding a recurrent agent with sensor readings.  If such an implementation was shown to be successful, it would facilitate a very straight-forward transition to an implementation on a physical vessel, as no specialized sensor equipment for measuring object velocities would be needed. However, even if sequentially feeding sensor readings to a recurrent network might sound relatively trivial, the motion of the vessel would induce rotations of the observed environment, complicating the situation. Initial experimentation with an off-the-shelf recurrent policy compatible with our simulation environment confirmed the difficulties with this approach. Even with a purely static environment, the recurrent agent was incapable of learning how to avoid collisions. 
	
	Thus, this preliminary study will focus on the explicit approach, i.e. providing the obstacles' velocities as features in the agent's observation vector. Admittedly, while the implementation of this is trivial in a simulated environment, as obstacle velocities can simply be accessed as object attributes, a real-world implementation will necessitate a reliable way of estimating obstacle velocities based on sensor data. However, even if this can be challenging due to uncertainty in the sensor readings, object tracking is a well-researched computer vision discipline. We reserve the implementation of such a method to future research, but refer the reader to \cite{granstrom2016extendedobjecttracking} for a comprehensive overview of the current state of the field.
	
\begin{figure}[ht!]
		\includegraphics[width=\linewidth]{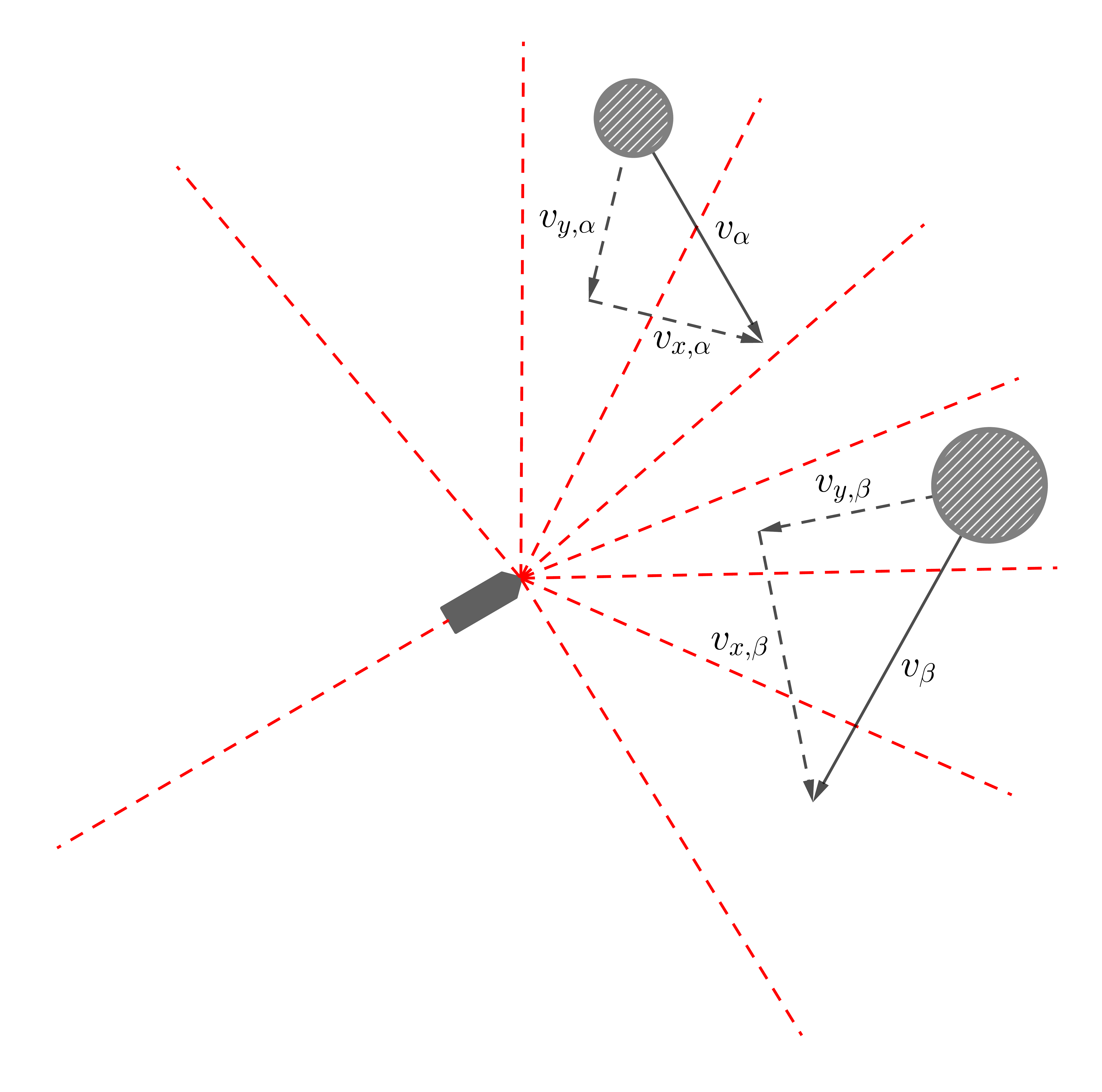}
		\caption[Illustration of the velocity decomposition for moving obstacles]{Velocity decomposition for two moving obstacles, $\alpha$ and $\beta$. For each obstacle, its velocity vector is decomposed into $x$ and $y$ components relative to the obstacle sector, such that the decomposed y-component is parallel to the center line of the corresponding sector, and has a positive value if it is moving \textit{towards} the vessel.}
		\label{fig:vel_decomposition_illustration}
	\end{figure}

	For each sector, we provide the decomposed velocity of the closest moving obstacle within the sector as features for the agent's observation vector. Specifically, the decomposition, which yields the x and y component of the obstacle velocity, is done with respect to the coordinate frame in which the y-axis is parallel to the center line of the sensor sector in which the obstacle is detected. This is illustrated in Figure \ref{fig:vel_decomposition_illustration}. For each sector $k$, we denote the corresponding decomposed $x$ and $y$ velocities as $v_{x, k}$ and $v_{y, k}$, respectively. Naturally, if there are no moving obstacles present within the sector, both components are zero.
	\subsubsection{Perception state vector}
	
	As having access to both obstacle distances and obstacles velocities is critical to achieve satisfactory obstacle-avoiding agent behavior, we include both in the perception state vector.
	
	To avoid discontinuities in the obstacle distance features caused by the sudden transition from $0$ to $S_r$ at the point of detection, we introduce the concept of obstacle \textit{closeness}. The \textit{closeness} to an obstacle is such that it is $0$ if the obstacle is undetected, i.e. further away from the vessel than the maximum range of the distance sensors, and $1$ if the vessel has collided with the obstacle. Furthermore, within this range, is it reasonable to map distance to closeness in a logarithmic fashion, such that, in accordance with human intuition, the difference between $10$m and $100$m is more significant than the difference between, for instance, $510$m and $600$m. Formally, we have that a distance $d$ maps to closeness $c(d) : \mathbb{R} \mapsto [0, 1]$ according to
	\begin{equation}
		c(d) = \clip{\left(1 - \frac{\log{(d+1)}}{\log{(S_r+1)}}, \;0, \;1\right)}
	\end{equation}
	By concatenating the reachable distance and the decomposed obstacle velocity from every sector, we then define the perception state vector $s_{p}$ as
	\begin{equation}
		s_{p}^{(t)} = \left[ \underbrace{c\left(\left(\bm{w}_1^{(t)}\right)\right),\; v_{x, 1}^{(t)},\; v_{y, 1}^{(t)}}_\text{First sector},\; \dots\right]^T
	\end{equation}
	\subsection{Reward function}
	
	Any RL agent is motivated by the pursuit of maximizing its reward. The simplest, and thus highly sought-after approach to rewarding RL agents is to reward it at the end of each episodes - at that point, one already knows if the agent succeeded or failed. However, given the length of a full episode, such a reward function turns out extremely sparse, leaving the agent with a near impossible learning task. This calls for a continuous reward signal, rewarding the agent based on its current adherence to its objectives, i.e. how well it is currently doing with respect to both path following and obstacle avoidance. Given the complexity of the dual-objective learning problem focused on in this study, as well as the general tendency of RL agents' to exploit the reward function in any way possible (e.g. standing still, going in circles), designing an appropriate rewards function $r^{(t)}$ is paramount to the agent exhibiting the desired behavior after training.
	
	It is natural to reward the agent separately for its performance in the two relevant domains: path following and collision avoidance. Thus, we introduce the independent reward terms $r_{path}^{(t)}$ and $r_{colav}^{(t)}$, representing the path-following and the obstacle-avoiding reward components, respectively, at time $t$. Furthermore, as suggested in \cite{meyer_ASV_IEEE}, we introduce the weighting coefficient $\lambda \in [0, 1]$ to regulate the trade-off between the two competing objectives. In addition, as it is crucial to penalize the agent whenever it collides with an obstacle, we represent this by the negative reward term $r_{collision}$, which is activated upon collision. This leads to the preliminary reward function 
	\begin{equation}\label{eq:preliminary_reward_function}
		r^{(t)} = \begin{cases}
			r_{collision}, & \text{if collision}\\
			\lambda r_{path}^{(t)} + \left( 1 - \lambda \right) r_{colav}^{(t)}, & \text{otherwise}
		\end{cases}
	\end{equation}
	\subsubsection{Path following performance}
\begin{figure}[ht]
		\centering
		\hspace{-1cm}\begin{subfigure}{0.8\linewidth}
			\centering
			\includegraphics[width=\textwidth]{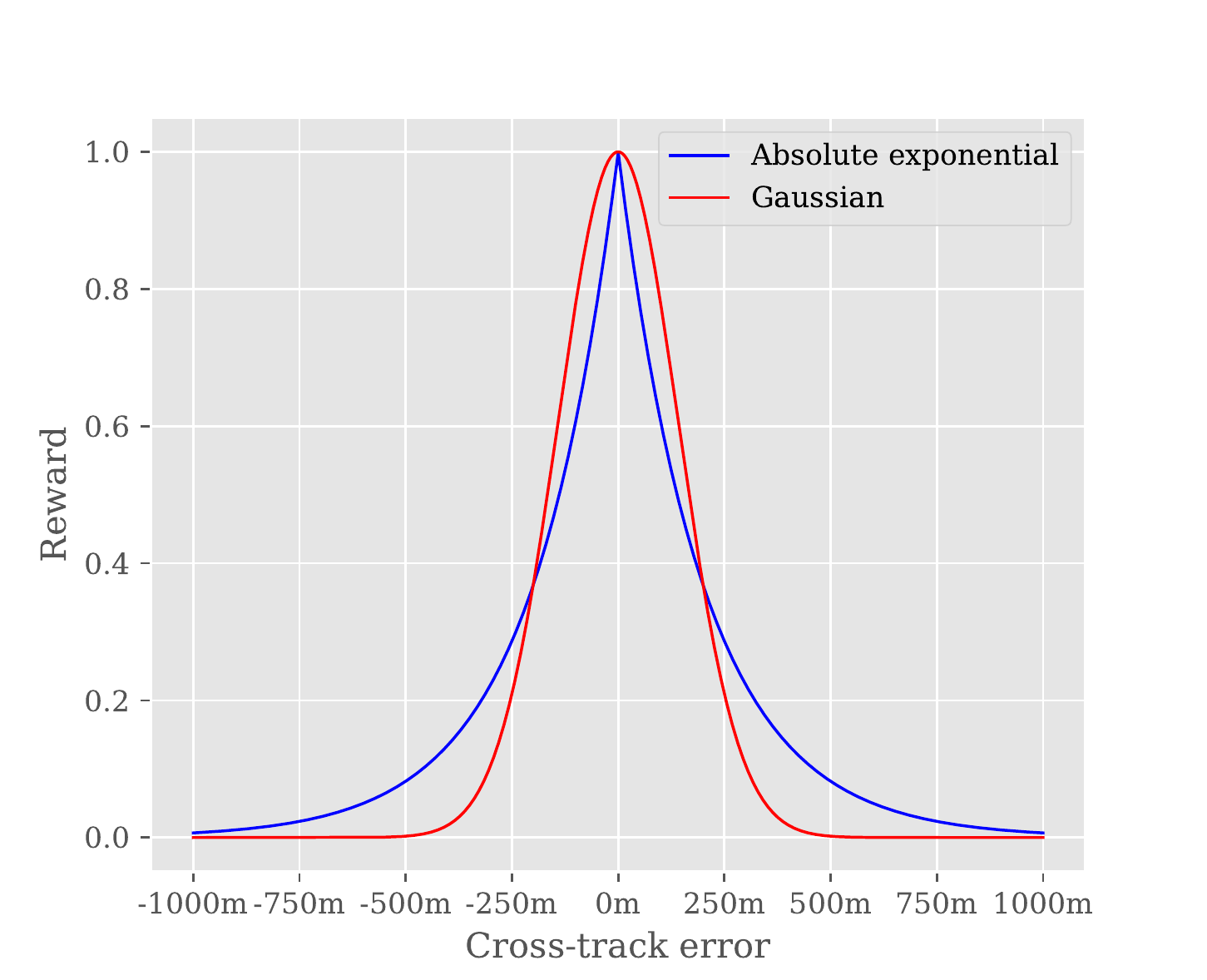}
			\subcaption{Cross-section of the path-following reward landscape assuming path-tangential full-speed motion visualized for both Gaussian and absolute exponential kernels for cross-track error rewarding.}
			\label{fig:reward_landscape_2d}
		\end{subfigure}
		\begin{subfigure}{\linewidth}
			\centering
			\includegraphics[width=\textwidth]{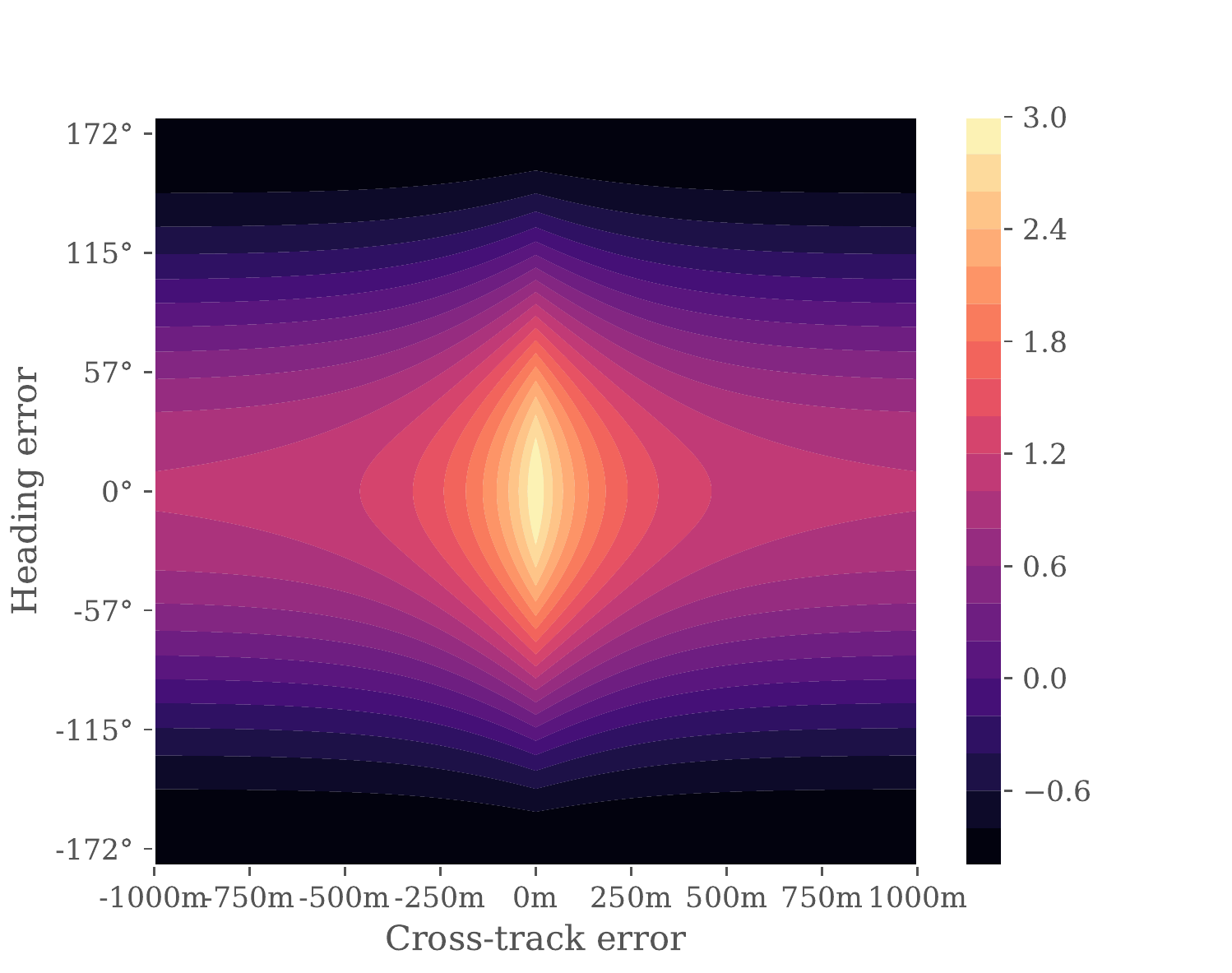}
			\subcaption{Path-following reward function assuming full-speed motion.}
			\label{fig:reward_landscape_3d}
		\end{subfigure}
		\caption{Cross-section and level curves for the path-following reward function with $\gamma_{\epsilon} = 0.05$.}
	\end{figure}

	A natural approach to incentivize path adherence is to reward the agent for minimizing the current absolute cross-track error $\left|\epsilon^{(t)}\right|$. In \cite{martinsen2018}, a Gaussian reward function centered at $\epsilon=0$ with standard deviation $\sigma_e$ was suggested. However we argue that the absolute exponential reward function $\exp{\left(-\gamma_{\epsilon} \left|\epsilon^{(t)}\right|\right)}$ has more desirable characteristics due to its fatter tails, as seen in Figure \ref{fig:reward_landscape_2d}. By avoiding the vanishing improvement gradient of the Gaussian reward occurring at large absolute cross-track errors, the absolute exponential reward function ensures that the agent is rewarded even for a slight improvement to a very unsatisfactory state.  
	However, this alone does not reflect our desire for the agent to actually make progress along the path - and thus, the RL agent, greedy as it is, will eventually develop a policy of standing still indefinitely after closing the gap to the path. Thus, the reward signal must be expanded upon so that it incorporates the incentivization of motion - and not just arbitrary motion, but movement in the right direction. 
	
	The already defined look-ahead heading error term $\tilde{\psi}$ is a natural basis for formalizing this. Specifically, we consider the term $\frac{u^{(t)}}{U_{max}} \cos{\tilde{\psi}^{(t)}}$, with $U_{max}$ being the maximum vessel speed, which effectively yields zero reward if the vessel is heading in a direction perpendicular to the path, and a negative reward if the agent is tracking backwards. Multiplying this with the cross-track error reward component defined earlier is a natural choice, and yields the provisional reward function
	\begin{equation*}
		r_{path}^{(t)} = \underbrace{\frac{u^{(t)}}{U_{max}} \cos{\tilde{\psi}^{(t)}}}_\text{Velocity-based reward} \underbrace{\exp{\left(-\gamma_{\epsilon} |\epsilon^{(t)}|\right)}}_\text{CTE-based reward}
	\end{equation*}
	Given this reward function, however, we note that, if the vessel is standing still (i.e. $u^{(t)} = 0$), or if it is heading in a direction perpendicular to the path (i.e. $\tilde{\psi}^{(t)} = \pm \frac{\pi}{2}$), the agent will receive zero reward regardless of the cross-track error, which is undesired. Similarly, if the cross-track error grows very large, i.e. $\exp{\left(-\gamma_{\epsilon} \left|\epsilon^{(t)}\right|\right)} \to 0$, the reward signal will be zero regardless of the vessel velocity and heading. Thus, we add constant multiplier terms $\gamma_r$ to both reward components, yielding the following expression for the final path-following reward function
	\begin{equation}
		r_{path}^{(t)} =  \underbrace{\left(\tfrac{u^{(t)}}{U_{max}} \cos{\tilde{\psi}^{(t)}} + \gamma_r \right)}_\text{Velocity-based reward} \underbrace{\left(\exp{\left(-\gamma_{\epsilon} |\epsilon^{(t)}|\right)} + \gamma_r\right)}_\text{CTE-based reward} -\gamma_r^2
	\end{equation}
	\noindent where the $-\gamma_r^2$ term is added to remove the constant reward bias implied by the function choice. 
	
	\subsubsection{Static obstacle avoidance performance}
	
	Collision avoidance involves both collisions with other vessels as well as avoiding running ashore (or colliding with some other static obstacle). However, the two aspects should be treated separately, as would any human sailor. In the following, we refer to the former as dynamic, and the latter as static obstacle avoidance.
	
	In order to encourage obstacle-avoiding guidance behavior, penalizing the agent for the closeness of nearby terrain in a strictly increasing manner seems reasonable. However, we note that the severity of closeness intuitively does not increase linearly with distance, but instead increases in some quasi-exponential fashion.
	
	Furthermore, given the presence of a nearby static obstacle, it seems clear that the penalty given to the agent must depend on the orientation of the vessel with regards to the obstacle in such a manner that obstacles located near the stern of the vessel are of significantly lower importance than obstacles that are currently right in front of the it.
	
	Thus, given a static obstacle located at distance $x$, at the angle $\theta$ with respect to the centerline of the vessel, we propose the penalty function
	\begin{equation}\label{eq:r_obst_stat}
		r_{obst, stat}^{(t)} = - \underbrace{\frac{1}{{1 + \gamma_{\theta, stat}|\theta|}}}_\text{Weighting term} \underbrace{\alpha_x \exp{(-\gamma_x x)}}_\text{Raw closeness penalty}
	\end{equation}
	
	where $\alpha_x$ is in the order of magnitude of the sensor range, such that sufficiently high negative rewards are given as objects get closer to the own-ship.
	
\begin{figure}[ht]
		\centering
		\includegraphics[width=\linewidth]{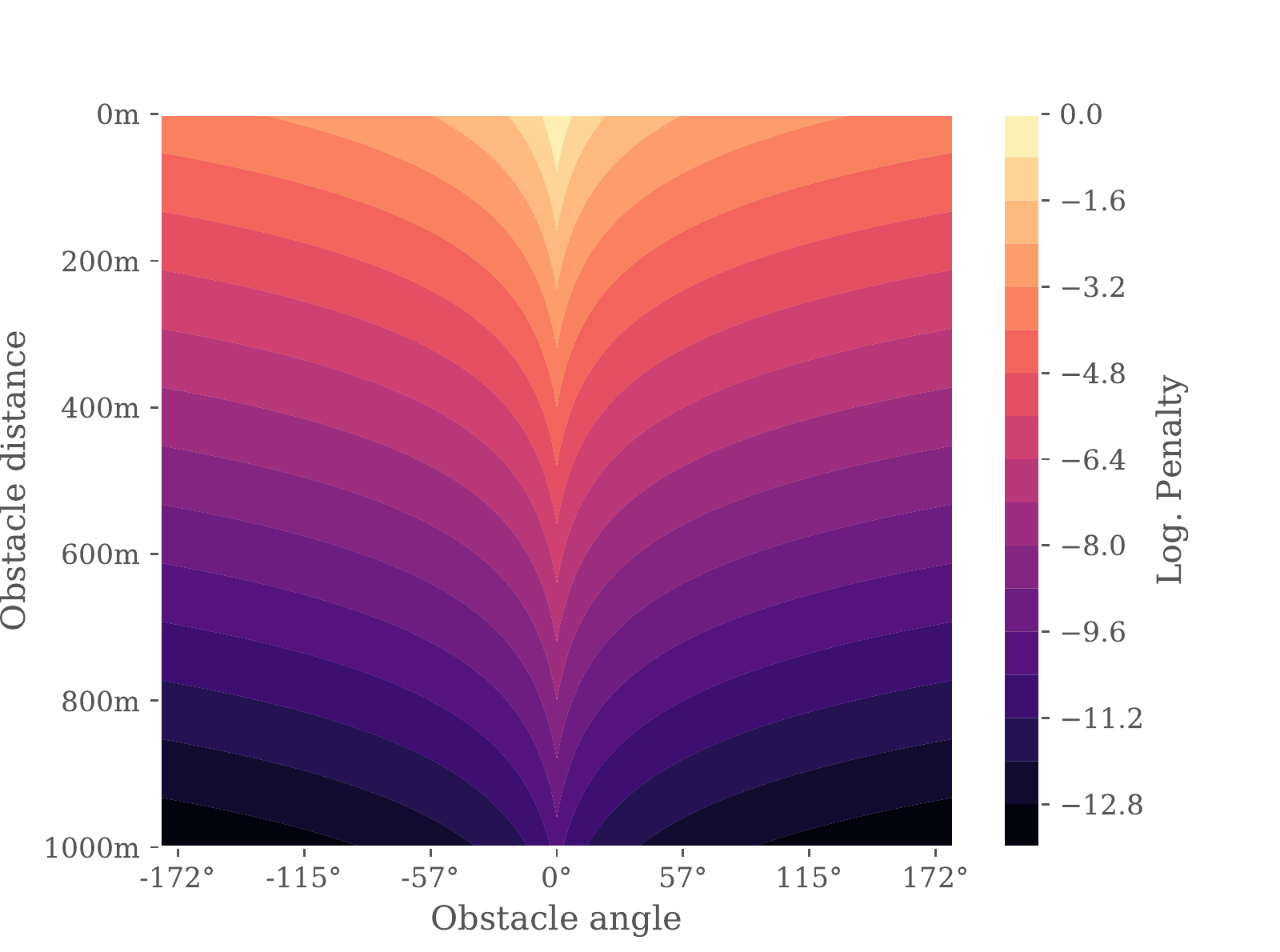}
		\caption[Static obstacle closeness penalty landscape]{Static obstacle closeness penalty landscape as a function of obstacle distance and angle relative to the vessel with the scale parameters $\gamma_{\theta} = 10$, $\gamma_x = 0.1$. The maximum penalty is imposed for obstacles located right in front of the vessel.}
		\label{fig:obst_reward_landscape}
	\end{figure}

	For practical reasons, we use the distances measured by the rangefinder sensors as surrogates for obstacle closeness, and penalize each sensor reading according to $r_{obst, stat}(x_i, \theta_i)$, where $x_i$ is the $i^{th}$ distance sensor measurement and $\theta_i$ is the vessel-relative angle of the corresponding sensor ray. In order to to cancel the dependency on the specific sensor suite configuration, i.e. the number of sensors and their vessel-relative angles, that arises when this penalty term is summed over all sensors, we compute the overall static obstacle-avoidance reward according to the weighted average
	\begin{equation}\label{eq:r_coll_stat}
		r_{colav, stat}^{(t)} = -\frac
		{\textstyle \displaystyle\sum_{i=1}^{N}{\frac{1}{1+\gamma_{\theta, stat} |\theta_i|} \alpha_x \exp{(-\gamma_x x_i )} }}
		{\textstyle \displaystyle\sum_{i=1}^{N}{\frac{1}{1+\gamma_{\theta, stat} |\theta_i|}}}
	\end{equation}
	\noindent which is visualised on a logarithmic scale in Figure \ref{fig:obst_reward_landscape}.  
	\subsubsection{Dynamic obstacle avoidance performance}
	
\begin{figure}[htb!]
		\centering
		\includegraphics[ width=0.7\linewidth]{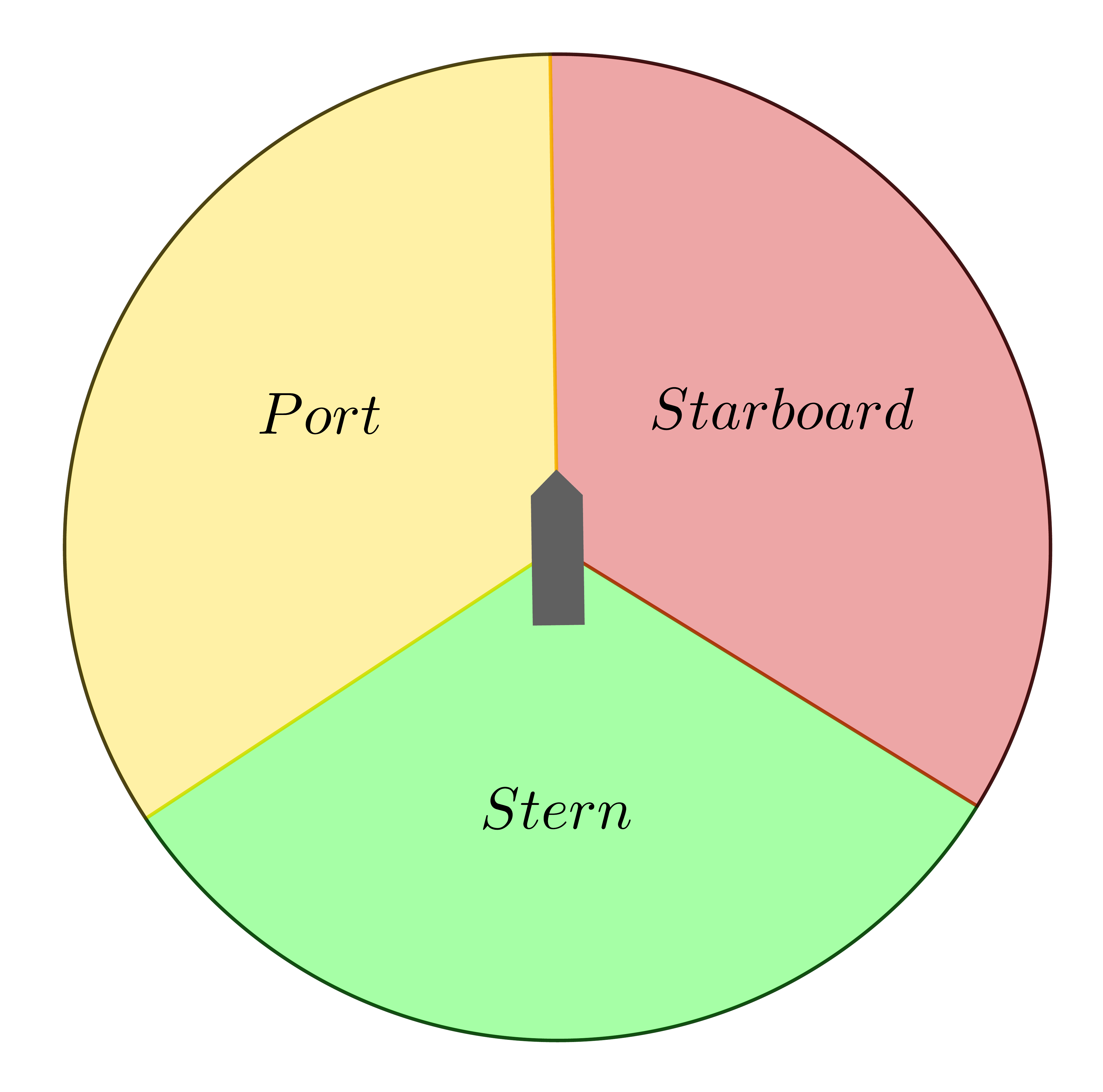}
		\subcaption[COLREG sectors]{Illustration of sectors around the own-ship.}
		\label{fig:colreg_sectors}
	\end{figure}

	For dynamic obstacle avoidance, we expand on the framework developed for static obstacles. Firstly, the penalty needs to reflect the relevant COLREGs. Since the COLREGs are defined according to the bearing of a target ship relative to the own-ship, an intuitive way to guide the RL agent towards COLREGs compliance is to adjust the static obstacle penalty (Equation \ref{eq:r_obst_stat}) according to the relative bearing of the dynamic obstacle. The area around a vessel is normally split into three sectors: port, starboard, and stern, as illustrated in Figure \ref{fig:colreg_sectors}. Therefore, a tunable parameter $\zeta_x$ was added to allow for differentiated weighting of these sectors.
	According to the COLREGs, it is desirable that crossings take place on the port side, meaning that the weighting of sensor readings on the starboard side should be higher. However, since it is assumed in this work that the target vessels have restricted maneuverability, sensor readings on the port side and astern must also be sufficiently penalized. Denoting starboard as "st.b.", we thus have that $\gamma_{x,st.b.} < \gamma_{x,port} \leq \gamma_{x,stern}$.
	\begin{equation}\label{eq:COLREG_weighting}
		\zeta_x(\theta) = \begin{cases}
			\gamma_{x, st.b.}, & \text{if $\theta \geq 0\degree$ and $\theta < 112.5\degree $}\\
			\gamma_{x, port}, & \text{if $\theta \geq -112.5\degree $ and $\theta < 0\degree $}\\
			\gamma_{x, stern}, & \text{if $\theta \geq 112.5\degree $ or $\theta < -112.5\degree $}
		\end{cases}
	\end{equation}

	Furthermore, the reward must reflect the variable risk associated with the direction of a target ship's velocity; an approaching target ship gives rise to a much higher risk than a receding one. In addition, the relatively steep function used as weighting term in Equation \ref{eq:r_obst_stat} was exchanged for a flatter function of the form $1/(1+\text{exp}(x))$, so as to give dynamic obstacles detected around the own-ship sufficient priority. Making adjustments to the static obstacle penalty to adhere to these requirements, the penalty for a single dynamic obstacle was chosen as
	\begin{equation}
		r_{obst, dyn} = -\underbrace{\frac{1}{1+\text{exp}(\gamma_{\theta,dyn} |\theta|)}}_\text{Weighting term} \underbrace{\alpha_x\exp{((\zeta_{v} v_y - \zeta_{x})x)}}_\text{Raw penalty}
	\end{equation}
	\noindent where $x$ is the distance to the obstacle, $\theta$ is the vessel-relative angle (azimuth angle), and $v_y$ is the velocity component in the direction towards the vessel. The scaling factor $\zeta_v$ is given as a function of the angle $\theta$ and the velocity $v_y$, such that the reward efficiently guides the agent towards COLREGs-compliant behavior. It was found that an algorithm with less explicit classification of situations and therefore fewer parameters was in fact harder to tune due to the subsequent high level of dependency between different encounter situations. For instance, since the starboard side is already heavily penalised, lighter weighting of velocity was needed to prevent the agent from reacting too strongly when detecting a target ship on the starboard side. The sign of the velocity component of the target ship towards the own-ship is therefore used to determine whether the target ship is moving towards the own-ship or moving away, which together with the sensor angle $\theta$ provides a good basis for determining a reasonable scaling factor for $v_y$. This scaling factor, $\zeta_v$, is therefore given as

	\begin{equation}\label{eq:vy_weighting}
		\zeta_v(\theta, v_y)=
		\begin{cases}
			\begin{cases}
				\makebox[\temp][l]{$\gamma_{v, st.b.}^+$} & \text{\small{if $v_y \geq$  0}} \\
				\gamma_{v, st.b.}^- & \text{\small{if $v_y$ < 0}}
			\end{cases}
			&\parbox[t]{.2\textwidth}{\vspace*{-0.38cm}\small{if $\theta > 0\degree$ and \\$\theta < 112.5\degree $}}\\
			\begin{cases}
				\makebox[\temp][l]{$\gamma_{v, port}^+$}& \text{\small{if $v_y \geq$  0}} \\
				\gamma_{v, port}^- & \text{\small{if $v_y$ < 0}}
			\end{cases}
			&\parbox[t]{.2\textwidth}{\vspace*{-0.37cm}\small{if $\theta > -112.5\degree $ \\and $\theta < 0\degree $}}\\
			\begin{cases}
				\makebox[\temp][l]{$\gamma_{v, stern}^+$} & \text{\small{if $v_y \geq$  0}} \\
				\gamma_{v, stern}^-  & \makebox[\temp+0.3\temp][l]{\small{if $v_y$ < 0}}
			\end{cases}
			&\parbox[t]{.2\textwidth}{\small{otherwise}}
		\end{cases}
	\end{equation}
	
	Finally, as was done for static obstacles, we then compute the dynamic obstacle-avoidance reward according to the weighted average
	\begin{equation}\label{eq:r_coll_dyn}
		r_{colav, dyn}^{(t)} = -\frac
		{\textstyle \displaystyle\sum_{i=1}^{N}{\frac{(1-\lambda_i)}{1+\text{exp}(\gamma_{\theta,dyn} |\theta_i|)} \alpha_x \exp{((\zeta_{v} v_y^i - \zeta_{x})x_i)} }}
		{\textstyle \displaystyle\sum_{i=1}^{N}{\frac{1}{1+\text{exp}(\gamma_{\theta,dyn} |\theta_i|)}}}
	\end{equation}

	where $\lambda_i$ is a parameter regulating the relative importance of path following and collision avoidance in an encounter situation. This parameter function depends on the velocity $v_y^i$ detected, and takes the distance $x_i$ measured by the sensor as input, according to the logistic function 

	\begin{equation}
		\lambda_i^{(t)} = \frac{1}{1 + \exp{\left(-\gamma_\lambda(v_y^i) x_i^{(t)} + \alpha_\lambda(v_y^i)\right)}}
	\end{equation}
	Here, $\alpha_\lambda(v_y)$ and $\gamma_\lambda(v_y)$ are tunable parameters. Two sets of constant values were chosen such that the overall function for $\lambda_i$ would depend solely on the sign of the speed $v_y$ of the target ship towards the own-ship, giving higher values when $v_y$ < 0. In other words, $\lambda_i$ incorporates the difference in risk between crossing ahead and astern of a target ship, allowing the agent to return to path following quicker in a situation where the target ship is moving away from the own-ship. Formally, we thus have

	\begin{equation}\label{eq:alpha_lambda}
		\alpha_\lambda(v_y) = \begin{cases}
			\alpha_\lambda^+, & \text{if $v_y \geq$ 0}\\
			\alpha_\lambda^-, & \text{if $v_y$ < 0}
		\end{cases}
	\end{equation}

	and

	\begin{equation}\label{eq:gamma_lambda}
		\gamma_\lambda(v_y) = \begin{cases}
			\gamma_\lambda^+, & \text{if $v_y \geq$ 0}\\
			\gamma_\lambda^-, & \text{if $v_y$ < 0}
		\end{cases}
	\end{equation}

	\subsubsection{Total reward}
	Combining the penalties for static and dynamic obstacle avoidance introduced in Eqs. \ref{eq:r_coll_stat} and \ref{eq:r_coll_dyn}, the total collision avoidance penalty function becomes

	\begin{equation}
		r_{colav}^{(t)} = \underbrace{r_{colav, stat}^{(t)}}_\text{Static component} + \underbrace{r_{colav, dyn}^{(t)}}_\text{Dynamic component}
	\end{equation}

	Further, in order to encourage the agent to complete the path within a reasonable time frame, a constant penalty $r_{exists} < 0$ was added. Combining all the elements presented, the expression for the final overall reward function then becomes

	\begin{equation}\label{eq:final_reward}
		r^{(t)} = \begin{cases}
			r_{collision}, & \text{if collision}\\
			\lambda^{(t)} r_{path}^{(t)} + r_{colav}^{(t)} + r_{exists}, & \text{otherwise}
		\end{cases}
	\end{equation}

	The relative weighting of the path and collision avoidance rewards regulated by $\lambda^{(t)}$ was, as previously discussed, found necessary to avoid more lenient collision avoidance manoeuvres when encountering a target ship close to the path. Note that each component $i$ in the sum $r_{colav}$ is multiplied by a weighting term $(1-\lambda_i)$.
	
	Since small values for $\lambda_i$ indicate a critical presence of another ship (and hence that less priority should be given to the path following objective), the smallest value of $\lambda_i$ is chosen to regulate $r_{path}$. Formally, this translates to

	\begin{equation}
		\lambda^{(t)} = \min_i \lambda_i^{(t)}
	\end{equation}

	\subsection{Software implementation}
	
	\subsubsection{Tools and libraries}
	
	Our solution is based on the Python framework OpenAI Gym \cite{OPENAIGYM}, which has become a de facto standard for DRL interfaces. By implementing our simulation environment as an extension of OpenAI Gym, it is straight-forward to train state-of-the-art, parallelizable RL agents on our scenarios. We use \textbf{Stable Baselines} \cite{stable-baselines}, a Python library providing a wide range of well-documented, off-the-shelf RL algorithms, including PPO, for training our agent. The most challenging aspect of the simulation, which is the calculation of the intersection points between the sensor rays and the boundaries of the nearby obstacles, is handled efficiently by the \textbf{shapely} Python library \cite{SHAPELY}, which offers an easy-to-use interface to a wide range of geometric analysis-related operations.
	
	\subsubsection{Simulation parameters}
	
	In our setup, both the policy network as well as the value network used in the PPO algorithm's advantage estimation have two hidden layers with 64 units each, and use the \textit{tanh} activation function across the networks. Furthermore, the hyperparameter values presented in Table \ref{tab:ppo_hyperparams} were used for the PPO algorithm.
	\begin{table}[h]
		\begin{tabular}{lll}
			\hline
			Parameter & Interpretation & Value\\
			\hline
			$\gamma$ & Discount factor & $0.999$ \\
			$T$ & Timesteps per training iteration & $1024$ \\
			$N_A$ & Number of parallel actors & $8$ \\
			$K$ & Training epochs & $10^6$ \\
			$\eta$ & Learning rate & $0.0002$ \\
			$N_{MB}$ & Number of minibatches & $32$ \\
			$\lambda_{PPO}$ & Bias vs. variance parameter& $0.95$ \\
			$c_1$ & Value function coefficient & $0.5$  \\
			$c_2$ & Entropy coefficient& $0.01$  \\
			$\epsilon$ & Clipping parameter & $0.2$ \\
		\end{tabular}
		\caption[Hyperparameters for PPO algorithm]{Hyperparameters for PPO algorithm.}\label{tab:ppo_hyperparams}
	\end{table}
	In terms of the vessel setup, the values in Table \ref{tab:vesselconfig_obstdect} were used.
	\begin{table}[h]
		\begin{tabular}{lll}
			\hline
			Parameter & Interpretation & Value \\
			\hline
			$U_{max}$ & Maximum vessel speed & $2$ m/s \\
			$N$ & Number of sensors & $180$ \\
			$S_r$ & Sensor distance & $1.5$ km\\
			$d$ & Number of sensor sectors & $9$\\
			$\Delta_{LA}$ & Look-ahead distance & $3$ km
		\end{tabular}
		\caption{Vessel configuration}\label{tab:vesselconfig_obstdect}
	\end{table}
	Finally, the parameters in Table \ref{tab:reward_parameters} were used for customizing the reward function. This choice of reward function parameters stems from intuitive reasoning about the desired characteristics of the agent's guidance behavior and how it relates to the parameters. In addition, adjustments were made based on observations made during testing.
	\begin{table}[h]
		\begin{tabular}{l L l}
			\hline
			Parameter & Interpretation & Value \\
			\hline
			$\gamma_e$ & Cross-track error scaling & 0.5 \\
			$\alpha_x$ & Raw COLAV penalty scaling & 75 \\
			$\gamma_{\theta, stat}$ & Sensor angle scaling for static obst. & 10 \\
			$\gamma_{\theta, dyn}$ & Sensor angle scaling for dyn. obst. & 1 \\
			$\gamma_x$ & Static obstacle distance scaling & 0.01 \\
			$\gamma_{v, st.b.}^+$ &Scaling of $v_y \geq 0$, s.b. side & 0.004 \\
			$\gamma_{v, st.b.}^-$ &Scaling of $v_y$ < 0, s.b. side & 0.05 \\
			$\gamma_{v, port}^+$ &Scaling of $v_y \geq 0$, port side & 0.007 \\
			$\gamma_{v, port}^-$ &Scaling of $v_y$ < 0, port side & 0.005\\
			$\gamma_{v, stern}^+$ &Scaling of $v_y \geq 0$, astern & 0.007 \\
			$\gamma_{v, stern}^-$ &Scaling of $v_y$ < 0, astern &  0.005\\
			$\gamma_{x, st.b.}$ & Dyn. obst. distance scaling, st.b. & 0.007 \\
			$\gamma_{x, port}$ & Dyn. obst. distance scaling, port & 0.009 \\
			$\gamma_{x, stern}$ & Dyn. obst. distance scaling, stern & 0.01\\
			$\alpha_\lambda^+$ & Translation of $\lambda$, $v_y \geq 0$ & 4 \\
			$\alpha_\lambda^-$ & Translation of $\lambda$, $v_y < 0$ & 2 \\
			$\gamma_\lambda^+$ & Distance scaling of $\lambda$, $v_y \geq 0$ & 0.003 \\
			$\gamma_\lambda^-$ &  Distance scaling of $\lambda$, $v_y < 0$ & 0.005 \\
			$r_{coll}$ & Collision reward & -10000\\
			$r_{exists}$ & Living penalty & -1\\
			
		\end{tabular}
		\caption{Reward configuration}\label{tab:reward_parameters}
	\end{table}

	\subsection{Evaluation}
	
	To provide a comprehensive basis for evaluating the agent's performance, we test the trained agent in three different test domains.
	
	\subsubsection{COLREGs compliance}
	
	First, artificial vessel encounter scenarios, in which COLREGs compliance easily can be categorized as a success or failure in binary terms, are created to quantify the trained agent's performance in a simple and unambiguous manner. Specifically, we simulate head-on and crossing scenarios, and expect the vessel to adhere to the relevant COLREGs rules. The head-on scenario and the first crossing scenario represent the scenarios illustrated in Figure \ref{fig:colreg_complianec}, which allows for easy comparison later on. 
	
	\subsubsection{Training environment performance}
	
	Next, we provide a statistical evaluation of the agents based on random samples of the training scenario. More precisely, we evaluate the degree to which the agent is avoiding collisions, as well as the degree to which it adheres to its path following objective, by simulating its behavior in new (i.e. unseen) permutations of the training scenario. As described,
	the training environment is challenging, with a dense scattering of both static and dynamic obstacles.

	\subsubsection{Real-world-based experiments}
	
	Finally, based on combining high-fidelity terrain data with AIS tracking data from the Trondheim Fjord area, we construct three digital real-world environments in which the vessel's performance can be evaluated.
	
	The dashed black line represents the desired vessel trajectory. Each other vessel is drawn as its initial position with an arrow whose length corresponds to its initial speed. Additionally, each other vessel's trajectory is drawn as a transparent and dotted red line.
	
	\textbf{Ørland-Agdenes}
	
	This scenario takes place in the heavily trafficked entrance region of the fjord: The region between the municipalities Ørland and Agdenes. After spawning near the coastline, the vessel must blend into two-way traffic and follow the path until it reaches the opening of the fjord. In particular, the agent will be tested on its ability to handle head-on and overtaking situations.
	\begin{figure}[htb!]
		\centering
		\hspace*{-0.2cm}\includegraphics[trim={1cm 0cm 0cm 1cm},clip, width=1.15\linewidth]{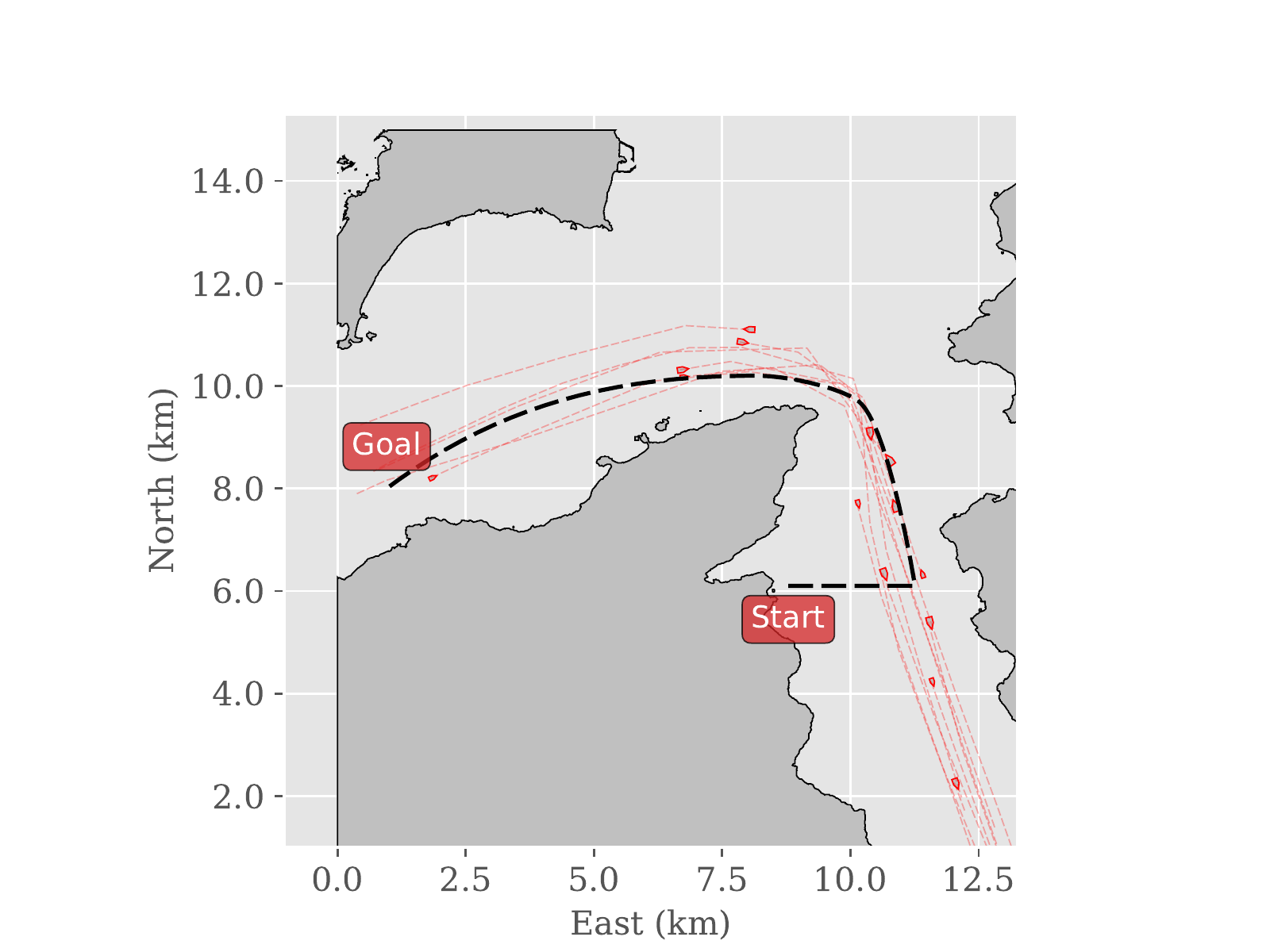}
		\subcaption[Ørland-Agdenes test scenario]{Map of the Ørland-Agdenes test scenario. The dashed black line represents the desired vessel trajectory. Each other vessel is drawn at its initial position. Also, each other vessel's trajectory is drawn as a transparent and dotted red line.}
		\label{fig:orlandagdenest_testscenario}
	\end{figure}
	\FloatBarrier
	
	\textbf{Trondheim}
	
	Spawning next to the Trondheim city center, the agent is expected to cross the fjord end and up at the village Vanvikan. In order to succeed in this scenario, the agent must avoid collisions with the crossing traffic, which is dominated by larger ships.
	\begin{figure}[htb!]
		\centering
		\hspace*{-0.2cm}\includegraphics[trim={1cm 0cm 0cm 1cm},clip, width=1.15\linewidth]{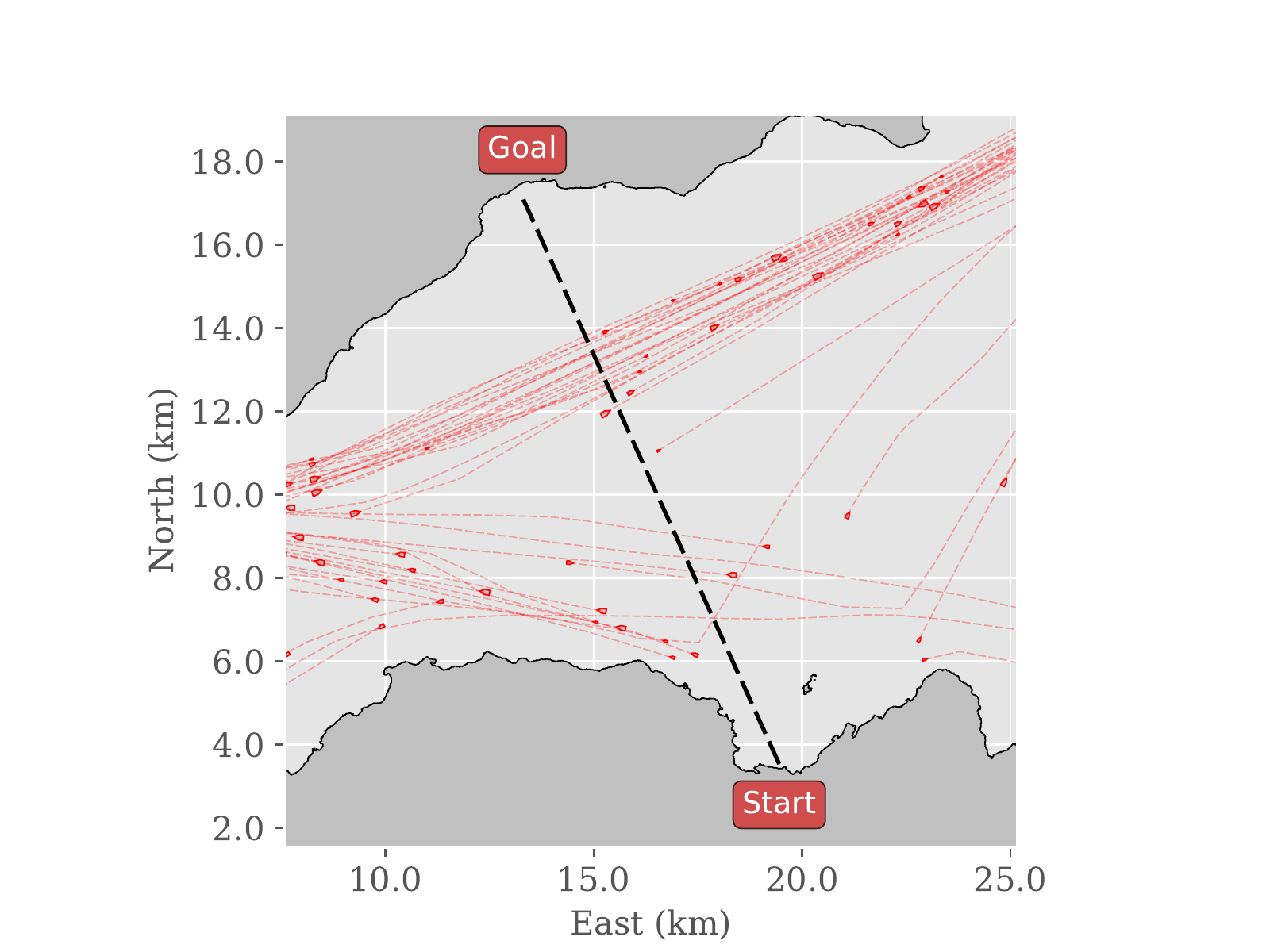}
		\subcaption[Trondheim test scenario]{Map of the Trondheim test scenario.}
		\label{fig:trondheim_testscenario}
	\end{figure}
	\textbf{Froan}
	Froan, which is located off the Trøndelag coast, is an archipelago encompassing hundreds of small, rocky islands. For this reason, it offers uniquely challenging terrain. In this scenario, the agent must carefully navigate through a cluster of small islands, before merging into traffic going to and from Sørburøy, the most populated island in the area. The challenging terrain will test the agent's ability to navigate static obstacles, whereas the traffic, comprised of smaller, fast-moving vessels, will lead to challenging head-on situations, especially in the narrow strait in which the goal is located.
	\begin{figure}[htb!]
		\hspace*{-1.1cm}\includegraphics[trim={0cm 0cm 0 1cm},clip, width=1.15\linewidth]{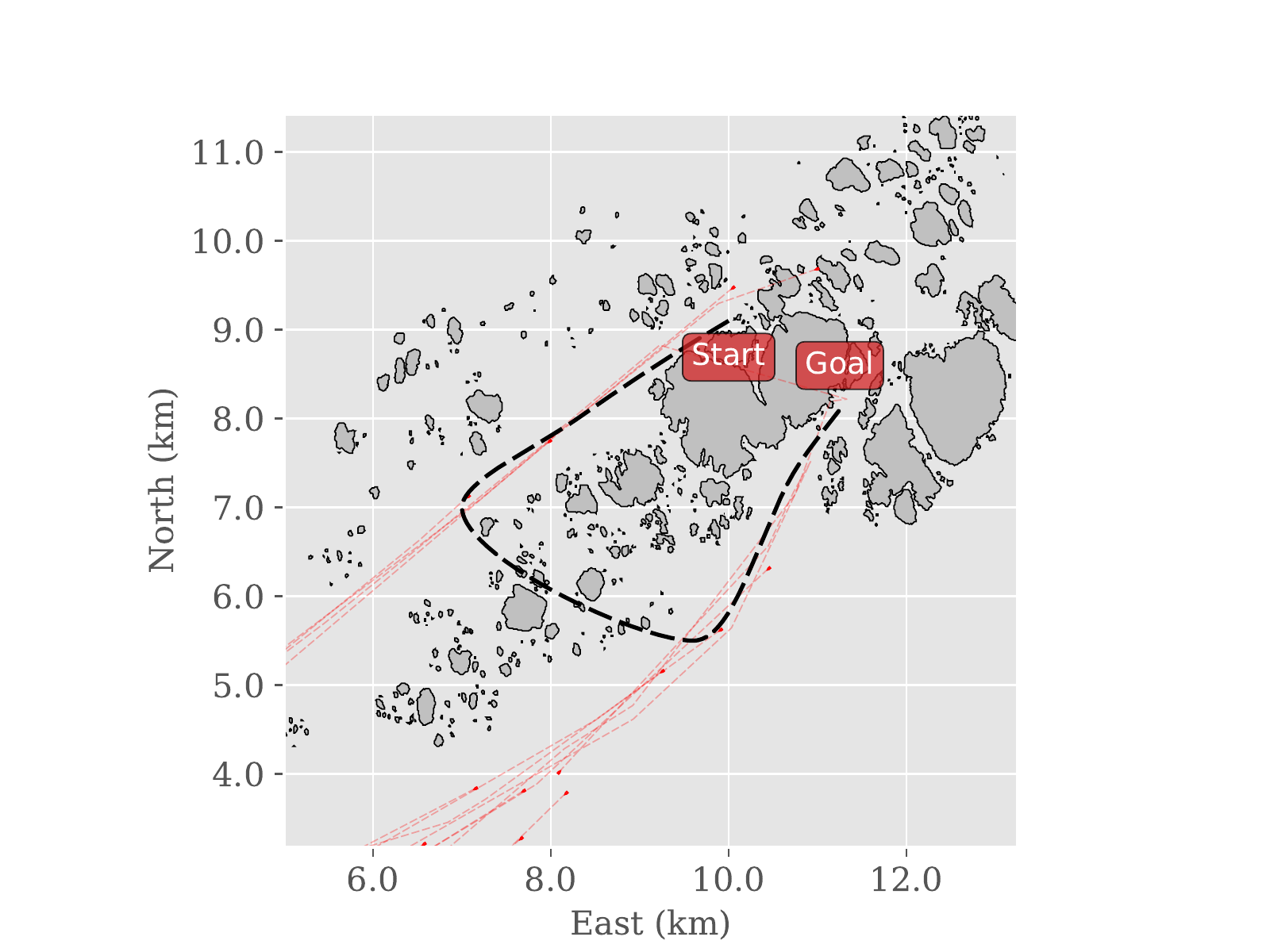}
		\subcaption[Froan test scenario]{Map of the Froan test scenario.}
		\label{fig:froan_testscenario}
	\end{figure}

	\FloatBarrier
	\section{Results and Conclusion}
	
	\subsection{Simulations}
	
	
	\subsubsection{COLREGs compliance}
	
\begin{figure}[H]
		\centering
		\begin{subfigure}{0.8\linewidth}
			\centering	\hspace*{-1.0cm}
			\begin{overpic}[trim={0cm 0.1cm 0 1cm},clip, width=1.2\textwidth]{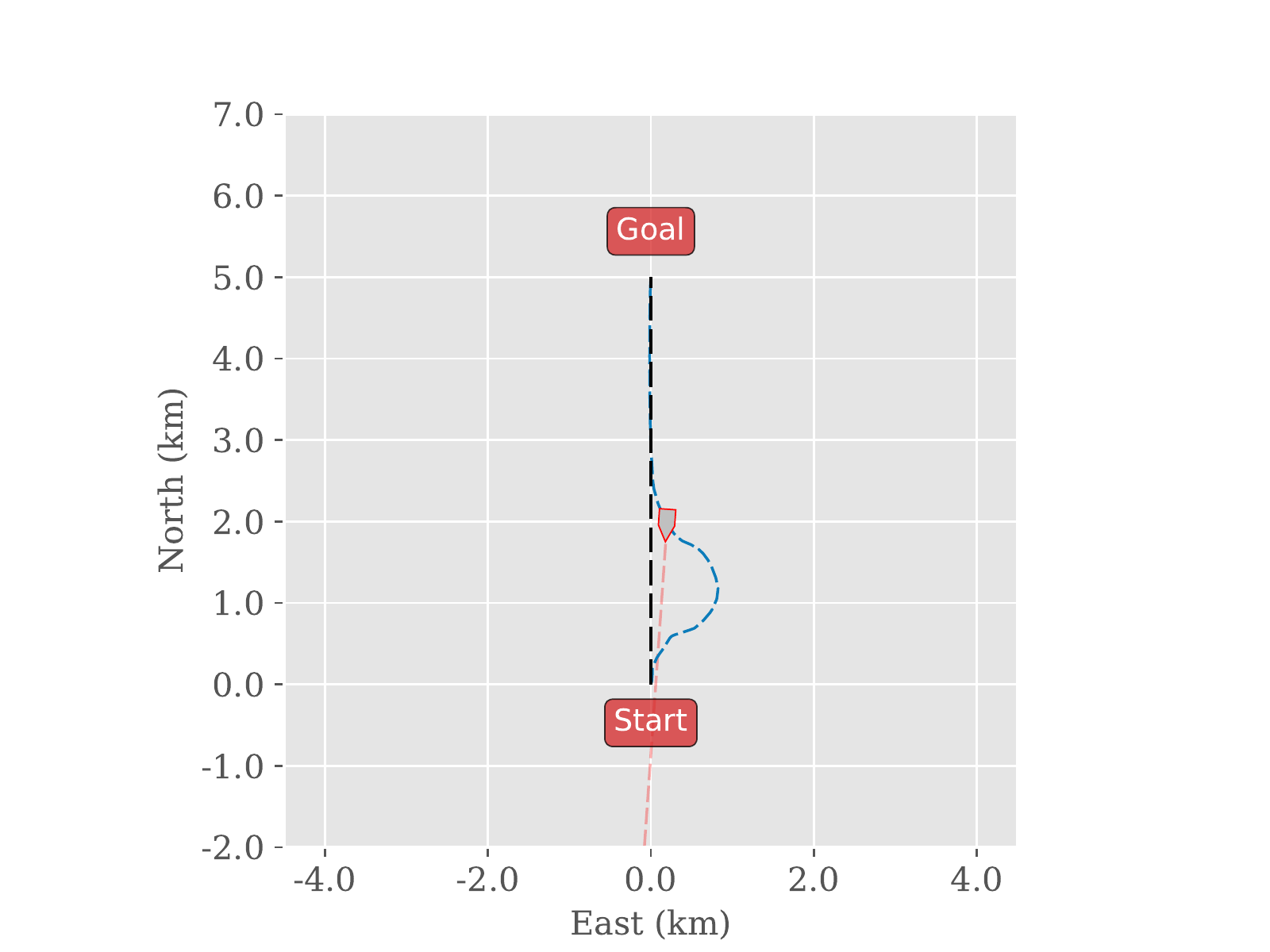}
				\put(65,35){\frame{\includegraphics[trim={5.5cm 3cm 6.8cm 3cm},clip, width=.23\textwidth, height=.35\textwidth]{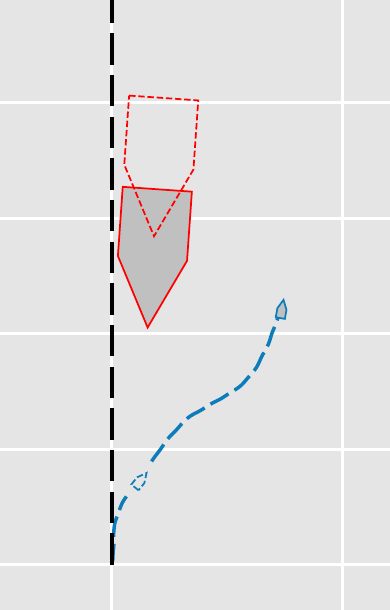}}}  
			\end{overpic}
			\subcaption[Head on test scenario]{Test scenario 1: Head on.}
			\label{fig:headon_testscenario}
		\end{subfigure}
		\begin{subfigure}{0.8\linewidth}
			\centering
			\hspace*{-1.0cm}
			\begin{overpic}[trim={0cm 0.1cm 0 1cm},clip, width=1.2\textwidth]{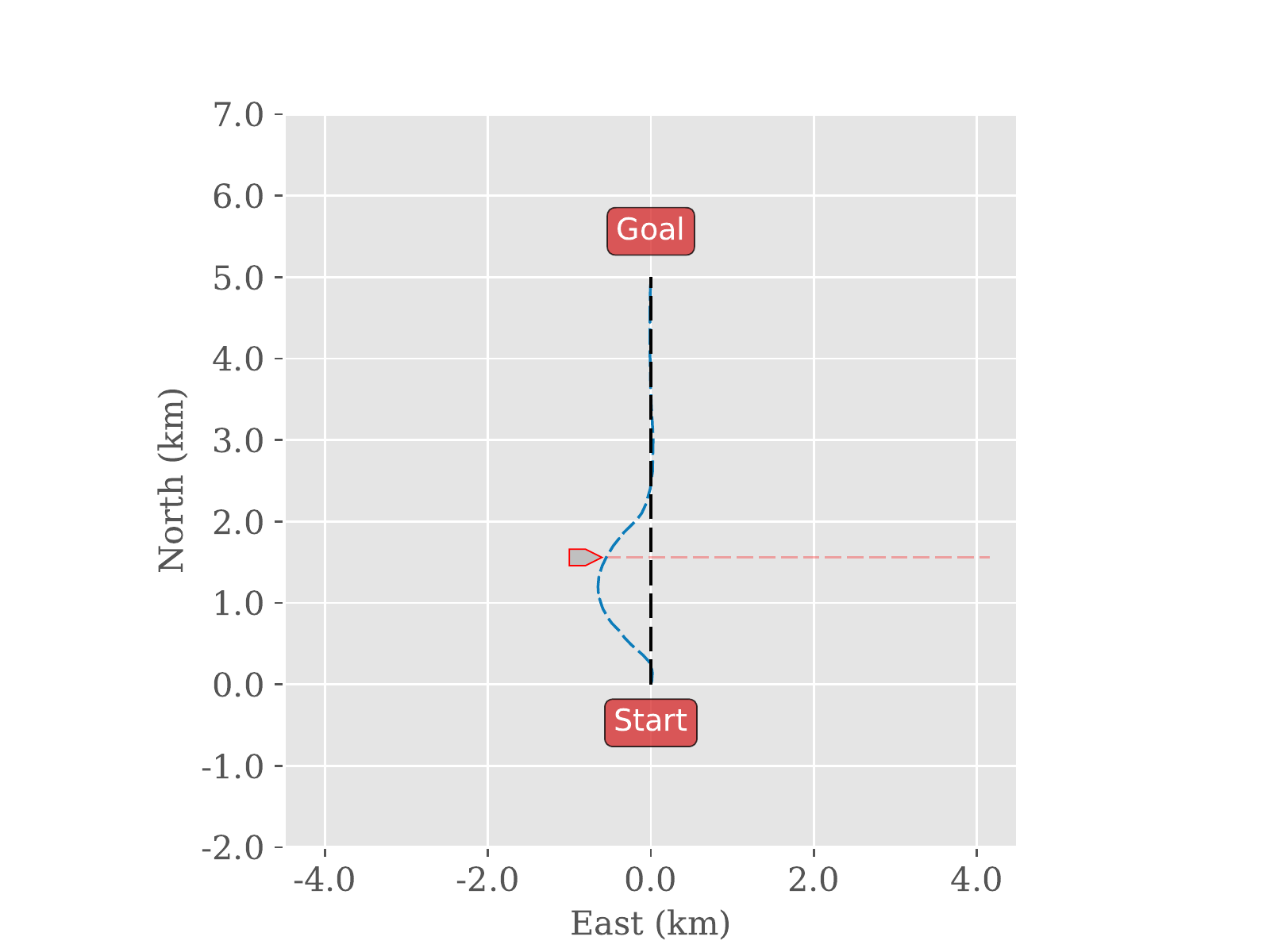}
				\put(65,35){\frame{\includegraphics[trim={7.2cm 2.5cm 4.5cm 3cm},clip, width=.23\textwidth, height=.35\textwidth]{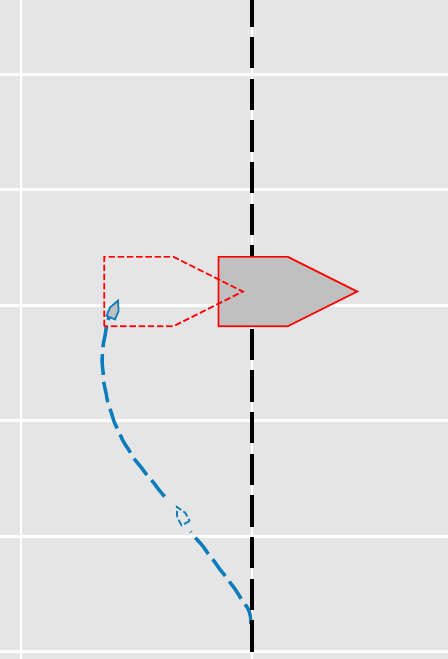}}}  
			\end{overpic}
			\subcaption[Crossing1 test scenario]{Test scenario 2: Crossing from starboard.}
			\label{fig:crossing1_testscenario}
		\end{subfigure}
		\begin{subfigure}{0.8\linewidth}
			\centering
			\hspace*{-1.0cm}
			\begin{overpic}[trim={0cm 0.1cm 0 1cm},clip, width=1.2\textwidth]{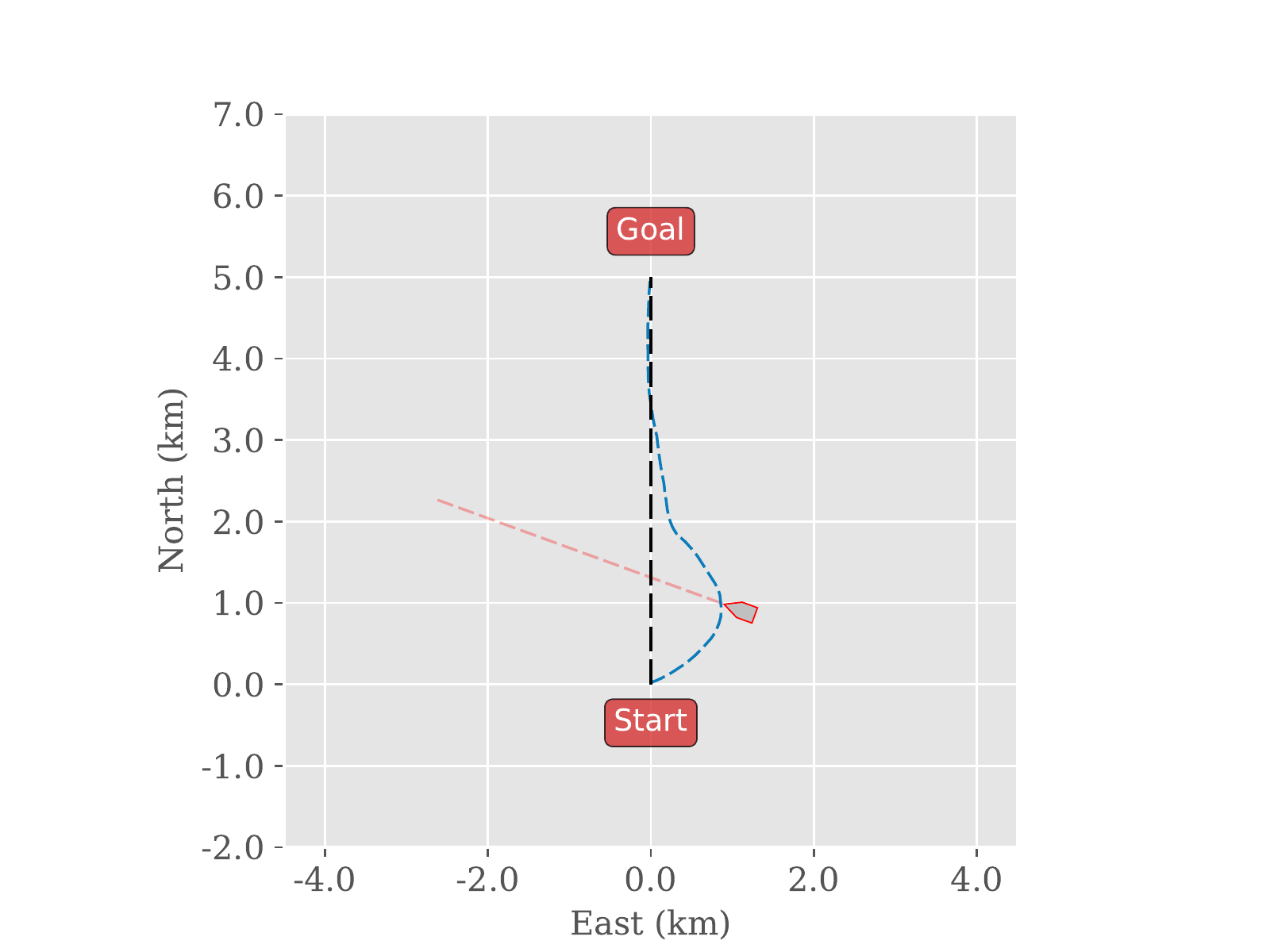}
				\put(65,35){\frame{\includegraphics[trim={6cm 3cm 6.7cm 4cm},clip, width=.23\textwidth, height=.35\textwidth]{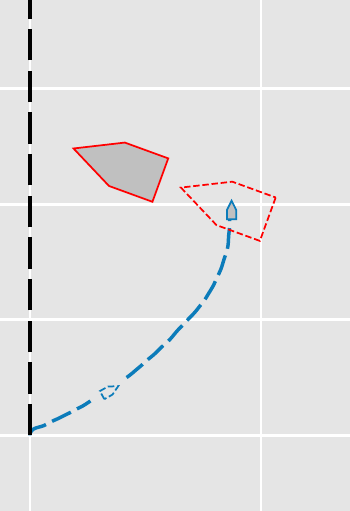}}}
			\end{overpic}
			\subcaption[Crossing2 test scenario]{Test scenario 2: Crossing from port.}
			\label{fig:crossing2_testscenario}
		\end{subfigure}
		\caption{Agent trajectories in the test scenarios are drawn as blue dashed lines, and the target ships with trajectories are drawn in red.}
		\label{fig:colregs_compliance_sims}
	\end{figure}

	To provide insight into the agent's fundamental COLREGs compliance, simple encounter scenarios similar to those seen in Figure \ref{fig:colreg_complianec} were constructed. The results of these simulations can be seen in Figure \ref{fig:colregs_compliance_sims}. Clearly, the agent adheres to the main COLREGs rules outlined. In Figure \ref{fig:headon_testscenario}, the own-ship adheres to \textbf{Rule 14} by altering her course to starboard in a head-on situation. This behaviour was reliably observed when varying the incoming angle of the target ship, denoted $\theta_t$, such that $\theta_t \in [-5\degree, 5\degree]$. Further, as seen in Figures \ref{fig:crossing1_testscenario} and \ref{fig:crossing2_testscenario}, the agent avoids crossing ahead of a target ship when it can make a reasonable maneuver to cross astern, as described by \textbf{Rules 15}, \textbf{16}, and \textbf{18}. It was noted, however, that there is a "cut-off" when the target ship approaches from an angle $\theta_t > 45\degree$. In these situations, it chooses to cross ahead, although with a good margin. This makes intuitive sense, as it effectively resolves the conflict without making sharp maneuvers. However, it is important to note the ambiguity of the COLREGs in these cases, stating that the give-way vessel should "keep well clear". 
	
	\subsubsection{Training environment}
	
\begin{figure}[ht!]
	\centering
	\begin{subfigure}{0.49\linewidth}
		{\includegraphics[trim={0.5cm 0.1cm 1cm 1cm},clip,width=\textwidth]{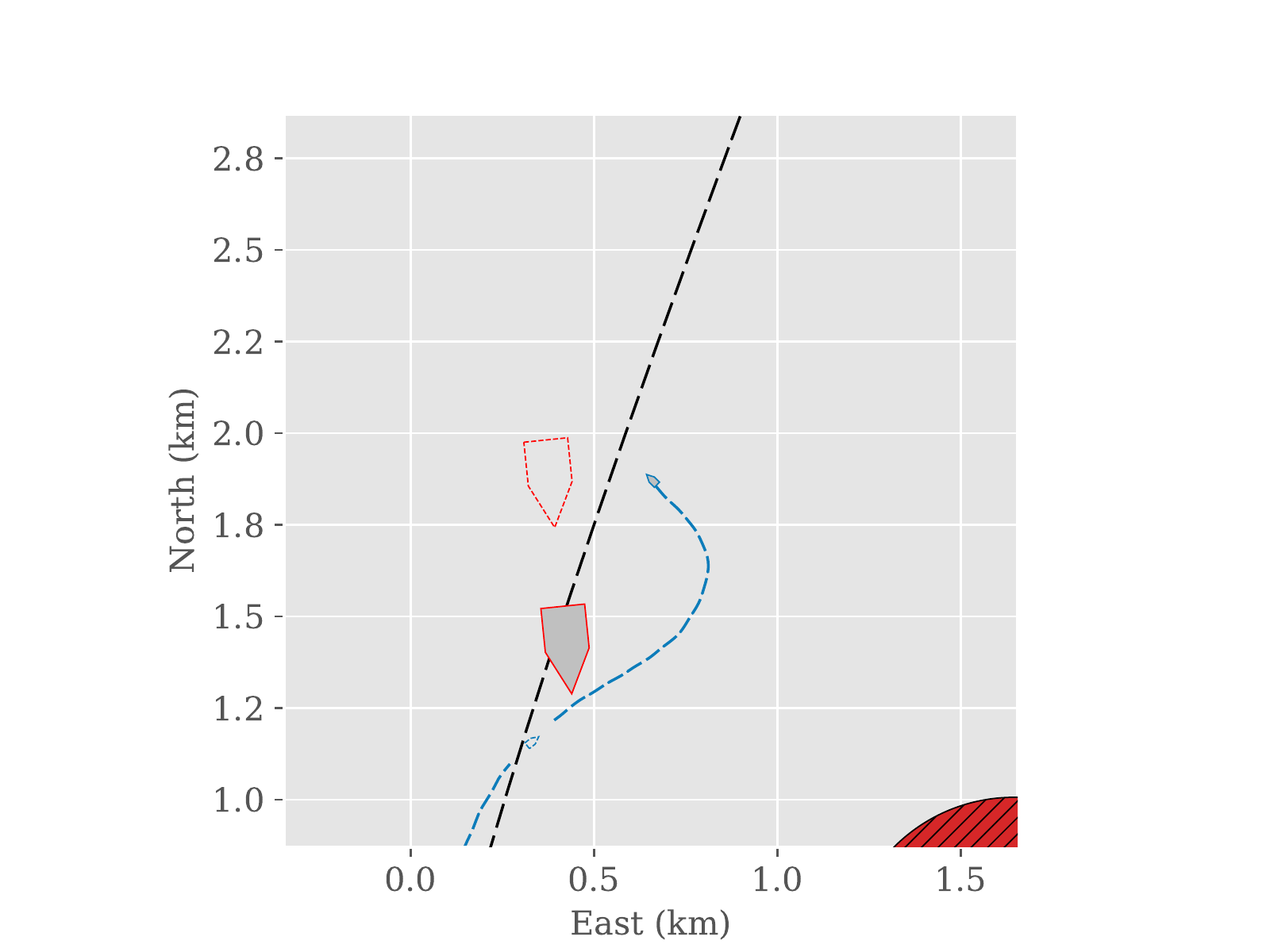}}
		\subcaption{Head-on situation}
	\end{subfigure}
	\begin{subfigure}{0.49\linewidth}
		{\includegraphics[trim={0.5cm 0.1cm 1cm 1cm},clip,width=\textwidth]{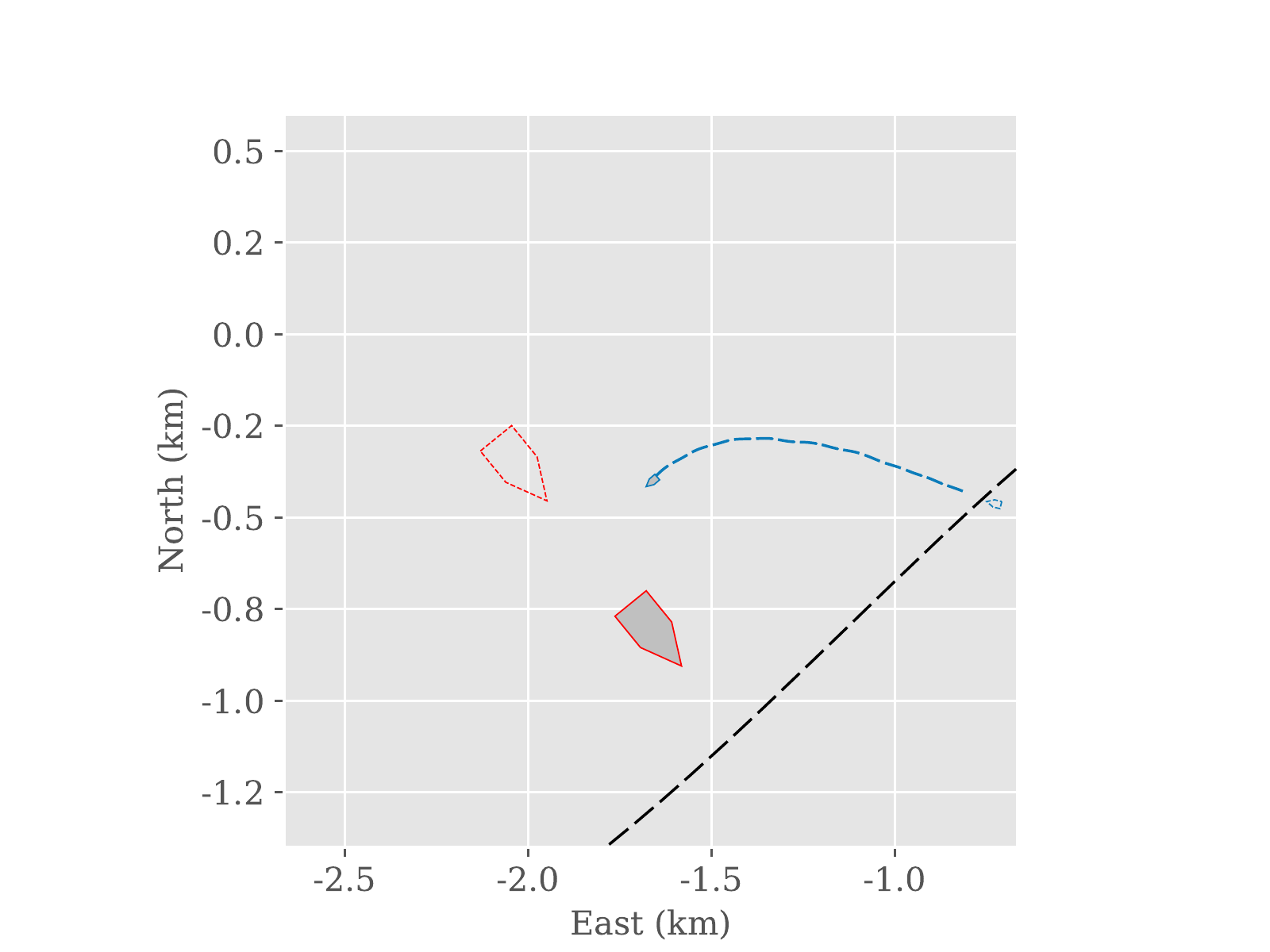}}
		\subcaption{Astern passing}
	\end{subfigure}
	\begin{subfigure}{0.49\linewidth}
		{\includegraphics[trim={0.5cm 0.1cm 1cm 1cm},clip,width=\textwidth]{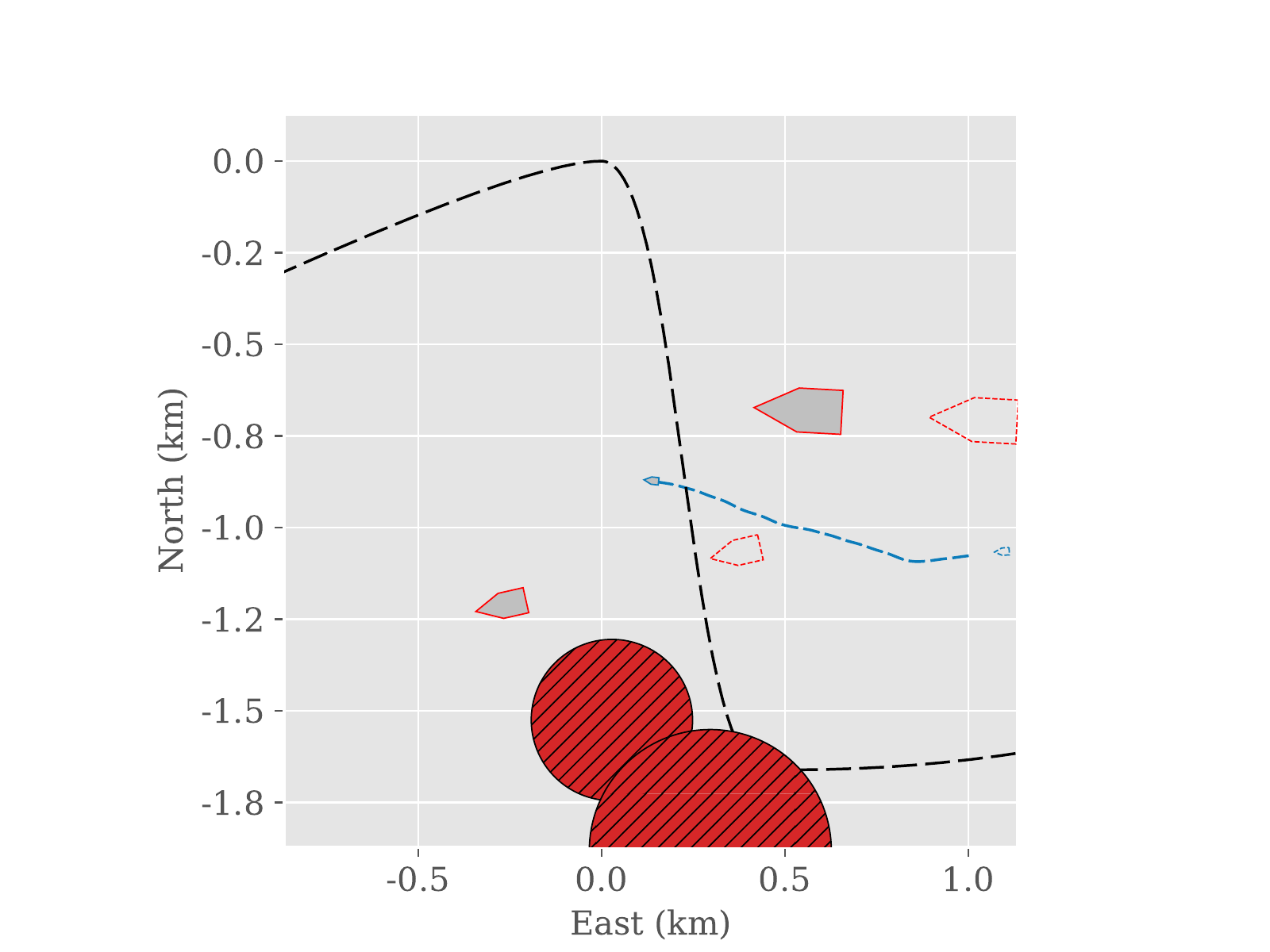}}
		\subcaption{Overtaking}
	\end{subfigure}
	\begin{subfigure}{0.49\linewidth}
		{\includegraphics[trim={0.5cm 0.1cm 1cm 1cm},clip,width=\textwidth]{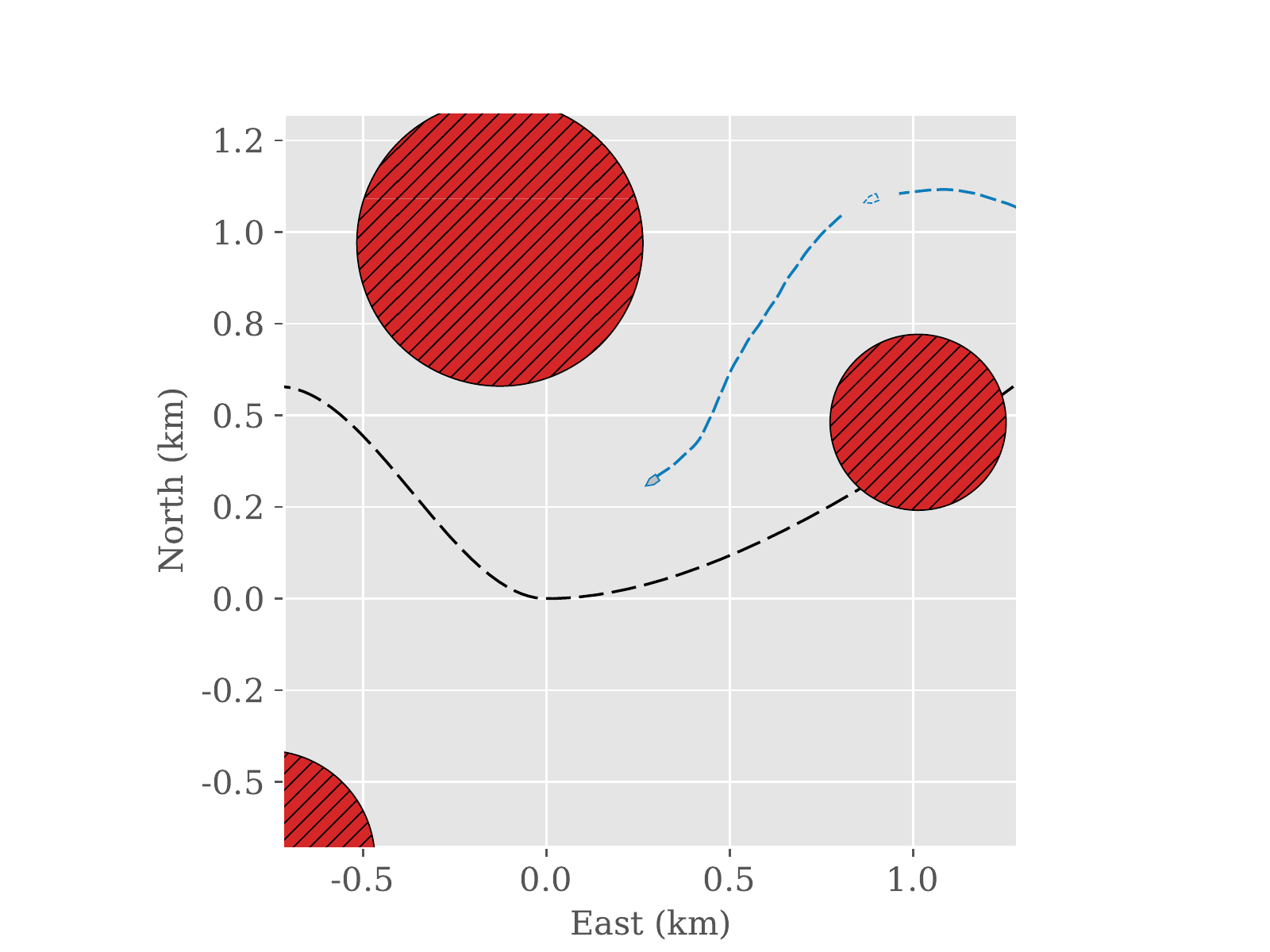}}
		\subcaption{Static COLAV}
	\end{subfigure}
	\caption{Agent performing common naval collision avoidance maneuvers in the training environment. Agent trajectories are drawn with blue dashed lines, and the target ships are drawn in red.}
	\label{fig:training_scenario_sims}
\end{figure}

	Next, common collision avoidance maneuvers from the training environment are shown in Figure \ref{fig:training_scenario_sims}. The snippets presented are representative of the agent's behavior in realistic encounter situations, and show that it is COLREGs-compliant in the situations where the rules can be accurately discerned. Due to the artificial nature of the training environment, the agent is subject to a wide variety of unrealistic situations during training. These have been discarded.

	%
	%
	
	\subsubsection{AIS-based environment}
	
\begin{figure}[ht!]
		\centering
		\begin{subfigure}{0.49\linewidth}
			{\includegraphics[trim={0.5cm 0.1cm 1cm 1cm},clip,width=\textwidth]{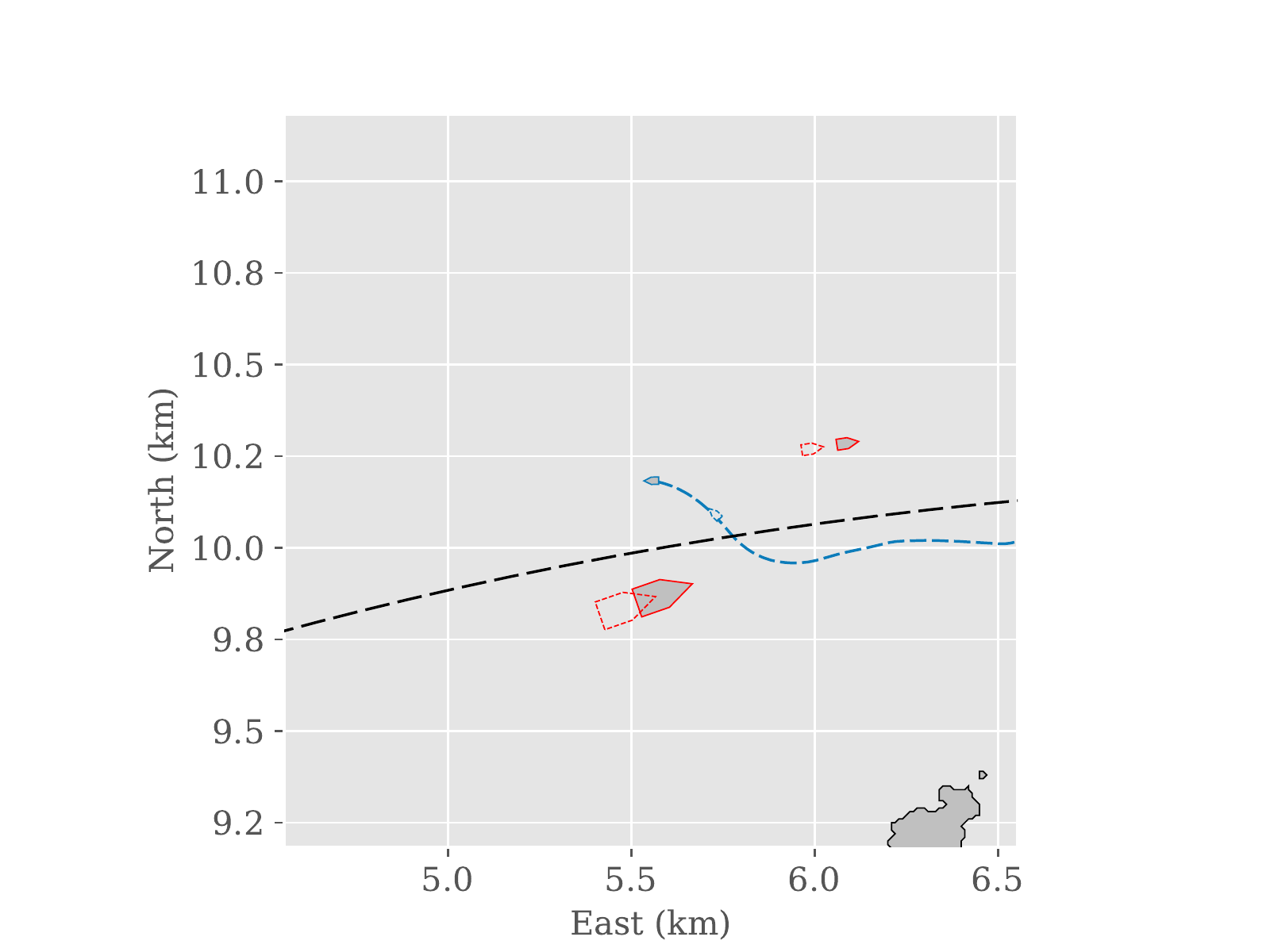}}
			\subcaption{Head-on situation}
		\end{subfigure}
		\begin{subfigure}{0.49\linewidth}
			{\includegraphics[trim={0.5cm 0.1cm 1cm 1cm},clip,width=\textwidth]{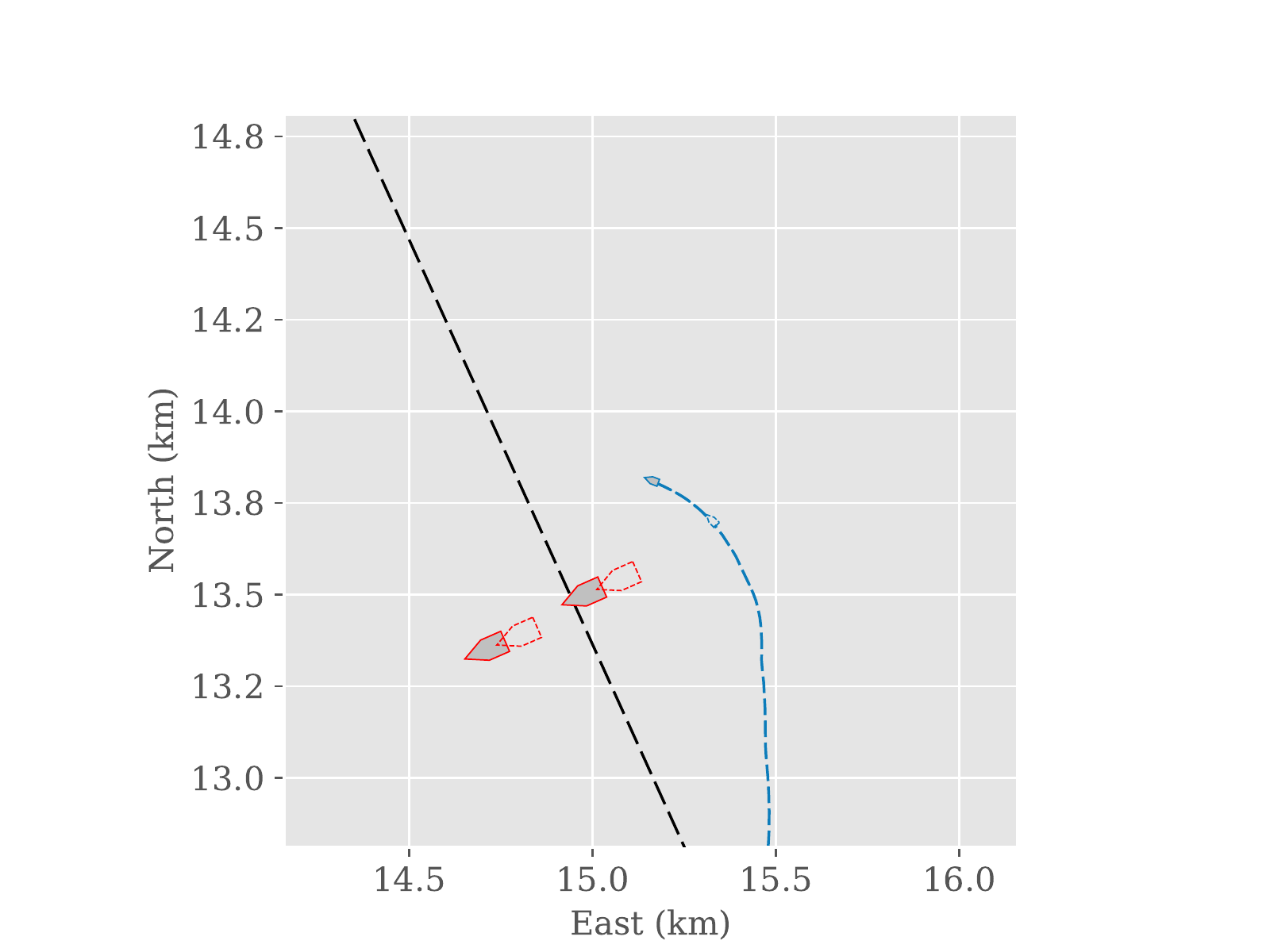}}
			\subcaption{Astern passing}
		\end{subfigure}
		\begin{subfigure}{0.49\linewidth}
			{\includegraphics[trim={0.5cm 0.1cm 1cm 1cm},clip,width=\textwidth]{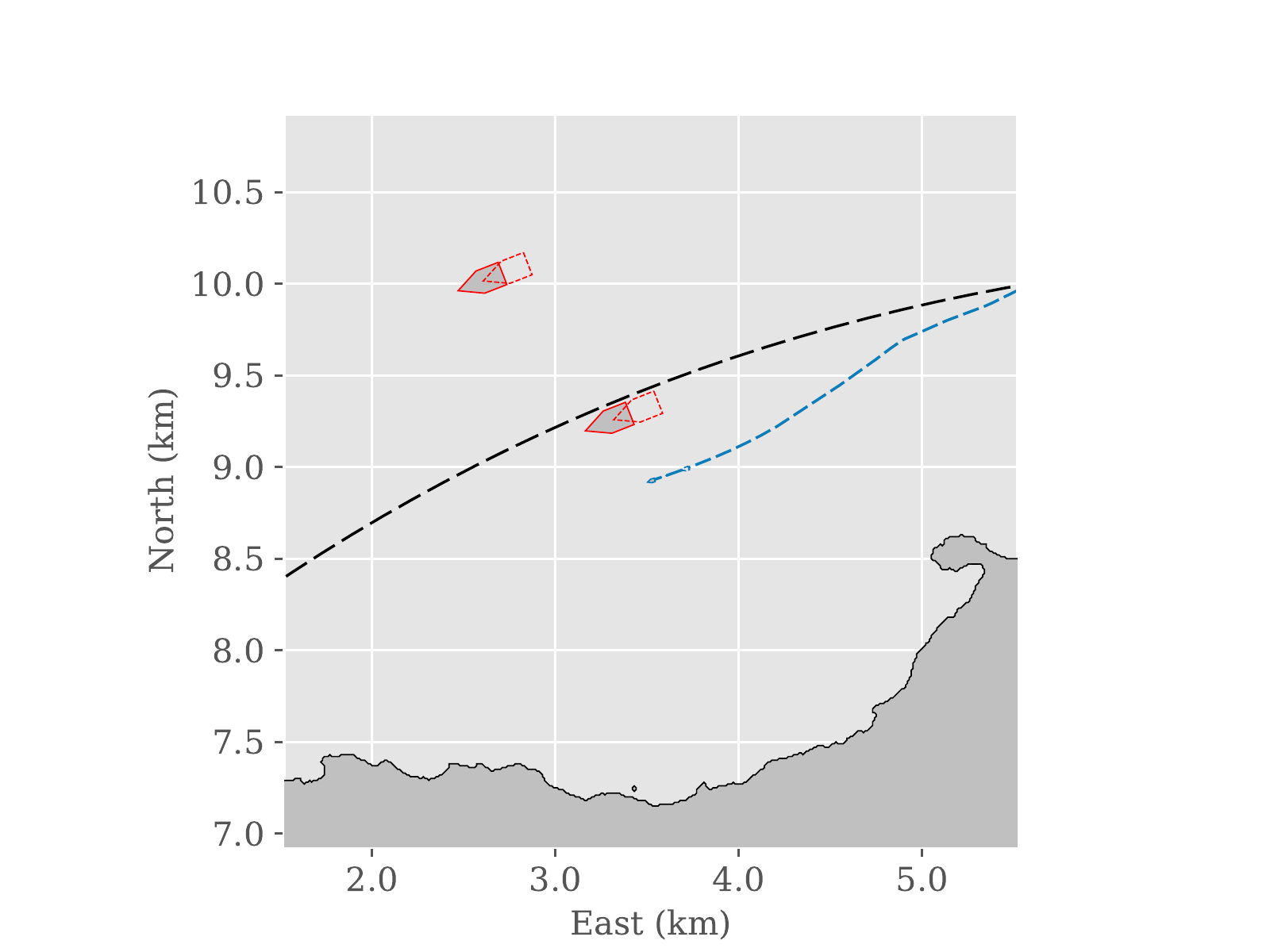}}
			\subcaption{Overtaking}
		\end{subfigure}
		\begin{subfigure}{0.49\linewidth}
			{\includegraphics[trim={0.5cm 0.1cm 1cm 1cm},clip,width=\textwidth]{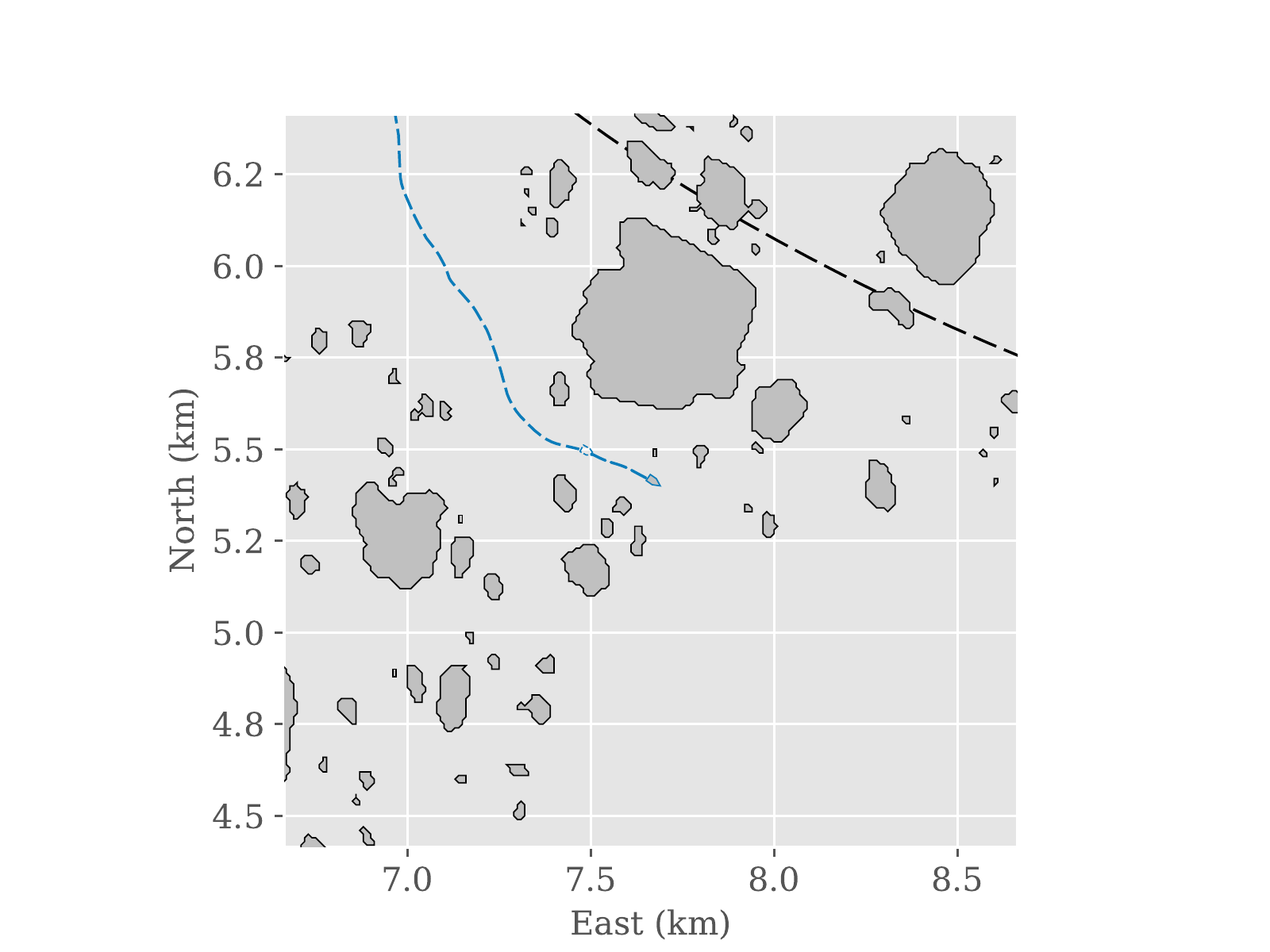}}
			\subcaption{Static COLAV}
		\end{subfigure}
		\caption{COLREGs-compliant agent performing common naval collision avoidance maneuvers in the AIS-based environment. Agent trajectories are drawn with blue dashed lines, and the target ships are drawn in red.}
		\label{fig:real_scenarios_sims_colregs}
	\end{figure}

\begin{figure}[!htb]
		\centering
		\includegraphics[trim={0cm 0cm 0 1cm},clip,width=0.5\textwidth]{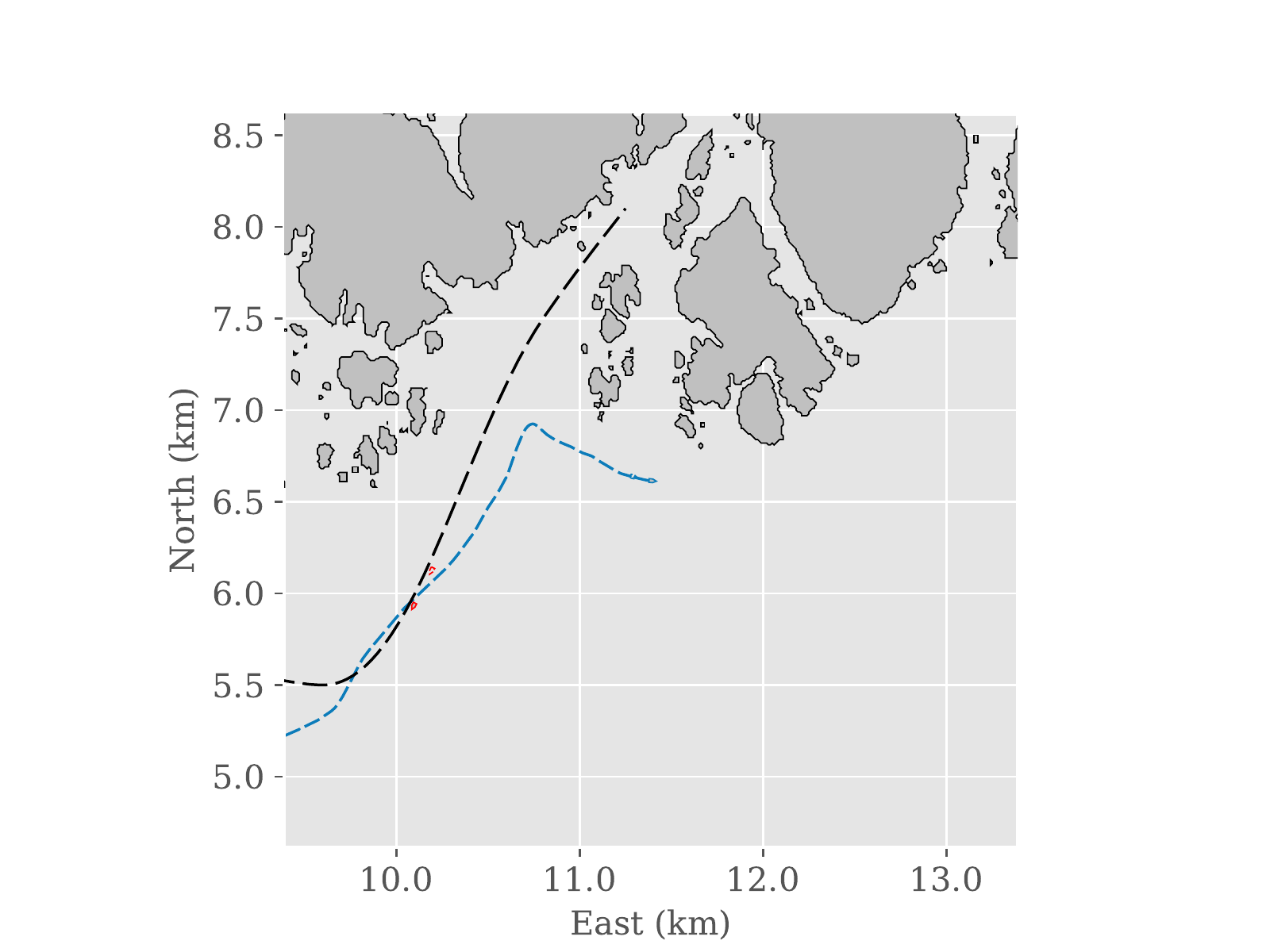}
		\caption[Training scenario simulations]{Agent (in blue) getting lost attempting to find an alternate route to the goal after encountering a target ship (in red).}
		\label{fig:sorbuoya_getting_lost}
	\end{figure}

\begin{figure}[htb!]
		\centering
		\begin{subfigure}{0.8\linewidth}
			\centering
			\hspace*{-1.0cm}\includegraphics[trim={0cm 0cm 0 1cm},clip, width=1.3\textwidth]{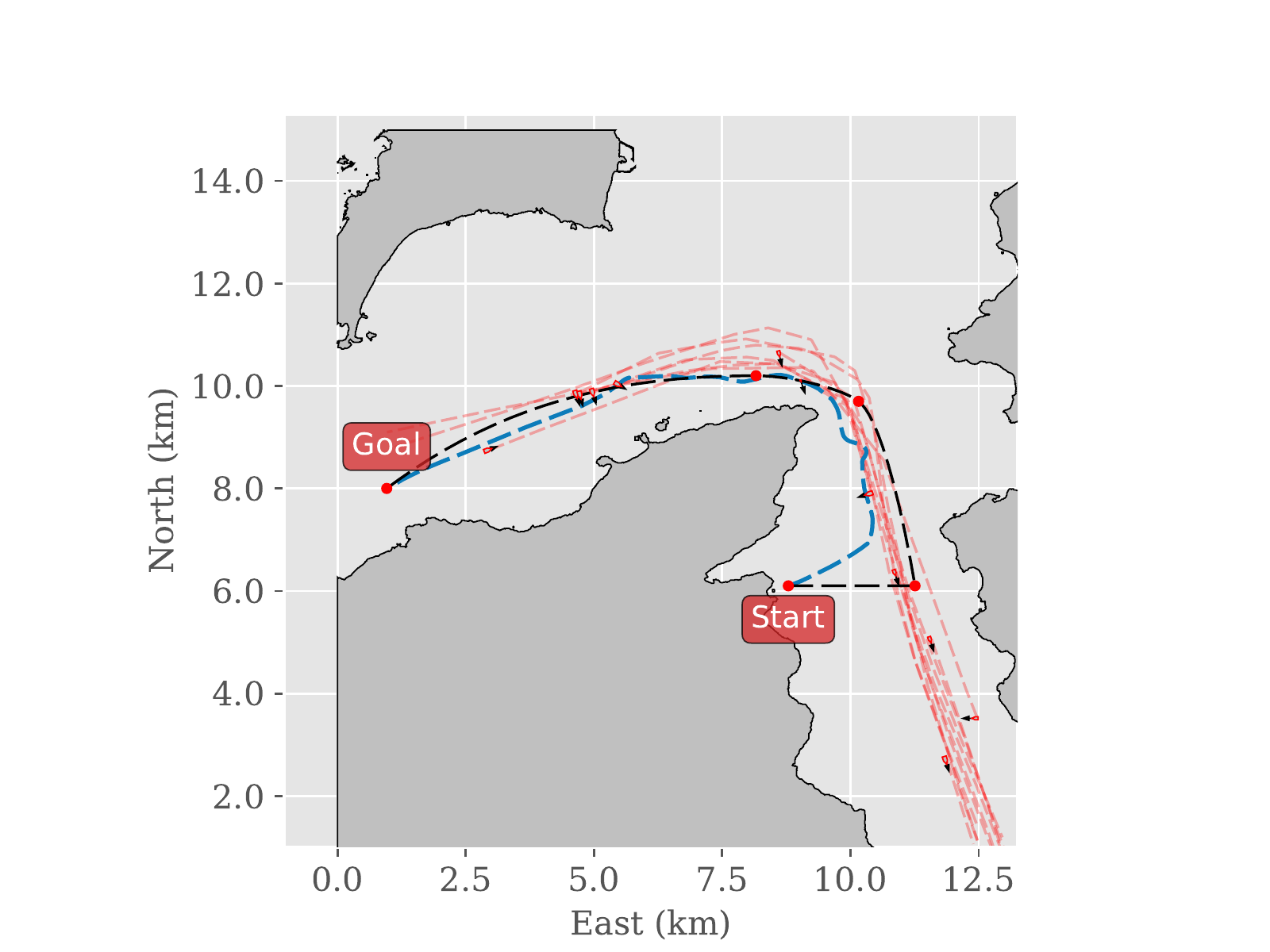}
			\subcaption[Ørland-Agdenes test scenario result]{Agent's trajectory in the Ørland-Agdenes test scenario.}
			\label{fig:orlandagdenest_testscenario_res_colregs}
		\end{subfigure}
		\begin{subfigure}{0.8\linewidth}
			\centering
			\hspace*{-1.0cm}\includegraphics[trim={0cm 0cm 0 1cm},clip,width=1.3\textwidth]{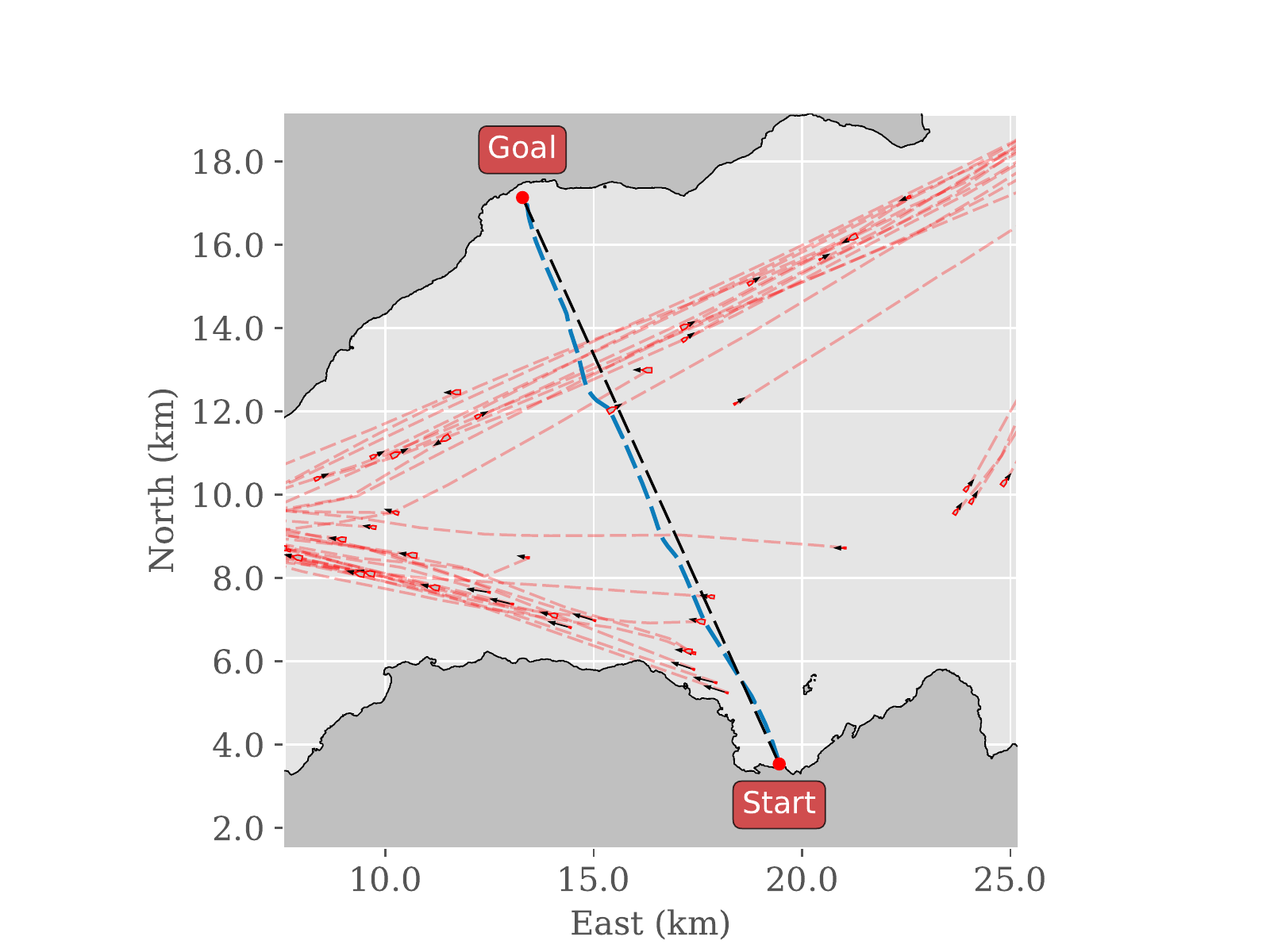}
			\subcaption[Trondheim test scenario result]{Agent's trajectory in the Trondheim test scenario.}
			\label{fig:trondheim_testscenario_res_colregs}
		\end{subfigure}
		\begin{subfigure}{0.8\linewidth}
			\centering
			\hspace*{-1.0cm}\includegraphics[trim={0cm 0cm 0 1cm},clip,width=1.3\textwidth]{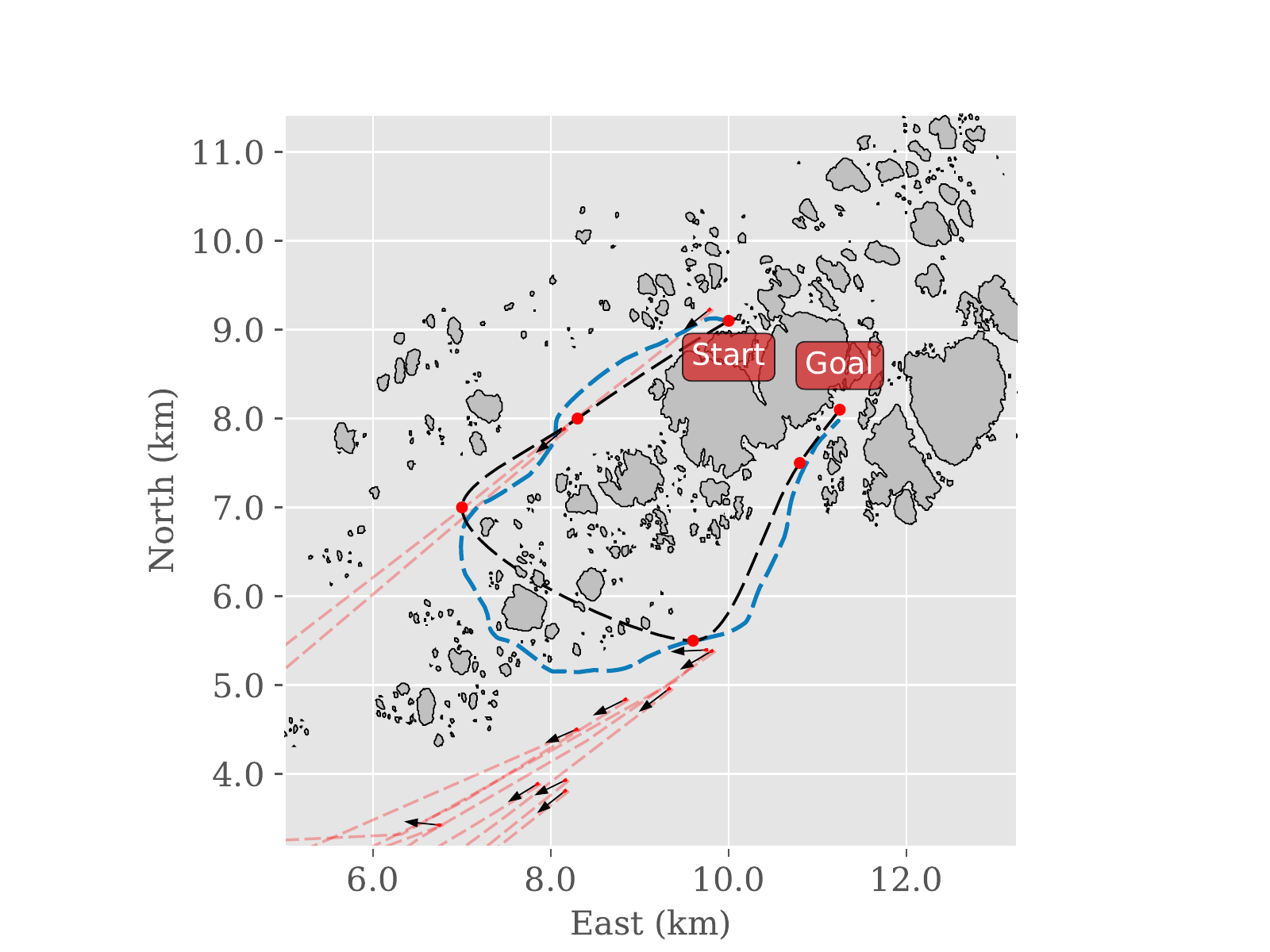}
			\subcaption[Froan test scenario result]{Agent's trajectory in the Froan test scenario.}
			\label{fig:sorbuoyatestscenario_res_colregs}
		\end{subfigure} 
		\caption{COLREGs-compliant agent trajectories in the test scenarios drawn as blue dashed lines, and target ships and trajectories are drawn in red.}
		\label{fig:real_world_sims_colregs}
	\end{figure}

	Extending the testing to scenarios based on real-world AIS data, it can be seen that the agent behaves in a COLREGs-compliant manner in situations where the COLREGs clearly define an expected behaviour. Some examples of this are presented in Figure \ref{fig:real_scenarios_sims_colregs}, where situations similar to those shown in Figure \ref{fig:training_scenario_sims} were chosen for comparison. The main difference between the training environment and the AIS-based environment, however, is the shapes and sizes of the static obstacles, which represent land and islands in the AIS-based environment. As seen in Figure \ref{fig:RW_static}, the agent has generalized sufficiently to tackle these scenarios with ease. Further, overall COLREGs-compliant trajectories undertaken by the agent in the Trondheim, Ørland-Agdenes and Froan scenarios can be seen in Figure \ref{fig:real_world_sims_colregs}. Although the agent had no issue traversing the complex geography of Froan, it struggled when encountering target ships in restricted waters (see Figure \ref{fig:sorbuoya_getting_lost}). The main explanation of this is likely that the training environment does not reflect these situations properly for the agent to be prepared for them. For instance, in the training environment, the own-ship can always sail around a circular obstacle when encountering a target ship close to such an obstacle. In the Froan scenario, this is not the case, and the own-ship is prone to get lost while attempting to find other ways to the goal. It should therefore be noted that in the scenario presented in Figure \ref{fig:sorbuoyatestscenario_res_colregs}, the own-ship did not encounter a target ship after entering the narrow end section of the desired path, but is included to showcase the agent's ability to navigate in restricted waters in the absence of target ships.
	
	%
	
	%
	
	\subsection{Conclusion}
	In this study, we demonstrated that an RL-based autonomous vessel can avoid collisions with other vessels, while at the same time follow a desired trajectory without getting stranded. With no a priori knowledge of the environment except for the waypoints of its desired path, the agent makes reactive control decisions based on rangefinder sensors measuring the distance to nearby obstacles, be it static obstacles such as the shoreline or dynamic obstacles such as other vessels. The agent was trained in an artificial, simulated environment, and evaluated in a digital reconstruction of the Trondheim Fjord area. Based on a representative sample of the marine traffic in the area, the trained vessel was evaluated by its performance in realistic encounter scenarios. Our results suggest that DRL agents, if trained in a stochastic, generic obstacle environment are capable of performing complex guidance tasks. 
	
	The successful demonstration in a simulated environment shows great promise for the viability of implementing it on a real-world vessel. As the approach requires no knowledge of the internal dynamics, and allows us to easily adapt the agent behavior by customizing the performance measure, our paper lays the groundwork for further research which may, given equally positive results, bring significant value to the field of autonomous guidance.
	\section*{Acknowledgment}
	The authors acknowledge the financial support from the Norwegian Research Council and the industrial partners: DNV GL, Kongsberg and Maritime Robotics of the Autosit project. (Grant No.: 295033).
	\bibliographystyle{unsrt}  
	\bibliography{references}
\end{document}